\documentclass{article}

\usepackage[preprint]{neurips_2026}

\usepackage[utf8]{inputenc} % allow utf-8 input
\usepackage[T1]{fontenc}    % use 8-bit T1 fonts
\usepackage{hyperref}       % hyperlinks
\usepackage{url}            % simple URL typesetting
\usepackage{booktabs}       % professional-quality tables
\usepackage{amsfonts}       % blackboard math symbols
\usepackage{nicefrac}       % compact symbols for 1/2, etc.
\usepackage{microtype}      % microtypography
\usepackage{color, xcolor}         % colors
\usepackage[most]{tcolorbox}
\usepackage{graphicx}
\usepackage{adjustbox}
\usepackage{colortbl}
\usepackage{multirow}
\usepackage{caption}
\usepackage{subcaption}
\usepackage{pifont}
\usepackage{makecell}
\usepackage{enumitem}

\definecolor{mygray}{gray}{.9}

\usepackage{xcolor}
\usepackage{pifont}
\usepackage{booktabs}
\usepackage{tabularx}
\usepackage{array}
\usepackage{ragged2e}
\usepackage[table]{xcolor}
\usepackage{amssymb}

\newcolumntype{Y}{>{\justifying\arraybackslash}X}

\definecolor{deepgreen}{RGB}{63, 126, 49}
\definecolor{deepred2}{RGB}{196, 49, 25}

\definecolor{tablegray}{RGB}{242,243,245}

\title{\textsc{HumanoidArena}: Benchmarking Egocentric Hierarchical Whole-body Learning}

\author{%
  \textbf{Taowen Wang}\textsuperscript{1,*}, 
  \textbf{Zikang Xie}\textsuperscript{1,*},
  \textbf{Bin Yang}\textsuperscript{1,*}, 
  \textbf{Yunheng Wang}\textsuperscript{1}, 
  \textbf{Zizhao Yuan}\textsuperscript{1}, 
  \\
  \textbf{Yuetong Fang}\textsuperscript{1},
  \textbf{Yixiao Feng}\textsuperscript{1}, 
  \textbf{Yichi Wang}\textsuperscript{2}, 
  \textbf{Xingyu Chen}\textsuperscript{1}, 
  \textbf{Haodong Chen}\textsuperscript{3}, 
  \textbf{Qiwei Wu}\textsuperscript{1},
  \\
  \textbf{Weisheng Xu}\textsuperscript{1}, 
  \textbf{Lihan Chen}\textsuperscript{4}, 
  \textbf{Lusong Li}\textsuperscript{5}, 
  \textbf{Zecui Zeng}\textsuperscript{5}, 
  \textbf{Renjing Xu}\textsuperscript{1,\textdagger}
  \\
  \\
  \textsuperscript{1}The Hong Kong University of Science and Technology (Guangzhou),
\\
  \textsuperscript{2}Beijing University of Technology,
  \textsuperscript{3}Harbin Institute of Technology, Shenzhen,
  \\
  \textsuperscript{4}Shenzhen MSU-BIT University,
  \textsuperscript{5}JD Explore Academy
}

\begin{document}

\maketitle
\begingroup
\renewcommand{\thefootnote}{}
\footnotetext{%
  \normalfont\footnotesize
  \begin{tabular}{@{}l@{\hspace{0.8em}}l@{}}
    \textsuperscript{*}           & Equal contribution. \\
    \textsuperscript{\textdagger} & Corresponding author.
  \end{tabular}%
}
\endgroup

\begin{abstract}

Humanoid robots promise whole-body interaction in human-centered environments, but scalable policy learning remains difficult because task-level decision-making and whole-body dynamic execution are tightly coupled. 
A practical solution is hierarchical control, where a high-level policy predicts intermediate whole-body actions and low-level general motion trackers (GMTs) execute them as stable humanoid motion.
% lack of hierarchical benchmark
However, existing benchmarks rarely evaluate the policy–tracker interface itself, leaving open whether intermediate whole-body actions are executable, robust under task distribution shifts, and transferable across different GMT backends.
% high-level definition
We introduce \textsc{HumanoidArena}, a simulation-first benchmark for egocentric hierarchical whole-body learning. The benchmark formulates policy learning as a hierarchical decision making problem: a high-level policy converts egocentric vision, proprioception, and instructions into a compact whole-body action, which is subsequently executed by a low-level GMT. 
% details design choose 
% Task-level (leg / HOI/ HSI)
Instead of treating the legs as planar transport tools, 
\textsc{HumanoidArena} emphasizes interactions where lower-body coordination is structurally necessary in task completion. 
We therefore design 7 leg-critical HOI/HSI tasks in which success requires foot placement, balance maintenance, posture adjustment, and whole-body reorientation. 
% details design choose 
% evalutation/generalization-level (pertubation dignosis, cross-gmt)
To further diagnose the hierarchical system, we evaluate policies from two complementary perspectives: perturbation-conditioned generalization and GMT-conditioned transfer.
% \textsc{HumanoidArena} evaluates policies along visual, semantic, and execution generalization axes, and introduces in-GMT and cross-GMT protocols for measuring policy--tracker compatibility. 
We benchmark representative imitation-learning and VLA-style policies under this shared interface. Experiments show that hierarchical control enables learned policies to solve diverse leg-critical interactions, but performance is strongly tracker-conditioned and \textit{cross-GMT} transfer remains fragile. These results position \textsc{HumanoidArena} as a benchmark for studying transferable intermediate action representations and scalable egocentric whole-body policy learning. Code and data are available at: \url{https://humanoidarena.github.io}.

\end{abstract}

\section{Introduction}

\begin{quote}
   \hspace{-1em}
   \begin{minipage}{0.85\textwidth} 
   \itshape
   ``The body is our general medium for having a world.'' \\
   \mbox{}\hfill \textemdash\ Maurice Merleau-Ponty \cite{merleau1996phenomenology}
   \end{minipage}
\end{quote}

A central goal of embodied intelligence is to build agents that can perceive, reason, and act in the physical world through their bodies~\cite{embodied1991}. 
Humanoid robots are a natural embodiment for human-centered environments because their morphology aligns with spaces, objects, tools, and affordances designed around the human body. This alignment is most consequential in interaction tasks where the lower body is not merely a locomotion module that transports the hands to a workspace. Foot placement, support transitions, balance-aware posture, and whole-body reorientation can directly determine whether an object or scene interaction is feasible~\cite{stilman2008humanoid,borras2015wholebody,OmniRetarget}. In such \emph{leg-critical} interaction settings, lower-body coordination is structurally involved in task completion rather than reducible to planar base motion. Evaluating humanoid intelligence under this setting requires policies that act through an integrated body under egocentric perception, rather than treating locomotion and manipulation as separable subproblems.

This whole-body requirement exposes a central tension at the interface between task intent and dynamic realization.
A high-level policy must produce actions that preserve task-relevant intent, such as which object or scene affordance to approach and what interaction to attempt, while the execution layer must realize these intentions as stable, coordinated motion on a high-DoF legged body under contact and viewpoint changes. This division suggests different learning regimes: visual and semantic policy learning benefits from scalable imitation learning~\cite{kim24openvla,pi05}, whereas dynamics-level execution benefits from fast reactive controllers trained with dense physical feedback~\cite{beyondmimic,konggubot,wang2026omnixtreme}. 
Recent hierarchical humanoid systems~\cite{ze2025twist2,luo2025sonic} make this division operational by letting high-level egocentric policies specify task-directed whole-body intent and low-level trackers realize that intent as dynamically feasible motion. This hierarchy provides a powerful abstraction for scalable humanoid control, decoupling visual-semantic decision making from high-frequency whole-body stabilization while retaining an executable intermediate representation for coordinated behavior.

Despite this progress, existing benchmarks do not yet provide a controlled setting for studying hierarchical humanoid whole-body learning. First, according to \emph{interface}, most systems evaluate an end-to-end control stack~\cite{HumanoidBench,leverb} or a fixed action representation~\cite{ze2025twist2,luo2025sonic}, making it difficult to separate high-level policy learning from low-level execution behavior. Without an explicit policy-facing whole-body action space and GMT-conditioned execution layer, comparisons are easily confounded by controller design, tracker dynamics, and backend-specific failure modes. Second, for \emph{task}, many embodied and humanoid benchmarks emphasize locomotion~\cite{HumanoidBench}, manipulation~\cite{li2022behavior,liu2023libero}, or specific task diversity~\cite{HumanoidVerse}, but do not center leg-critical interactions where lower-body coordination is structurally required for success. Third, for \emph{evaluation}, existing protocols rarely disentangle visual perception, semantic grounding, spatial execution, and tracker-induced deployment shifts within the same benchmark. As a result, task success under one controller and one distribution can obscure whether a policy has learned robust hierarchical whole-body behavior.

To address this gap, we introduce \textsc{HumanoidArena}, a simulation-first benchmark for egocentric hierarchical whole-body learning. For the \emph{interface}, \textsc{HumanoidArena} makes the hierarchical decomposition explicit: high-level policies predict compact intermediate whole-body actions from egocentric observations, proprioception, and task instructions, while low-level GMTs execute these actions as dynamically feasible humanoid motions. 
For the \emph{task} design, the benchmark instantiates seven leg-critical Human-Object Interaction (HOI) / Human-Scene Interaction (HSI) tasks that require lower-limb participation as an essential part of interaction, rather than treating the legs merely as a navigation module. 
For \emph{evaluation}, \textsc{HumanoidArena} defines controlled perturbation-conditioned evaluation protocols across visual, semantic, and execution shifts and further supports \textit{in/cross-GMT} deployment to assess both policy robustness and policy--tracker compatibility. 
Together, these design choices establish \textsc{HumanoidArena} as a principled benchmark for egocentric hierarchical whole-body humanoid learning.
Specifically, we make the following contributions.

\textbf{\ding{182}} \textbf{Hierarchical benchmark formulation.} We introduce a shared hierarchical interface for egocentric humanoid whole-body learning, where high-level policies predict intermediate whole-body actions and low-level general motion trackers execute them as dynamically feasible motions.

\textbf{\ding{183}} \textbf{Leg-critical egocentric task suite and data pipeline.} We design seven leg-critical HOI/HSI tasks in which successful execution requires lower-body participation, balance maintenance, posture adjustment, and whole-body coordination. 
Built in simulation, our GMT-based teleoperation pipeline collects closed-loop egocentric demonstrations while supporting scalable task and scene variation, failure-rich exploration, and synchronized first-person and third-person observations.

\textbf{\ding{184}} \textbf{Controlled hierarchical evaluation protocols.} We introduce perturbation-conditioned evaluation protocols across visual, semantic, and execution shifts. 
Together with \textit{in/cross-GMT} deployment, these protocols make policy robustness and policy–tracker compatibility measurable under a shared intermediate whole-body action interface.

\textbf{\ding{185}} \textbf{Baselines and findings.} We benchmark representative imitation-learning and VLA-style policies across tasks, trackers, and perturbations. Our results show that GMT-based hierarchy enables learned policies to complete diverse leg-critical interactions, while performance remains sensitive to the choice of execution backend. This sensitivity highlights \textit{cross-GMT} transfer and transferable intermediate action representations as important open challenges for hierarchical whole-body learning.
\begin{figure*}[!t]
    \centering
    \includegraphics[width=1\linewidth]{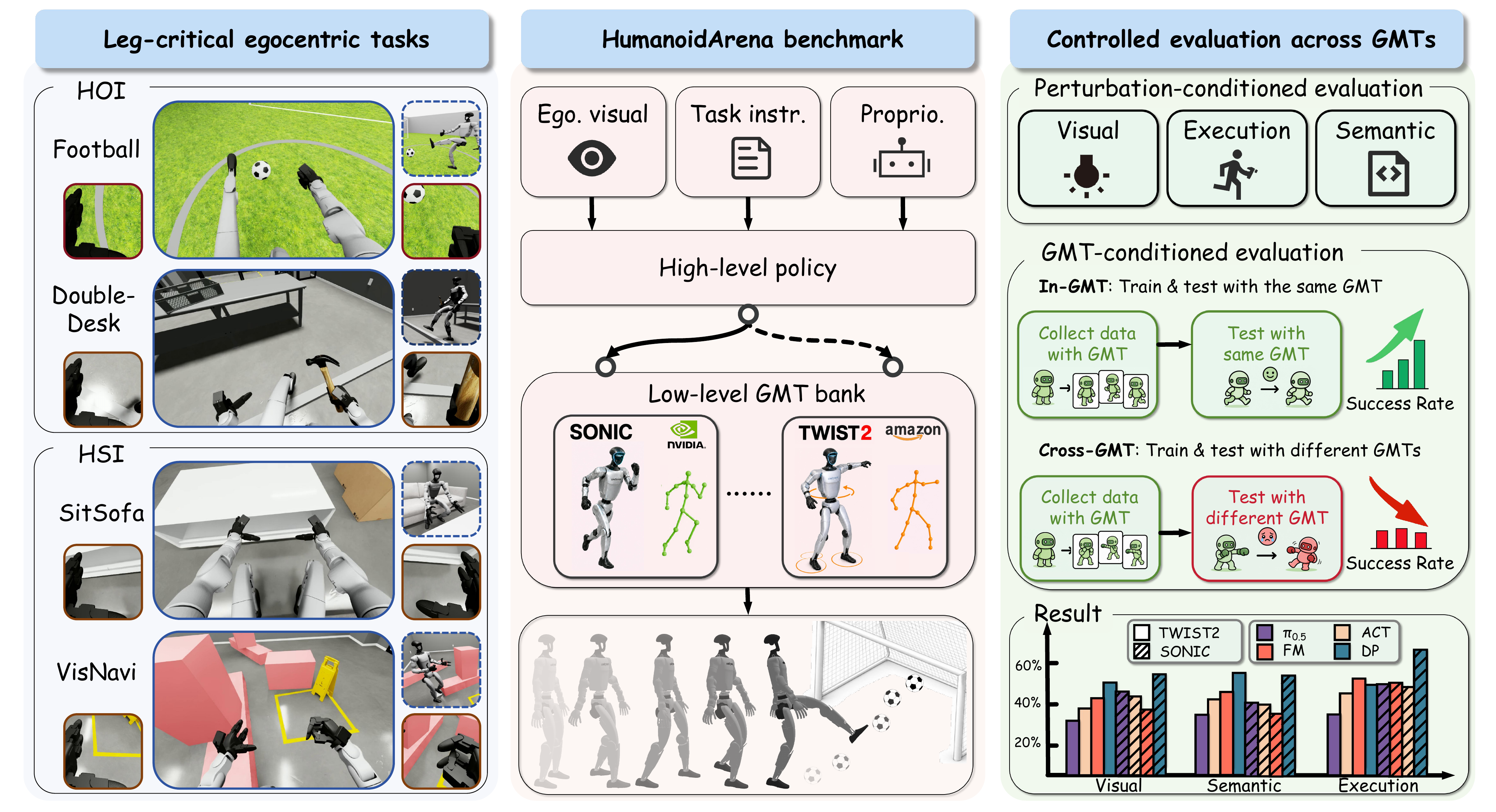}
    \caption{\textbf{Overview of \textsc{HumanoidArena}.}
The benchmark studies leg-critical HOI/HSI tasks where success requires coordinated perception, foot placement, balance, and whole-body motion.
Within this hierarchical formulation, high-level policies predict intermediate whole-body actions from egocentric visual observations, task instructions, and proprioception, while low-level GMTs stabilize and track them into feasible humanoid motions.
To evaluate this interface, perturbation-conditioned evaluation and GMT-conditioned evaluation diagnose task generalization, tracker-induced distribution shifts, and policy--tracker compatibility.
}
    % \vspace{-20pt}
    \label{fig:main}
\end{figure*}

\section{Related Work}
\label{gen_inst}

\textbf{Egocentric Humanoid Whole-Body Interaction.} 
Recent advances in humanoid whole-body control have gradually transitioned from proprioceptive motion tracking toward perception-driven interaction. Prior arts~\cite{OmniRetarget, hdmi, zhao2025resmimic} achieved significant milestones by reinforcement learning, and recent work has expanded into the field of underactuated object interaction~\cite{li2026haic}. To ensure perception accuracy, several approaches rely on external sensors~\cite{HumanX, su2025hitter, zhang2026learning}, while egocentric perception has increasingly emerged for HOI~\cite{wang2025physhsi, yin2025visualmimic, he2025viral, lin2026prohoi, lin2026lessmimic, he2026ultra, ren2026smash}. In parallel, egocentric vision has also proven crucial for HSI, empowering humanoids to navigate various terrains~\cite{rudin2025parkour, liu2025ego, ben2025gallant, zhu2026hiking, zhuang2026deep, wu2026perceptive}. These works demonstrate impressive interaction capabilities, since they inherently couple visual perception, decision-making, and whole-body execution within one stack, making it difficult to evaluate the generalization of the learned intent across different execution backends.

A second line of work makes the hierarchy explicit.
TWIST2~\cite{ze2025twist2} uses a general motion tracker as the execution layer for egocentric whole-body teleoperation and visuomotor policy learning, while SONIC~\cite{luo2025sonic} scales motion tracking into a general low-level controller that can interface with multiple input modalities. Other humanoid VLA and whole-body policy systems instantiate the high-level/low-level split through different intermediate abstractions: motion generation followed by tracking~\cite{zerowbc,PhysiFlow}, latent vision-language verbs with RL whole-body skills~\cite{leverb}, or VLA/action experts paired with specialized lower-body or whole-body controllers~\cite{HumanoidVLA,jiang2025wholebodyvla,Psi0}. 
Together, these works mark a clear shift toward hierarchical whole-body learning, where high-level models handle egocentric perception, language, and task intent, while low-level controllers maintain dynamic feasibility.
However, hierarchical whole-body learning still lacks a dedicated benchmark.
Existing systems vary in tasks, data, action spaces, and GMT backends, making it difficult to evaluate high-level policies under a shared policy--tracker interface.
\textsc{HumanoidArena} fills this gap by turning egocentric hierarchical whole-body learning into a controlled benchmark.

\textbf{Humanoid Benchmarks and Platforms.} Broader robot-learning benchmarks have greatly expanded task diversity and standardized evaluation. 
LIBERO~\cite{liu2023libero} studies knowledge transfer in language-conditioned manipulation with 130 procedurally generated tasks, while BEHAVIOR-1K~\cite{li2022behavior} scales household activity evaluation to 1,000 tasks in realistic simulation. 
However, they either rely on fixed-arm manipulation setups or abstract mobility into simplified navigation actions, leaving hierarchical whole-body control with humanoid full-body execution largely unstudied.
A related line of work studies whole-body mobile manipulation beyond fixed-arm settings. 
BEHAVIOR Robot Suite~\cite{jiang2025brs} targets real-world household whole-body manipulation with a bimanual wheeled robot and a 4-DoF torso, emphasizing bimanual coordination, stable navigation, and end-effector reachability. 
AgentWorld~\cite{zhang2025agentworld} supports scalable scene construction and mobile manipulation data collection with both wheeled bases and humanoid locomotion policies. 
These platforms expand the scope of mobile manipulation, but their lower-body control is either wheeled or abstracted through locomotion/navigation interfaces, rather than evaluated through leg-critical humanoid execution under a GMT-based hierarchical interface.

Furthermore, HumanoidBench~\cite{HumanoidBench} provides a high-dimensional simulated benchmark for humanoid locomotion and manipulation. LeVERB~\cite{leverb} introduces a sim-to-real-ready vision-language closed-loop benchmark for humanoid whole-body control and a latent hierarchical policy,
directly motivating the need for expressive high-level action interfaces. HumanoidVerse~\cite{HumanoidVerse} studies egocentric language-guided multi-object rearrangement across 350 tasks, while Humanoid Everyday~\cite{zhao2025humanoideveryday} contributes a large real-world humanoid manipulation dataset with multimodal sensing and cloud-based evaluation. Although these resources substantially advance humanoid whole-body learning, they do not explicitly benchmark GMT-based hierarchical control.
\textsc{HumanoidArena} complements them by making the policy--tracker interface itself a benchmark axis, enabling systematic evaluation of high-level policies across task-relevant perception, grounding, spatial execution, and tracker-induced deployment shifts.

\vspace{-10pt}
\section{\textsc{HumanoidArena}}
\vspace{-5pt}
\label{headings}

This section introduces \textsc{HumanoidArena} by first outlining the research topics that motivate its design (Sec.~\ref{sec:researchtopic}).
We then instantiate these topics through four benchmark components: a hierarchical whole-body protocol that couples high-level policies with low-level general motion trackers (Sec.~\ref{sec:protocol}), a closed-loop egocentric data collection pipeline for multi-GMT demonstrations (Sec.~\ref{sec:pipeline}), leg-critical HOI/HSI task suites (Sec.~\ref{sec:tasks}), and evaluation protocols for perturbation-conditioned and GMT-conditioned generalization (Sec.~\ref{sec:eval}).

\subsection{Research Topics in HumanoidArena}
\label{sec:researchtopic}

\textbf{(T1) Goal-directed hierarchical humanoid interaction.}
Humanoid robots are expected to accomplish goals through coordinated whole-body interaction rather than isolated locomotion or manipulation. In leg-critical HOI/HSI tasks, success may depend on foot placement, balance-aware posture adjustment, whole-body reorientation, and coordinated arm--leg motion. \textsc{HumanoidArena} provides a hierarchical setting in which a high-level policy predicts intermediate whole-body actions from egocentric observations and task instructions, while a low-level general motion tracker executes them into dynamically feasible motion.

\textbf{(T2) Robust policy learning with egocentric perception.}
A key question is how robust policies remain when task execution depends on egocentric visual observations.
\textsc{HumanoidArena} enables this study through three controlled generalization dimensions, visual, semantic, and execution, which test robustness to lighting changes, similar distractors, and various object initializations.

\textbf{(T3) \textit{Cross-GMT} generalization in hierarchical control.}
Hierarchical control raises the question of how policy outputs change when executed by different low-level general motion trackers.
Through a shared intermediate whole-body action interface, \textsc{HumanoidArena} offers a \textit{cross-GMT} testbed where policies can be trained with one tracker and deployed with another, directly evaluating whether intermediate actions remain valid across different GMTs.

\textbf{(T4) Multimodal coordination under whole-body motion.}
Humanoid perception is tightly coupled with body motion. Unlike stationary-arm manipulation, humanoid egocentric and wrist cameras can undergo drastic viewpoint changes due to walking, arm swing, torso rotation, squatting, balance recovery, and contact-rich interaction. \textsc{HumanoidArena} enables research on multimodal fusion and visual--proprioceptive alignment under rapidly changing egocentric viewpoints.

\subsection{Hierarchical Whole-Body Protocol}
\label{sec:protocol}
\textsc{HumanoidArena} adopts a hierarchical whole-body control protocol that separates perception-conditioned decision making from low-level dynamic execution. At time step $t$, a high-level policy receives an egocentric observation $I^{\mathrm{ego}}_t$, proprioception $p_t$, and a task instruction $\ell$. The proprioception $p_t$ consists of the root orientation in 6D rotation representation, the 29D G1 joint positions, and their corresponding joint velocities. Then the high-level policy predicts a 40D intermediate whole-body action $u_t \in \mathbb{R}^{40}$. This action specifies root movement in the horizontal plane, root height, root orientation, 29D body joint targets, and binary open/close commands for both hands. This action serves as a tracker-compatible interface that specifies the intended whole-body motion to be executed. A low-level GMT then tracks this action and converts it into dynamically feasible humanoid motion. This separation allows \textsc{HumanoidArena} to evaluate not only task success, but also the tracker robustness of intermediate whole-body actions under different GMT backends.

\textbf{High-level policy baselines.}
The high-level policy $\pi_\theta$ is the main learning component evaluated in \textsc{HumanoidArena}. At each policy step, it receives egocentric visual observations, task instructions, and a 64D canonical proprioceptive state, and predicts a 40D intermediate whole-body action:
\begin{equation}
    u_t \sim \pi_\theta(\cdot \mid I^{\mathrm{ego}}_t, p_t, \ell), 
\quad p_t \in \mathbb{R}^{64}, \quad u_t \in \mathbb{R}^{40}.
\end{equation}

We benchmark $\pi_\theta$ with four representative visuomotor policy families: ACT~\cite{ACT}, Diffusion Policy~\cite{chi2023diffusionpolicy}, Flow Matching~\cite{FM_objective}, and $\pi_{0.5}$~\cite{pi05}, covering chunked action prediction, diffusion-based modeling, multi-step action prediction, and VLA-style instruction-conditioned policy learning. Since high-dimensional action spaces substantially increase learning complexity, we adopt a compact 40D canonical action interface instead of an SMPL-based action representation. This interface still requires policies to predict coordinated whole-body actions over the root, torso, arms, legs, and hands under egocentric perception, while keeping the action space tractable for systematic baseline evaluation. All baselines share the same observation space, action interface, and GMT backends, so performance differences primarily reflect policy design rather than changes in the execution interface.

\textbf{Low-level general motion tracker bank.}
The low-level layer contains a bank of GMTs that execute the canonical whole-body action predicted by the high-level policy. Since each GMT follows its own input protocol, $u_t$ is first mapped by a GMT-specific adapter $\psi_m$ before being passed to GMT $G^{(m)}$:
\begin{equation}
q^{(m)}_{t+1} = G^{(m)}\!\left(\psi_m(u_t)\right),
\end{equation}
where $q^{(m)}_{t+1}$ is the G1 joint-position target produced by the $m$-th GMT. The adapter $\psi_m$ implements the required conversion from the canonical action space to the tracker-specific reference format. After this conversion, each GMT applies its own tracking policy to produce executable joint targets. Since trackers differ in motion priors, stabilization strategies, and failure modes, this bank enables evaluation of tracker robustness under a shared high-level interface.

\subsection{Data Collection Pipeline}
\label{sec:pipeline}

\textsc{HumanoidArena} collects data in Isaac Lab~\cite{mittal2025isaaclab} through a VR-based egocentric teleoperation pipeline, where the operator receives the humanoid's egocentric visual stream and controls the robot in closed loop via GMR~\cite{joao2025gmr,ze2025twist}. The retargeted robot reference motion is canonicalized into the policy-facing action form defined in Section~\ref{sec:protocol}, and then executed by a selected GMT. Under the same teleoperation and action interface, switching the GMT backend yields multi-GMT data with shared policy-facing action semantics but different low-level tracking dynamics. In our implementation, to reduce retargeting errors caused by height variation across operators, both backends are conditioned on robot reference motions: TWIST2~\cite{ze2025twist2} adopts a mimic-style tracking command, while SONIC~\cite{luo2025sonic}, which supports multiple conditioning interfaces, is used in its robot motion encoder mode rather than its human motion encoder mode. The full data specification is provided in Appendix~\ref{app:data_spec}.

\subsection{Leg-Critical Egocentric Task Suites}
\label{sec:tasks}
\vspace{-2pt}

\textsc{HumanoidArena} focuses on leg-critical egocentric tasks, by which we refer to HOI/HSI settings where lower-body coordination is directly involved in task completion rather than serving only as planar transport for the upper body.
In contrast to mobile-base manipulation settings, these tasks require the policy to predict intermediate whole-body actions that jointly coordinate root motion, body posture, legs, arms, and hands.
Detailed task design, including layouts, initialization ranges, success conditions, and the lower-body requirements of each task, is provided in Appendix~\ref{app:tasks}.

The \textbf{HOI} tasks test how legged motion changes object interaction. \textsc{HOI-DoubleDesk} couples cross-surface object transfer with terrain-aware stepping across a height discontinuity. \textsc{HOI-Football} evaluates coordinated leg-object interaction, where success depends on visual localization, balance, and kicking motion. \textsc{HOI-P\&PBox} tests leg-assisted high placement, requiring the robot to bend its legs and adjust whole-body posture to reach the target shelf height.

The \textbf{HSI} tasks test how to interact with scenes. \textsc{HSI-VisNavi} evaluates obstacle-aware egocentric navigation in constrained space. \textsc{HSI-OpenDoor} couples handle manipulation with body turning and doorway traversal. \textsc{HSI-SitSofa} requires obstacle avoidance, body alignment, and a stable sitting transition. \textsc{HSI-Boxing} evaluates height-adaptive striking, where low targets require crouching and whole-body adjustment. Together, these HOI and HSI tasks probe complementary forms of leg-critical interaction, from terrain-aware object transfer to posture-dependent scene interaction.

\vspace{-4pt}
\subsection{Evaluation Protocols}
\vspace{-2pt}
\label{sec:eval}

\textbf{Perturbation-conditioned evaluation.}
\textsc{HumanoidArena} defines three perturbation axes for diagnosing policy generalization: \textit{visual}, \textit{semantic}, and \textit{execution}.
These axes are designed to isolate failures in scene perception, target grounding, and whole-body execution, respectively.
\textit{Visual generalization} changes the lighting direction while preserving the goal and object identity, testing robustness to egocentric appearance shifts.
\textit{Semantic generalization} introduces semantically similar distractors under the same instruction, testing whether the policy grounds the commanded target.
\textit{Execution generalization} expands the initialization range of task-relevant objects, requiring policies to adapt foot placement, body pose, and whole-body motion to new spatial conditions. Full perturbation configurations and rollout seeds are detailed in Appendix~\ref{app:eval_protocol}.

\textbf{GMT-conditioned evaluation.}
Enabled by the shared whole-body action interface defined in Section~\ref{sec:protocol}, 
\textsc{HumanoidArena} introduces GMT-conditioned protocols for evaluating policy--tracker compatibility.
This design allows the same policy-facing action space to be executed by different low-level GMT backends, making \textit{cross-GMT} transfer a measurable property rather than an implementation detail.
In the \textit{in-GMT} protocol, training and inference use the same GMT, measuring performance under matched tracking dynamics.
In the \textit{cross-GMT} protocol, policies are trained with one GMT and deployed with another, testing whether the learned intermediate actions encode tracker-transferable whole-body intent or overfit to the motion priors, behavior, and failure modes of a single GMT.
Together, these protocols assess \textit{cross-GMT} transfer of intermediate whole-body actions and reveal compatibility failures that are hidden under standard matched-backend evaluation.

\vspace{-4pt}
\section{Experiments}
\vspace{-2pt}
In this section, we conduct experiments as an initial study of the proposed benchmark setting and hierarchical control formulation. We first introduce the evaluation protocols and then analyse empirical results across tasks, trackers, and generalization settings. Our experiments focus on the following research questions:
\begin{itemize}
    \item[\textbf{Q1:}] How well do existing imitation-learning and VLA-style policy baselines perform on leg-critical egocentric HOI/HSI tasks?
    \item[\textbf{Q2:}] How robust are different policy models under visual, semantic, and execution perturbations?
    \item[\textbf{Q3:}] How do high-level policies trained on demonstrations collected with a single GMT perform when deployed with different GMT backends?
\end{itemize}

\begin{figure*}[!t]
    \centering
    \includegraphics[width=1\linewidth]{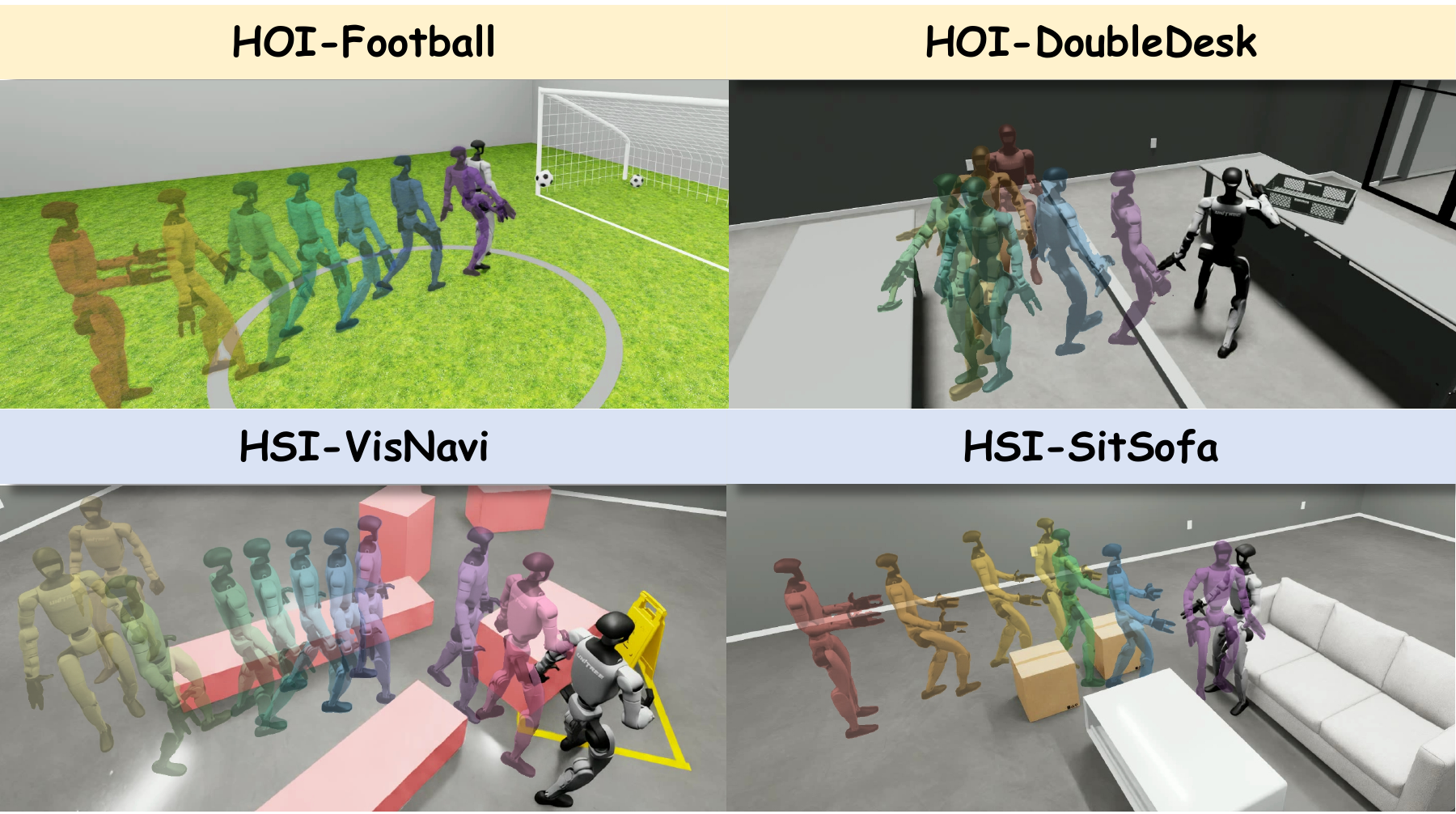}
    \caption{\textbf{Qualitative rollouts.} We visualize four representative successful episodes generated by Diffusion Policy with SONIC as the low-level GMT. The examples cover both HOI and HSI tasks, showing that the high-level policy can coordinate egocentric perception, task interaction, foot placement, and whole-body motion through the shared GMT-based execution interface.}
    \vspace{-23pt}
    \label{fig:qualitative}
\end{figure*}
\vspace{-8pt}

\subsection{Experimental Setup}
\vspace{-3pt}
\textbf{Evaluation matrix.}
We evaluate policies across task categories, GMT backends, policy architectures, and perturbations. 
The task suite contains leg-critical HOI and HSI tasks introduced in Section~\ref{sec:tasks}; we report both category-level performance and the overall average. 
For each task, all methods use the same demonstration split and are evaluated on the same held-out initializations. Each experimental configuration is evaluated with 3 random seeds and 20 rollout trials per seed, resulting in 60 rollouts per configuration.
To study robustness beyond the in-distribution setting, we evaluate on the visual, semantic, and execution generalization splits defined in Section~\ref{sec:eval}.

\textbf{Policy instantiations and training.}
We instantiate the high-level policy with four representative policy families: ACT~\cite{ACT}, Diffusion Policy~(DP,~\cite{chi2023diffusionpolicy}), Flow Matching~(FM,~\cite{FM_objective}), and $\pi_{0.5}$~\cite{pi05}. 
All policies receive the same egocentric RGB observation, proprioceptive state, and task instruction, and predict the same 40D intermediate whole-body action. 
This keeps the observation space, action interface, and GMT execution layer fixed, so that performance differences primarily reflect policy design rather than changes in the control interface. 
For fair comparison, all policies are trained with the same number of demonstrations per task and the same total training budget of 100k gradient steps. 
Model-specific training details are provided in Appendix~\ref{app:implementation}.

\textbf{GMT deployment protocols.}
We evaluate each policy under multiple GMT deployment settings. 
In the \textit{in-GMT} protocol, training and inference use the same GMT, measuring standard task performance under a matched data-collection and GMT backend. 
In the \textit{cross-GMT} protocol, policies are trained on demonstrations collected with one GMT and deployed with another, measuring whether the learned high-level policy and intermediate action representation transfer across GMT backends.

\textbf{Metrics.}
We utilize success rate~(SR) as the primary metric across all tasks and evaluation protocols.
For cross-GMT evaluation, we report the absolute cross-GMT drop 
$\Delta_{\mathrm{cross}}=\mathrm{SR}_{\mathrm{in}}-\mathrm{SR}_{\mathrm{cross}}$ 
and relative transfer retention 
$\mathrm{Ret}=\mathrm{SR}_{\mathrm{cross}}/\mathrm{SR}_{\mathrm{in}}$,
where $\mathrm{SR}_{\mathrm{in}}$ is the matched in-GMT success rate and $\mathrm{SR}_{\mathrm{cross}}$ is the success rate after replacing the deployment GMT.
For visual, semantic, and execution generalization, we report success rate on each held-out split together with the relative performance drop from the in-distribution split.
In addition to task completion, we report average fall rate~(AFR) as a key stability metric, since successful humanoid interaction requires not only reaching task goals but also maintaining balance and dynamically feasible whole-body motion throughout execution.

\textbf{Implementation and reproducibility.}
To ensure that comparisons reflect policy design rather than evaluation artifacts, all methods are evaluated with identical rollout seeds, held-out object configurations, and initial robot states.
All policies share the same visual preprocessing, canonical proprioceptive input, and intermediate whole-body action interface.
Detailed training and deployment settings, including policy frequency, GMT control frequency, action chunking, optimization hyperparameters, and compute resources, are provided in Appendix~\ref{app:implementation}.

\begin{table*}[t]
  \caption{
  \textbf{\textit{In-GMT} evaluation.}
  We evaluate high-level policy baselines with matched training and inference GMTs, using TWIST2 or SONIC as the low-level execution backend. Results are reported as success rate (SR, $\uparrow$; mean $\pm$ standard deviation). AVG denotes the average SR within each HOI or HSI task suite.
  The best and second-best AVG are highlighted in \textbf{bold} and \underline{underlined}, respectively.
  }
  \label{table:ingmt}
  \centering
  \begin{adjustbox}{max width=\textwidth}
  \setlength{\tabcolsep}{4.2pt}
  \renewcommand{\arraystretch}{1.08}

  \begin{tabular}{ccccccc|ccccc}
  \toprule
  & & &
  \multicolumn{4}{c|}{HOI} &
  \multicolumn{5}{c}{HSI} \\
  \cmidrule(lr){4-7} \cmidrule(lr){8-12}

  \multirow{-2}{*}{Difficulty} &
  \multirow{-2}{*}{Method} &
  \multirow{-2}{*}{AFR~($\downarrow$)} &
  \textsc{Football} & \textsc{DoubleDesk} & \textsc{P\&Pbox} & \textsc{Avg~($\uparrow$)} &
  \textsc{OpenDoor} & \textsc{SitSofa} & \textsc{Boxing} & \textsc{VisNavi} & \textsc{Avg~($\uparrow$)} \\
  \midrule

  \multirow{4}{*}{TWIST2}     & ACT~\cite{ACT}                          & 6.43\%  & 26.7$\pm$6.2\% & 15.0$\pm$8.2\% & 43.3$\pm$8.5\% & \cellcolor{mygray}28.33$\pm$13.94\% & 50.0$\pm$4.1\% & 66.7$\pm$10.3\% & 58.3$\pm$10.3\% & 13.3$\pm$4.7\% & \cellcolor{mygray}47.08$\pm$21.84\% \\
                              & DP~\cite{chi2023diffusionpolicy}        & 10.48\% & 46.7$\pm$6.2\% & 10.0$\pm$4.1\% & 50.0$\pm$0.0\% & \underline{\cellcolor{mygray}35.56$\pm$18.63\%} & 68.3$\pm$6.2\% & 73.3$\pm$6.2\% & 51.7$\pm$11.8\% & 30.0$\pm$4.1\% & \underline{\cellcolor{mygray}55.83$\pm$18.58\%} \\
                              & FM~\cite{FM_objective}                  & 7.38\%  & 53.3$\pm$8.5\% & 16.7$\pm$8.5\% & 38.3$\pm$8.5\% & \textbf{\cellcolor{mygray}36.11$\pm$17.28\%} & 76.7$\pm$2.4\% & 68.3$\pm$6.2\% & 63.3$\pm$9.4\% & 26.7$\pm$20.9\% & \textbf{\cellcolor{mygray}58.75$\pm$22.56\%} \\
                              & $\pi_{0.5}$~\cite{pi05}                 & 3.33\%  & 26.7$\pm$4.7\% & 1.7$\pm$2.4\% & 46.7$\pm$8.5\% & \cellcolor{mygray}25.00$\pm$19.29\% & 28.3$\pm$9.4\% & 60.0$\pm$4.1\% & 51.7$\pm$8.5\% & 13.3$\pm$8.5\% & \cellcolor{mygray}38.33$\pm$20.14\% \\

  \midrule

  \multirow{4}{*}{SONIC}      & ACT~\cite{ACT}                          & 8.57\%  & 16.7$\pm$6.2\% & 18.3$\pm$4.7\% & 56.7$\pm$6.2\% & \cellcolor{mygray}30.56$\pm$19.36\% & 78.3$\pm$9.4\% & 73.3$\pm$8.5\% & 56.7$\pm$6.2\% & 33.3$\pm$2.4\% & \underline{\cellcolor{mygray}60.42$\pm$18.98\%} \\
                              & DP~\cite{chi2023diffusionpolicy}        & 8.33\%  & 45.0$\pm$10.8\% & 36.7$\pm$4.7\% & 75.0$\pm$4.1\% & \textbf{\cellcolor{mygray}52.22$\pm$17.97\%} & 85.0$\pm$10.8\% & 78.3$\pm$14.3\% & 76.7$\pm$2.4\% & 23.3$\pm$6.2\% & \textbf{\cellcolor{mygray}65.83$\pm$26.52\%} \\
                              & FM~\cite{FM_objective}                  & 5.71\%  & 13.3$\pm$2.4\% & 38.3$\pm$4.7\% & 73.3$\pm$6.2\% & \underline{\cellcolor{mygray}41.67$\pm$25.06\%} & 70.0$\pm$4.1\% & 15.0$\pm$7.1\% & 70.0$\pm$8.2\% & 38.3$\pm$14.3\% & \cellcolor{mygray}48.33$\pm$24.94\% \\
                              & $\pi_{0.5}$~\cite{pi05}                 & 5.24\%  & 10.0$\pm$4.1\% & 43.3$\pm$6.2\% & 71.7$\pm$11.8\% & \cellcolor{mygray}41.67$\pm$26.46\% & 66.7$\pm$6.2\% & 73.3$\pm$2.4\% & 70.0$\pm$0.0\% & 23.3$\pm$6.2\% & \cellcolor{mygray}58.33$\pm$20.85\% \\

  \bottomrule
  \end{tabular}
  \end{adjustbox}
  \vspace{-10pt}
  \end{table*}

\subsection{Main Results}
\label{sec:experiments}

\textbf{\textit{In-GMT} evaluation.}
Table~\ref{table:ingmt} evaluates high-level policy baselines under matched training and inference GMTs.
Overall, the results show that existing egocentric policies can complete leg-critical HOI/HSI tasks with GMT-based execution, while performance varies substantially across policy architectures and tracker backends. Under TWIST2, FM~\cite{FM_objective} achieves the best suite averages on both HOI and HSI, reaching 36.11\% and 58.75\%, respectively. Under SONIC, DP~\cite{chi2023diffusionpolicy} becomes the strongest baseline on both suites, reaching 52.22\% on HOI and 65.83\% on HSI. Notably, $\pi_{0.5}$ achieves the best result on DoubleDesk under SONIC, reaching 43.3\% SR. This may benefit from its manipulation-oriented pretraining, since DoubleDesk involves hand-centric grasping, carrying, and placing. However, this advantage is less evident on tasks dominated by whole-body motion, where success depends on foot placement, balance, and coordinated body execution. Comparing the best results under each tracker, SONIC outperforms TWIST2 on both HOI and HSI, improving the best average SR by 16.11\% and 7.08\%, respectively. These results suggest that performance should be interpreted as a property of the policy–tracker pair rather than the high-level policy alone. Beyond SR, AFR provides an important safety-oriented view: $\pi_{0.5}$~\cite{pi05} achieves the lowest AFR under both TWIST2 and SONIC in the in-GMT setting, making it the most stable baseline. This highlights the need to evaluate both interaction success and physical stability in humanoid benchmarks.

\noindent\textcolor[HTML]{BE023F}{\textbf{\textit{Finding 1:}}} \textbf{\textit{In-GMT} performance is feasible but tracker-conditioned.}
The same policy family can behave differently when paired with different GMTs, and the best policy under one tracker is not always the best under another.
This supports the benchmark design choice of treating the general motion tracker as a first-class evaluation axis rather than a hidden low-level controller.

\begin{figure*}[h]
    \centering
    \includegraphics[width=1\linewidth]{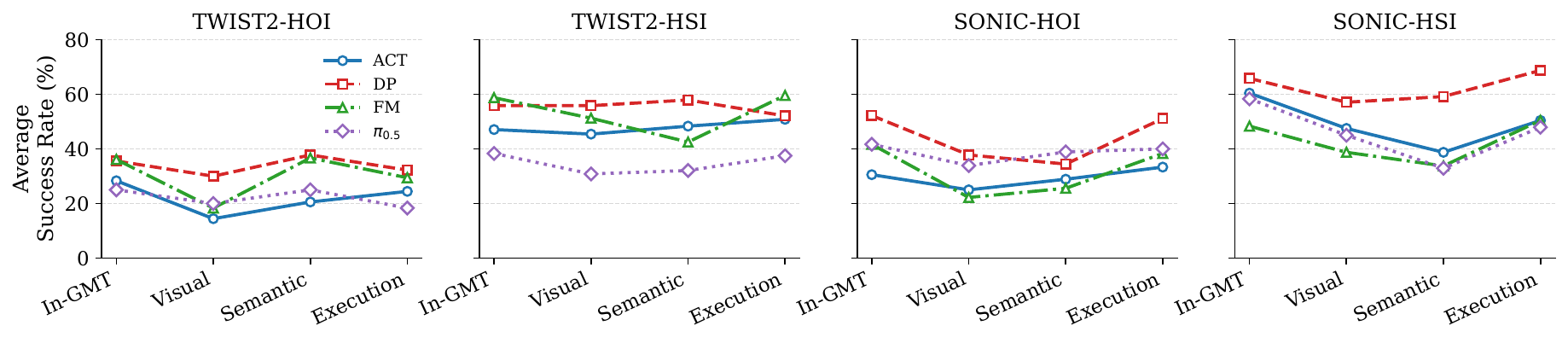}
\caption{
\textbf{Perturbation-conditioned evaluation.}
We report suite-level average success rates under the in-distribution, visual, semantic, and execution settings. Each curve denotes one high-level policy, characterizing how different policy architectures degrade under the three benchmark-defined sources of perturbation with matched GMT execution.
}
    \vspace{-5pt}
    \label{fig:perturbation}
\end{figure*}
\textbf{Perturbation-conditioned evaluation.}
Figure~\ref{fig:perturbation} summarizes \textit{in-GMT} robustness using average SR under the visual, semantic, and execution settings. Overall, SR often decreases under perturbation, especially under visual shifts, but the degradation patterns are not uniform across policies and tracker backends. Visual perturbations produce the most consistent drop, especially on HOI; for example, under TWIST2-HOI, ACT~\cite{ACT} drops from 28.33\% to 14.44\% and FM from 36.11\% to 18.33\%. DP~\cite{chi2023diffusionpolicy} shows the most consistent profile in SONIC settings, remaining strongest on HSI across all perturbation splits and on HOI except the semantic split, where $\pi_{0.5}$ achieves the highest HOI average. These results show that visual, semantic, and execution perturbations expose different failure modes rather than a single notion of robustness, motivating evaluation of egocentric appearance robustness, target grounding, and spatial adaptation of whole-body actions. Full per-task results are provided in Appendix~\ref{app:moreresults}.

\textbf{\textit{Cross-GMT} evaluation.}
Table~\ref{table:crossgmt} evaluates whether intermediate whole-body actions remain transferable when replacing the deployment GMT.
Each policy entry reports $\Delta_{\mathrm{cross}}$~($\downarrow$) / Ret~($\uparrow$), so the two metrics should be interpreted jointly. Overall, tracker replacement causes substantial degradation in both directions: TWIST2$\rightarrow$SONIC has a slightly larger average drop than SONIC$\rightarrow$TWIST2 (39.9\% vs. 36.0\%), and it produces more zero-retention entries. For example, on Football, TWIST2$\rightarrow$SONIC yields drops ranging from 25.0\% to 46.6\%, whereas SONIC$\rightarrow$TWIST2 ranges from 10.0\% to 43.3\% with low retention values of 0.0\%--12.78\%.
This shows that smaller $\Delta_{\mathrm{cross}}$ does not necessarily indicate better transfer when matched-GMT performance is already low. Furthermore, AFR provides a complementary view of tracker-induced stability.
SONIC$\rightarrow$TWIST2 has a higher average AFR than TWIST2$\rightarrow$SONIC (7.56\% vs. 1.36\%), driven by instability on DoubleDesk (32.08\%) and VisNavi (11.68\%).
In contrast, TWIST2$\rightarrow$SONIC remains stable, with its highest AFR values on DoubleDesk (2.90\%), Boxing (2.50\%), and VisNavi (2.48\%). However, the lower AFR of TWIST2$\rightarrow$SONIC does not translate into improved task transfer, as retention remains low across tasks. 

\noindent\textcolor[HTML]{BE023F}{\textbf{\textit{Finding 2:}}}
\textbf{\textit{Cross-GMT} transfer is asymmetric and metric-dependent.}
Replacing the deployment GMT causes substantial performance drops and low retention, but the two transfer directions differ in absolute drop, retention, and fall rate.
This indicates that current high-level policies do not learn fully tracker-invariant intermediate intent by default; instead, they partially specialize to the dynamics and correction behavior of the demonstration tracker.

\textbf{Qualitative Results.}
Figure~\ref{fig:qualitative} presents successful Diffusion Policy~\cite{chi2023diffusionpolicy} rollouts on Football, DoubleDesk, SitSofa, and VisNavi. 
The examples show that the learned policy can generate temporally coherent intermediate whole-body actions that are executable by the low-level general motion tracker. 
In Football, the robot coordinates approach, stance adjustment, and leg swing to kick the ball; in DoubleDesk, it moves through constrained scene geometry while maintaining stable body orientation; in SitSofa, it aligns with the sofa and lowers its body into a seated posture; and in VisNavi, it uses egocentric observations to guide stepping and turning toward the goal region. 
Together, these cases illustrate that \textsc{HumanoidArena} evaluates integrated perception, foot placement, balance, and whole-body interaction, rather than isolated manipulation or base navigation. We further provide failure case analysis in Appendix~\ref{app:failure}.

\begin{table*}[t]
  \centering
  \small
  \setlength{\tabcolsep}{3.2pt}
  \renewcommand{\arraystretch}{1.12}
  \caption{
  \textbf{\textit{Cross-GMT} evaluation.}
  T$\rightarrow$S denotes policies trained on TWIST2 demonstrations and deployed with the SONIC backend, while S$\rightarrow$T denotes the reverse direction. Each policy entry reports absolute \textit{cross-GMT} drop $\Delta_{\mathrm{cross}}$~($\downarrow$) / relative transfer retention Ret~($\uparrow$), computed with
  respect to the corresponding matched in-GMT performance.
  AFR~($\downarrow$) denotes the task-level average fall rate under each \textit{cross-GMT} deployment direction, measuring tracker-induced stability failures.
  }
  \label{table:crossgmt}
  \begin{adjustbox}{max width=\textwidth}
  \begin{tabular}{llccccc|ccccc}
  \toprule
  \multirow{2}{*}{Task Type} &
  \multirow{2}{*}{Task} &
  \multicolumn{5}{c|}{TWIST2 $\rightarrow$ SONIC} &
  \multicolumn{5}{c}{SONIC $\rightarrow$ TWIST2} \\
  \cmidrule(lr){3-7} \cmidrule(lr){8-12}
  & & ACT~\cite{ACT} & DP~\cite{chi2023diffusionpolicy} & FM~\cite{FM_objective} & $\pi_{0.5}$~\cite{pi05} & AFR~($\downarrow$)
    & ACT~\cite{ACT} & DP~\cite{chi2023diffusionpolicy} & FM~\cite{FM_objective} & $\pi_{0.5}$~\cite{pi05} & AFR~($\downarrow$) \\
  \midrule
  \multirow{3}{*}{HOI}
  & \textsc{Football}
  & 26.70 / 0.00\% & 25.00 / 46.47\% & 46.60 / 12.57\% & 26.70 / 0.00\% & \cellcolor{mygray}1.25\%
  & 16.70 / 0.00\% & 43.30 / 3.78\% & 11.60 / 12.78\% & 10.00 / 0.00\% & \cellcolor{mygray}7.08\% \\
  & \textsc{DoubleDesk}
  & 15.00 / 0.00\% & 10.00 / 0.00\% & 16.70 / 0.00\% & 1.70 / 0.00\% & \cellcolor{mygray}2.90\%
  & 18.30 / 0.00\% & 36.70 / 0.00\% & 38.30 / 0.00\% & 41.60 / 3.93\% & \cellcolor{mygray}32.08\% \\
  & \textsc{P\&Pbox}
  & 43.30 / 0.00\% & 50.00 / 0.00\% & 38.30 / 0.00\% & 46.70 / 0.00\% & \cellcolor{mygray}0.00\%
  & 31.70 / 44.09\% & 48.30 / 35.60\% & 45.00 / 38.61\% & 21.70 / 69.74\% & \cellcolor{mygray}0.40\% \\
  \midrule
  \multirow{4}{*}{HSI}
  & \textsc{OpenDoor}
  & 50.00 / 0.00\% & 68.30 / 0.00\% & 75.00 / 2.22\% & 28.30 / 0.00\% & \cellcolor{mygray}0.00\%
  & 78.30 / 0.00\% & 81.70 / 3.88\% & 63.30 / 9.57\% & 51.70 / 22.49\% & \cellcolor{mygray}0.00\% \\
  & \textsc{SitSofa}
  & 66.70 / 0.00\% & 73.30 / 0.00\% & 68.30 / 0.00\% & 60.00 / 0.00\% & \cellcolor{mygray}0.40\%
  & 45.00 / 38.61\% & 60.00 / 23.37\% & 1.00 / 93.33\% & 68.30 / 6.82\% & \cellcolor{mygray}1.65\% \\
  & \textsc{Boxing}
  & 58.30 / 0.00\% & 50.00 / 3.29\% & 55.00 / 13.11\% & 51.70 / 0.00\% & \cellcolor{mygray}2.50\%
  & 23.40 / 58.73\% & 31.70 / 58.67\% & 11.70 / 83.29\% & 15.00 / 78.57\% & \cellcolor{mygray}0.00\% \\
  & \textsc{VisNavi}
  & 13.30 / 0.00\% & 13.30 / 55.67\% & 26.70 / 0.00\% & 13.30 / 0.00\% & \cellcolor{mygray}2.48\%
  & 33.30 / 0.00\% & 23.30 / 0.00\% & 35.00 / 8.62\% & 23.30 / 0.00\% & \cellcolor{mygray}11.68\% \\
  \bottomrule
  \end{tabular}
  \end{adjustbox}
  \vspace{-10pt}
  \end{table*}
\section{Conclusion}

We presented \textsc{HumanoidArena}, a simulation-first benchmark for egocentric hierarchical whole-body humanoid learning. 
Built around a shared policy--tracker interface, \textsc{HumanoidArena} provides leg-critical HOI/HSI tasks, closed-loop multi-GMT demonstrations, and controlled evaluation protocols across visual, semantic, execution, \textit{in-GMT}, and \textit{cross-GMT} settings. 
Our experiments show that current policies can learn diverse whole-body interactions through GMT-based execution, but also reveal that performance depends strongly on the paired general motion tracker. 
In particular, cross-GMT evaluation exposes substantial and asymmetric transfer failures that are hidden under matched-backend evaluation. 
These findings highlight the need for intermediate whole-body action representations that are robust, tracker-transferable, and physically compatible with dynamic execution. 
We hope \textsc{HumanoidArena} provides a common testbed for advancing egocentric humanoid policies that coordinate perception, task interaction, balance, and whole-body motion.
\bibliographystyle{unsrt}
\bibliography{ref}

\begin{thebibliography}{10}

\bibitem{merleau1996phenomenology}
M.~Merleau-Ponty and C.~Smith.
\newblock {\em Phenomenology of Perception}.
\newblock Motilal Banarsidass Publishers (Pvt. Limited), 1996.

\bibitem{embodied1991}
Rodney~A. Brooks.
\newblock Intelligence without representation.
\newblock {\em Artificial Intelligence}, 1991.

\bibitem{stilman2008humanoid}
Mike Stilman, Koichi Nishiwaki, and Satoshi Kagami.
\newblock Humanoid teleoperation for whole body manipulation.
\newblock In {\em ICRA}, 2008.

\bibitem{borras2015wholebody}
J{\'u}lia Borr{\`a}s and Tamim Asfour.
\newblock A whole-body pose taxonomy for loco-manipulation tasks.
\newblock In {\em IROS}, 2015.

\bibitem{OmniRetarget}
Lujie Yang, Xiaoyu Huang, Zhen Wu, Angjoo Kanazawa, Pieter Abbeel, Carmelo Sferrazza, C.~Karen Liu, Rocky Duan, and Guanya Shi.
\newblock Omniretarget: Interaction-preserving data generation for humanoid whole-body loco-manipulation and scene interaction.
\newblock In {\em ICRA}, 2026.

\bibitem{kim24openvla}
{Moo Jin} Kim, Karl Pertsch, Siddharth Karamcheti, Ted Xiao, Ashwin Balakrishna, Suraj Nair, Rafael Rafailov, Ethan Foster, Grace Lam, Pannag Sanketi, Quan Vuong, Thomas Kollar, Benjamin Burchfiel, Russ Tedrake, Dorsa Sadigh, Sergey Levine, Percy Liang, and Chelsea Finn.
\newblock Openvla: An open-source vision-language-action model.
\newblock {\em arXiv preprint arXiv:2406.09246}, 2024.

\bibitem{pi05}
Physical Intelligence, Kevin Black, Noah Brown, James Darpinian, Karan Dhabalia, Danny Driess, Adnan Esmail, Michael Equi, Chelsea Finn, Niccolo Fusai, Manuel~Y. Galliker, Dibya Ghosh, Lachy Groom, Karol Hausman, Brian Ichter, Szymon Jakubczak, Tim Jones, Liyiming Ke, Devin LeBlanc, Sergey Levine, Adrian Li-Bell, Mohith Mothukuri, Suraj Nair, Karl Pertsch, Allen~Z. Ren, Lucy~Xiaoyang Shi, Laura Smith, Jost~Tobias Springenberg, Kyle Stachowicz, James Tanner, Quan Vuong, Homer Walke, Anna Walling, Haohuan Wang, Lili Yu, and Ury Zhilinsky.
\newblock $\pi_{0.5}$: a vision-language-action model with open-world generalization.
\newblock {\em arXiv preprint arXiv:2504.16054}, 2025.

\bibitem{beyondmimic}
Qiayuan Liao, Takara~E. Truong, Xiaoyu Huang, Yuman Gao, Guy Tevet, Koushil Sreenath, and C.~Karen Liu.
\newblock Beyondmimic: From motion tracking to versatile humanoid control via guided diffusion.
\newblock {\em arXiv preprint arXiv:2508.08241}, 2025.

\bibitem{konggubot}
Weiji Xie, Jinrui Han, Jiakun Zheng, Huanyu Li, Xinzhe Liu, Jiyuan Shi, Weinan Zhang, Chenjia Bai, and Xuelong Li.
\newblock Kungfubot: Physics-based humanoid whole-body control for learning highly-dynamic skills.
\newblock {\em arXiv preprint arXiv:2506.12851}, 2025.

\bibitem{wang2026omnixtreme}
Yunshen Wang, Shaohang Zhu, Peiyuan Zhi, Yuhan Li, Jiaxin Li, Yong-Lu Li, Yuchen Xiao, Xingxing Wang, Baoxiong Jia, and Siyuan Huang.
\newblock Omnixtreme: Breaking the generality barrier in high-dynamic humanoid control.
\newblock {\em arXiv preprint arXiv:2602.23843}, 2026.

\bibitem{ze2025twist2}
Yanjie Ze, Siheng Zhao, Weizhuo Wang, Angjoo Kanazawa, Rocky Duan, Pieter Abbeel, Guanya Shi, Jiajun Wu, and C.~Karen Liu.
\newblock Twist2: Scalable, portable, and holistic humanoid data collection system.
\newblock {\em arXiv preprint arXiv:2511.02832}, 2025.

\bibitem{luo2025sonic}
Zhengyi Luo, Ye~Yuan, Tingwu Wang, Chenran Li, Sirui Chen, Fernando Casta\~neda, Zi-Ang Cao, Jiefeng Li, David Minor, Qingwei Ben, Xingye Da, Runyu Ding, Cyrus Hogg, Lina Song, Edy Lim, Eugene Jeong, Tairan He, Haoru Xue, Wenli Xiao, Zi~Wang, Simon Yuen, Jan Kautz, Yan Chang, Umar Iqbal, Linxi Fan, and Yuke Zhu.
\newblock Sonic: Supersizing motion tracking for natural humanoid whole-body control.
\newblock {\em arXiv preprint arXiv:2511.07820}, 2025.

\bibitem{HumanoidBench}
Carmelo Sferrazza, Dun-Ming Huang, Xingyu Lin, Youngwoon Lee, and Pieter Abbeel.
\newblock Humanoidbench: Simulated humanoid benchmark for whole-body locomotion and manipulation.
\newblock {\em arXiv preprint arXiv:2403.10506}, 2024.

\bibitem{leverb}
Haoru Xue, Xiaoyu Huang, Dantong Niu, Qiayuan Liao, Thomas Kragerud, Jan~Tommy Gravdahl, Xue~Bin Peng, Guanya Shi, Trevor Darrell, Koushil Sreenath, and Shankar Sastry.
\newblock Leverb: Humanoid whole-body control with latent vision-language instruction.
\newblock {\em arXiv preprint arXiv:2506.13751}, 2025.

\bibitem{li2022behavior}
Chengshu Li, Ruohan Zhang, Josiah Wong, Cem Gokmen, Sanjana Srivastava, Roberto Mart{\'\i}n-Mart{\'\i}n, Chen Wang, Gabrael Levine, Michael Lingelbach, Jiankai Sun, Mona Anvari, Minjune Hwang, Manasi Sharma, Arman Aydin, Dhruva Bansal, Samuel Hunter, Kyu-Young Kim, Alan Lou, Caleb~R Matthews, Ivan Villa-Renteria, Jerry~Huayang Tang, Claire Tang, Fei Xia, Silvio Savarese, Hyowon Gweon, Karen Liu, Jiajun Wu, and Li~Fei-Fei.
\newblock {BEHAVIOR}-1k: A benchmark for embodied {AI} with 1,000 everyday activities and realistic simulation.
\newblock In {\em CoRL}, 2022.

\bibitem{liu2023libero}
Bo~Liu, Yifeng Zhu, Chongkai Gao, Yihao Feng, Qiang Liu, Yuke Zhu, and Peter Stone.
\newblock Libero: Benchmarking knowledge transfer for lifelong robot learning.
\newblock {\em arXiv preprint arXiv:2306.03310}, 2023.

\bibitem{HumanoidVerse}
Haozhuo Zhang, Jingkai Sun, Michele Caprio, Jian Tang, Shanghang Zhang, Qiang Zhang, and Wei Pan.
\newblock Humanoidverse: A versatile humanoid for vision-language guided multi-object rearrangement.
\newblock {\em arXiv preprint arXiv:2508.16943v1}, 2025.

\bibitem{hdmi}
Haoyang Weng, Yitang Li, Nikhil Sobanbabu, Zihan Wang, Zhengyi Luo, Tairan He, Deva Ramanan, and Guanya Shi.
\newblock Hdmi: Learning interactive humanoid whole-body control from human videos.
\newblock {\em arXiv preprint arXiv:2509.16757}, 2025.

\bibitem{zhao2025resmimic}
Siheng Zhao, Yanjie Ze, Yue Wang, C.~Karen Liu, Pieter Abbeel, Guanya Shi, and Rocky Duan.
\newblock Resmimic: From general motion tracking to humanoid whole-body loco-manipulation via residual learning.
\newblock {\em arXiv preprint arXiv:2510.05070}, 2025.

\bibitem{li2026haic}
Dongting Li, Xingyu Chen, Qianyang Wu, Bo~Chen, Sikai Wu, Hanyu Wu, Guoyao Zhang, Liang Li, Mingliang Zhou, Diyun Xiang, et~al.
\newblock Haic: Humanoid agile object interaction control via dynamics-aware world model.
\newblock {\em arXiv preprint arXiv:2602.11758}, 2026.

\bibitem{HumanX}
Yinhuai Wang, Qihan Zhao, Yuen~Fui Lau, Runyi Yu, Hok~Wai Tsui, Qifeng Chen, Jingbo Wang, Jiangmiao Pang, and Ping Tan.
\newblock Humanx: Toward agile and generalizable humanoid interaction skills from human videos.
\newblock {\em arXiv preprint arXiv:2602.02473}, 2026.

\bibitem{su2025hitter}
Zhi Su, Bike Zhang, Nima Rahmanian, Yuman Gao, Qiayuan Liao, Caitlin Regan, Koushil Sreenath, and S~Shankar Sastry.
\newblock Hitter: A humanoid table tennis robot via hierarchical planning and learning.
\newblock {\em arXiv preprint arXiv:2508.21043}, 2025.

\bibitem{zhang2026learning}
Zhikai Zhang, Haofei Lu, Yunrui Lian, Ziqing Chen, Yun Liu, Chenghuai Lin, Han Xue, Zicheng Zeng, Zekun Qi, Shaolin Zheng, et~al.
\newblock Learning athletic humanoid tennis skills from imperfect human motion data.
\newblock {\em arXiv preprint arXiv:2603.12686}, 2026.

\bibitem{wang2025physhsi}
Huayi Wang, Wentao Zhang, Runyi Yu, Tao Huang, Junli Ren, Feiyu Jia, Zirui Wang, Xiaojie Niu, Xiao Chen, Jiahe Chen, et~al.
\newblock Physhsi: Towards a real-world generalizable and natural humanoid-scene interaction system.
\newblock {\em arXiv preprint arXiv:2510.11072}, 2025.

\bibitem{yin2025visualmimic}
Shaofeng Yin, Yanjie Ze, Hong-Xing Yu, C~Karen Liu, and Jiajun Wu.
\newblock Visualmimic: Visual humanoid loco-manipulation via motion tracking and generation.
\newblock {\em arXiv preprint arXiv:2509.20322}, 2025.

\bibitem{he2025viral}
Tairan He, Zi~Wang, Haoru Xue, Qingwei Ben, Zhengyi Luo, Wenli Xiao, Ye~Yuan, Xingye Da, Fernando Casta{\~n}eda, Shankar Sastry, et~al.
\newblock Viral: Visual sim-to-real at scale for humanoid loco-manipulation.
\newblock {\em arXiv preprint arXiv:2511.15200}, 2025.

\bibitem{lin2026prohoi}
Yuhang Lin, Jiyuan Shi, Dewei Wang, Jipeng Kong, Yong Liu, Chenjia Bai, and Xuelong Li.
\newblock Pro-hoi: Perceptive root-guided humanoid-object interaction.
\newblock {\em arXiv preprint arXiv:2603.01126}, 2026.

\bibitem{lin2026lessmimic}
Yutang Lin, Jieming Cui, Yixuan Li, Baoxiong Jia, Yixin Zhu, and Siyuan Huang.
\newblock Lessmimic: Long-horizon humanoid interaction with unified distance field representations.
\newblock {\em arXiv preprint arXiv:2602.21723}, 2026.

\bibitem{he2026ultra}
Xialin He, Sirui Xu, Xinyao Li, Runpei Dong, Liuyu Bian, Yu-Xiong Wang, and Liang-Yan Gui.
\newblock Ultra: Unified multimodal control for autonomous humanoid whole-body loco-manipulation.
\newblock {\em arXiv preprint arXiv:2603.03279}, 2026.

\bibitem{ren2026smash}
Junli Ren, Yinghui Li, Kai Zhang, Penglin Fu, Haoran Jiang, Yixuan Pan, Guangjun Zeng, Tao Huang, Weizhong Guo, Peng Lu, et~al.
\newblock Smash: Mastering scalable whole-body skills for humanoid ping-pong with egocentric vision.
\newblock {\em arXiv preprint arXiv:2604.01158}, 2026.

\bibitem{rudin2025parkour}
Nikita Rudin, Junzhe He, Joshua Aurand, and Marco Hutter.
\newblock Parkour in the wild: Learning a general and extensible agile locomotion policy using multi-expert distillation and rl fine-tuning.
\newblock {\em arXiv preprint arXiv:2505.11164}, 2025.

\bibitem{liu2025ego}
Hang Liu, Yuman Gao, Sangli Teng, Yufeng Chi, Yakun~Sophia Shao, Zhongyu Li, Maani Ghaffari, and Koushil Sreenath.
\newblock Ego-vision world model for humanoid contact planning.
\newblock {\em arXiv preprint arXiv:2510.11682}, 2025.

\bibitem{ben2025gallant}
Qingwei Ben, Botian Xu, Kailin Li, Feiyu Jia, Wentao Zhang, Jingping Wang, Jingbo Wang, Dahua Lin, and Jiangmiao Pang.
\newblock Gallant: Voxel grid-based humanoid locomotion and local-navigation across 3d constrained terrains.
\newblock {\em arXiv preprint arXiv:2511.14625}, 2025.

\bibitem{zhu2026hiking}
Shaoting Zhu, Ziwen Zhuang, Mengjie Zhao, Kun-Ying Lee, and Hang Zhao.
\newblock Hiking in the wild: A scalable perceptive parkour framework for humanoids.
\newblock {\em arXiv preprint arXiv:2601.07718}, 2026.

\bibitem{zhuang2026deep}
Ziwen Zhuang, Shaoting Zhu, Mengjie Zhao, and Hang Zhao.
\newblock Deep whole-body parkour.
\newblock {\em arXiv preprint arXiv:2601.07701}, 2026.

\bibitem{wu2026perceptive}
Zhen Wu, Xiaoyu Huang, Lujie Yang, Yuanhang Zhang, Koushil Sreenath, Xi~Chen, Pieter Abbeel, Rocky Duan, Angjoo Kanazawa, Carmelo Sferrazza, et~al.
\newblock Perceptive humanoid parkour: Chaining dynamic human skills via motion matching.
\newblock {\em arXiv preprint arXiv:2602.15827}, 2026.

\bibitem{zerowbc}
Haoran Yang, Jiacheng Bao, Yucheng Xin, Haoming Song, Yuyang Tian, Bin Zhao, Dong Wang, and Xuelong Li.
\newblock Zerowbc: Learning natural visuomotor humanoid control directly from human egocentric video.
\newblock {\em arXiv preprint arXiv:2603.09170}, 2026.

\bibitem{PhysiFlow}
Weikai Qin, Sichen Wu, Ci~Chen, Mengfan Liu, Linxi Feng, Xinru Cui, Haoqi Han, and Hesheng Wang.
\newblock Physiflow: Physics-aware humanoid whole-body vla via multi-brain latent flow matching and robust tracking.
\newblock {\em arXiv preprint arXiv:2603.05410}, 2026.

\bibitem{HumanoidVLA}
Pengxiang Ding, Jianfei Ma, Xinyang Tong, Binghong Zou, Xinxin Luo, Yiguo Fan, Ting Wang, Hongchao Lu, Panzhong Mo, Jinxin Liu, Yuefan Wang, Huaicheng Zhou, Wenshuo Feng, Jiacheng Liu, Siteng Huang, and Donglin Wang.
\newblock Humanoid-vla: Towards universal humanoid control with visual integration.
\newblock {\em arXiv preprint arXiv:2502.14795}, 2025.

\bibitem{jiang2025wholebodyvla}
Haoran Jiang, Jin Chen, Qingwen Bu, Li~Chen, Modi Shi, Yanjie Zhang, Delong Li, Chuanzhe Suo, Chuang Wang, Zhihui Peng, and Hongyang Li.
\newblock Wholebodyvla: Towards unified latent vla for whole-body loco-manipulation control.
\newblock {\em arXiv preprint arXiv:2512.11047}, 2025.

\bibitem{Psi0}
Songlin Wei, Hongyi Jing, Boqian Li, Zhenyu Zhao, Jiageng Mao, Zhenhao Ni, Sicheng He, Jie Liu, Xiawei Liu, Kaidi Kang, Sheng Zang, Weiduo Yuan, Marco Pavone, Di~Huang, and Yue Wang.
\newblock $\psi_0$: An open foundation model towards universal humanoid loco-manipulation.
\newblock {\em arXiv preprint arXiv:2603.12263}, 2026.

\bibitem{jiang2025brs}
Yunfan Jiang, Ruohan Zhang, Josiah Wong, Chen Wang, Yanjie Ze, Hang Yin, Cem Gokmen, Shuran Song, Jiajun Wu, and Li~Fei-Fei.
\newblock {BEHAVIOR} robot suite: Streamlining real-world whole-body manipulation for everyday household activities.
\newblock In {\em CoRL}, 2025.

\bibitem{zhang2025agentworld}
Yizheng Zhang, Zhenjun Yu, Jiaxin Lai, Cewu Lu, and Lei Han.
\newblock Agentworld: An interactive simulation platform for scene construction and mobile robotic manipulation.
\newblock In {\em CoRL}, 2025.

\bibitem{zhao2025humanoideveryday}
Zhenyu Zhao, Hongyi Jing, Xiawei Liu, Jiageng Mao, Abha Jha, Hanwen Yang, Rong Xue, Sergey Zakharov, Vitor Guizilini, and Yue Wang.
\newblock Humanoid everyday: A comprehensive robotic dataset for open-world humanoid manipulation.
\newblock {\em arXiv preprint arXiv:2510.08807}, 2025.

\bibitem{ACT}
Tony~Z. Zhao, Vikash Kumar, Sergey Levine, and Chelsea Finn.
\newblock Learning fine-grained bimanual manipulation with low-cost hardware.
\newblock In {\em RSS}, 2023.

\bibitem{chi2023diffusionpolicy}
Cheng Chi, Siyuan Feng, Yilun Du, Zhenjia Xu, Eric Cousineau, Benjamin Burchfiel, and Shuran Song.
\newblock Diffusion policy: Visuomotor policy learning via action diffusion.
\newblock In {\em RSS}, 2023.

\bibitem{FM_objective}
Boston Dynamics and TRI~Research Team.
\newblock Large behavior models and atlas find new footing, 2025.
\newblock Blog post.

\bibitem{mittal2025isaaclab}
Mayank Mittal, Pascal Roth, James Tigue, Antoine Richard, Octi Zhang, Peter Du, Antonio Serrano-Muñoz, Xinjie Yao, René Zurbrügg, Nikita Rudin, Lukasz Wawrzyniak, Milad Rakhsha, Alain Denzler, Eric Heiden, Ales Borovicka, Ossama Ahmed, Iretiayo Akinola, Abrar Anwar, Mark~T. Carlson, Ji~Yuan Feng, Animesh Garg, Renato Gasoto, Lionel Gulich, Yijie Guo, M.~Gussert, Alex Hansen, Mihir Kulkarni, Chenran Li, Wei Liu, Viktor Makoviychuk, Grzegorz Malczyk, Hammad Mazhar, Masoud Moghani, Adithyavairavan Murali, Michael Noseworthy, Alexander Poddubny, Nathan Ratliff, Welf Rehberg, Clemens Schwarke, Ritvik Singh, James~Latham Smith, Bingjie Tang, Ruchik Thaker, Matthew Trepte, Karl~Van Wyk, Fangzhou Yu, Alex Millane, Vikram Ramasamy, Remo Steiner, Sangeeta Subramanian, Clemens Volk, CY~Chen, Neel Jawale, Ashwin~Varghese Kuruttukulam, Michael~A. Lin, Ajay Mandlekar, Karsten Patzwaldt, John Welsh, Huihua Zhao, Fatima Anes, Jean-Francois Lafleche, Nicolas Moënne-Loccoz, Soowan Park, Rob Stepinski, Dirk~Van Gelder,
  Chris Amevor, Jan Carius, Jumyung Chang, Anka~He Chen, Pablo de~Heras~Ciechomski, Gilles Daviet, Mohammad Mohajerani, Julia von Muralt, Viktor Reutskyy, Michael Sauter, Simon Schirm, Eric~L. Shi, Pierre Terdiman, Kenny Vilella, Tobias Widmer, Gordon Yeoman, Tiffany Chen, Sergey Grizan, Cathy Li, Lotus Li, Connor Smith, Rafael Wiltz, Kostas Alexis, Yan Chang, David Chu, Linxi~"Jim" Fan, Farbod Farshidian, Ankur Handa, Spencer Huang, Marco Hutter, Yashraj Narang, Soha Pouya, Shiwei Sheng, Yuke Zhu, Miles Macklin, Adam Moravanszky, Philipp Reist, Yunrong Guo, David Hoeller, and Gavriel State.
\newblock Isaac lab: A gpu-accelerated simulation framework for multi-modal robot learning.
\newblock {\em arXiv preprint arXiv:2511.04831}, 2025.

\bibitem{joao2025gmr}
Joao~Pedro Araujo, Yanjie Ze, Pei Xu, Jiajun Wu, and C.~Karen Liu.
\newblock Retargeting matters: General motion retargeting for humanoid motion tracking.
\newblock {\em arXiv preprint arXiv:2510.02252}, 2025.

\bibitem{ze2025twist}
Yanjie Ze, Zixuan Chen, João~Pedro Araújo, Zi~ang Cao, Xue~Bin Peng, Jiajun Wu, and C.~Karen Liu.
\newblock Twist: Teleoperated whole-body imitation system.
\newblock {\em arXiv preprint arXiv:2505.02833}, 2025.

\end{thebibliography}

%%%%%%%%%%%%%%%%%%%%%%%%%%%%%%%%%%%%%%%%%%%%%%%%%%%%%%%%%%%%

\appendix

\clearpage
\setcounter{page}{1}

\newcolumntype{C}[1]{>{\centering\arraybackslash}p{#1}}
\newcolumntype{L}[1]{>{\raggedright\arraybackslash}p{#1}}
\renewcommand{\thesection}{S\arabic{section}}
\renewcommand{\thetable}{S\arabic{table}}
\renewcommand{\thefigure}{S\arabic{figure}}
\setcounter{table}{0}
\setcounter{figure}{0}
\setcounter{section}{0}
\centerline{\textbf{SUMMARY OF THE APPENDIX}} 
\vspace{0.5em}

This appendix contains additional experimental results and discussions of our work, organized as:
\begin{itemize}[leftmargin=*]
\setlength{\itemsep}{0pt}
  \item \S\ref{app:tasks} details the \textbf{leg-critical HOI/HSI task suite}, including task layouts, initialization ranges, success conditions, termination criteria, and why each task requires lower-body participation.
  \item \S\ref{app:data_spec} describes the \textbf{data collection pipeline and dataset specification}, including VR teleoperation, GMR retargeting, synchronized observations, episode format, data fields, train/test splits, and multi-GMT demonstration statistics.
  \item \S\ref{app:action_gmt} explains the \textbf{intermediate whole-body action interface and GMT adapters}, covering the 40D canonical action, 64D proprioceptive state, tracker-specific mappings, and the SONIC/TWIST2 execution interfaces.
  \item \S\ref{app:eval_protocol} outlines the \textbf{evaluation protocol details}, including visual, semantic, and execution perturbation configurations, held-out splits, rollout seeds, \textit{in/cross-GMT} settings, and metric definitions.
  \item \S\ref{app:qualitative} provides \textbf{additional experimental analysis}, examining representative rollout behaviors across diverse HOI/HSI tasks.
  \item \S\ref{app:implementation} provides \textbf{policy implementation and training details}, including ACT, Diffusion Policy, Flow Matching, and $\pi_{0.5}$ hyperparameters, learning rates, batch sizes, action chunk lengths, inference horizons, preprocessing, training steps, and compute resources.
  \item \S\ref{app:failure} analyses \textbf{failure cases}, separating perception failures, semantic grounding failures, execution instability, fall cases, tracker incompatibility, and compounding errors under long-horizon whole-body interaction.
  \item \S\ref{app:release} provides \textbf{reproducibility and release details}, including code structure, dataset format, environment setup, asset dependencies, random seeds, license information, and expected compute requirements.
  \item \S\ref{app:moreresults} presents \textbf{additional results}, including full per-task perturbation results for in-GMT evaluation and merge-all training analysis across TWIST2, SONIC, and combined GMT
  datasets.
  \item \S\ref{app:discussion} discusses \textbf{limitations and future directions}.
\end{itemize}

\appendix

\section{Task Suite Specification}
\label{app:tasks}

\newcommand{\taskrolloutimg}[1]{%
    \begin{minipage}[t]{0.19\textwidth}
        \centering
        \includegraphics[width=\linewidth]{#1}
    \end{minipage}%
}

This section specifies the seven leg-critical egocentric tasks in \textsc{HumanoidArena}, designed to evaluate whole-body robotic capabilities that require tight coordination between locomotion, perception, balance control, and task-oriented interaction. The task suite consists of three Human-Object Interaction (HOI) tasks and four Human-Scene Interaction (HSI) tasks. The HOI tasks include \textbf{Football}, where the robot kicks a soccer ball into a goal net; \textbf{DoubleDesk}, where it transfers a hammer from one desk to a basket on another desk; and \textbf{P\&PBox}, where it picks up a box from a table and places it onto a shelf. The HSI tasks include \textbf{OpenDoor}, which requires opening an articulated door and passing through the doorway; \textbf{SitSofa}, which requires navigating to and sitting on a sofa; \textbf{Boxing}, which requires striking a randomized target mounted on a punching bag; and \textbf{VisNavi}, which requires visually navigating through obstacles and stepping into a target sign zone. Together, these tasks cover a diverse set of lower-body-dependent skills, including approach-and-kick motion, object transport during walking, articulated contact interaction, obstacle-aware navigation, precise foot placement, whole-body posture adjustment, and controlled sitting or striking behaviors. The task-specific success criteria, failure/termination conditions, and maximum episode lengths are summarized in Table~\ref{tab:termination}. Each task is evaluated under a consistent protocol with 3 random seeds and 20 repeated trials per seed, resulting in 60 episodes for each model configuration.

\begin{center}
\captionof{table}{Task-specific success and termination conditions. Fall detection uses a 60\textdegree\ torso tilt threshold, confirmed over 5 consecutive steps; an additional 75\textdegree\ hard-tilt condition or 50N body contact force with tilt triggers immediate termination. All tasks share a standing criterion of root height $\geq$ 0.45m and torso up-axis alignment $\geq$ 0.60.}
\label{tab:termination}
\footnotesize
\setlength{\tabcolsep}{2pt}
\setlength{\extrarowheight}{2pt}
\renewcommand{\arraystretch}{1.15}
\resizebox{0.8\textwidth}{!}{%
\begin{tabular}{@{}C{1.50cm}C{1.20cm}C{0.55cm}C{1.00cm}L{3.0cm}L{3.50cm}@{}}
\toprule
\rowcolor{gray!15}
& 
& \multicolumn{3}{c}{\textbf{Termination}} 
& \\
\rowcolor{gray!15}
\multirow{-2}{*}{\cellcolor{gray!15}\textbf{Task}} 
& \multirow{-2}{*}{\cellcolor{gray!15}\textbf{Max Steps}} 
& \textbf{Fall} 
& \textbf{Timeout} 
& \multicolumn{1}{c}{\textbf{Note}} 
& \multirow{-2}{*}{\cellcolor{gray!15}\textbf{Success Condition}} \\
\midrule
\textsc{Football} 
& 1300 
& \checkmark 
& \checkmark 
& Ball leaves the field 
& Ball center crosses the goal plane and remains within the net bounds \\
\textsc{DoubleDesk} 
& 2000 
& \checkmark 
& \checkmark 
& -- 
& Hammer center of mass lies inside the basket bounding box \\
\textsc{P\&PBox} 
& 2000 
& \checkmark 
& \checkmark 
& Non-terminal box drop
& Box rests on the shelf support surface, with its center within the support span \\
\textsc{OpenDoor} 
& 1800 
& \checkmark 
& \checkmark 
& -- 
& Robot root passes through the door-frame plane while remaining upright \\
\textsc{SitSofa} 
& 1050 
& \checkmark 
& \checkmark 
& -- 
& Pelvis or hip bodies maintain stable contact with the sofa seat surface \\
\textsc{Boxing} 
& 1500 
& \checkmark 
& \checkmark 
& -- 
& Hand reaches the dynamic hit radius of the target \\
\textsc{VisNavi} 
& 1800 
& \checkmark 
& \checkmark 
& -- 
& Both feet are inside the target bounding box while the robot remains upright \\
\bottomrule
\end{tabular}%
}
\end{center}

\begin{center}
    \taskrolloutimg{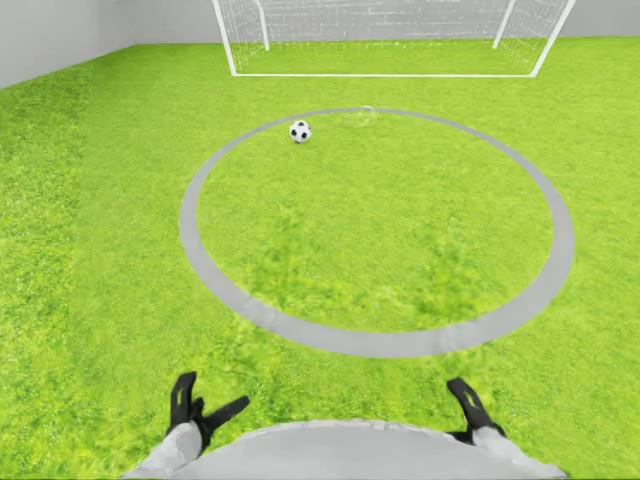}\hfill
    \taskrolloutimg{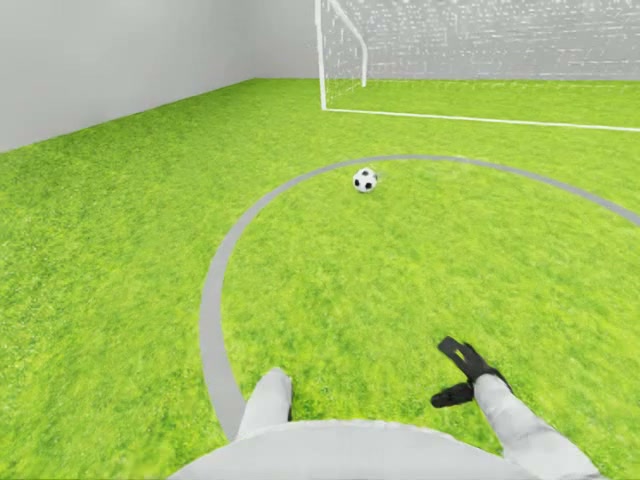}\hfill
    \taskrolloutimg{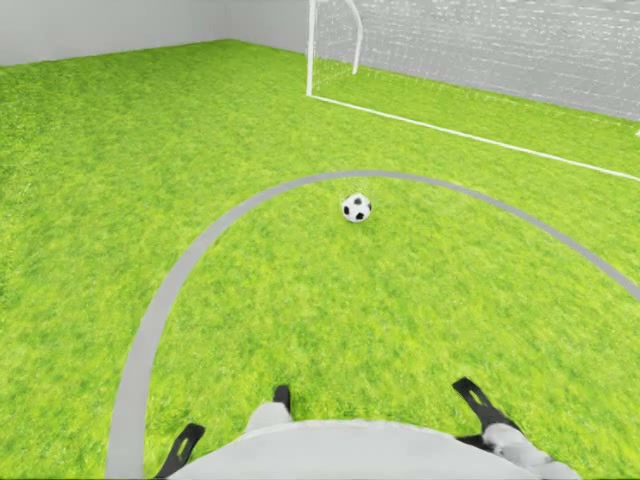}\hfill
    \taskrolloutimg{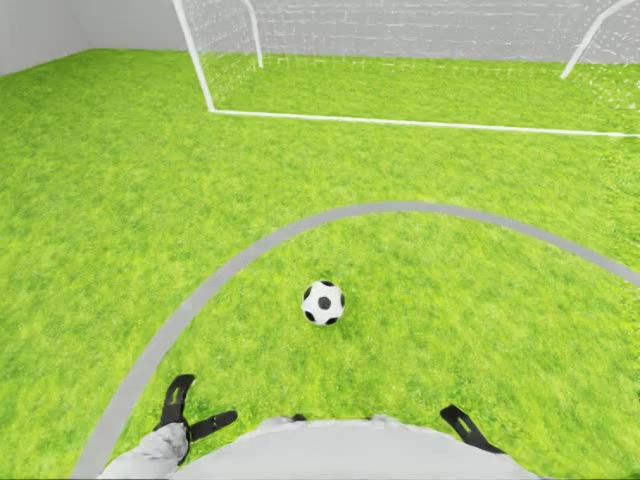}\hfill
    \taskrolloutimg{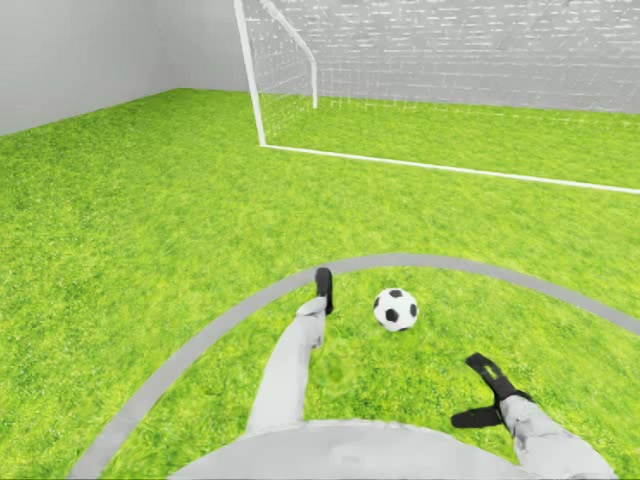}

    \vspace{0.35em}

    \taskrolloutimg{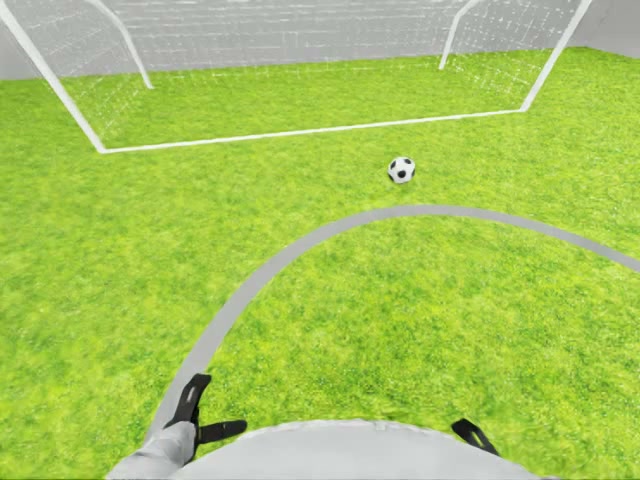}\hfill
    \taskrolloutimg{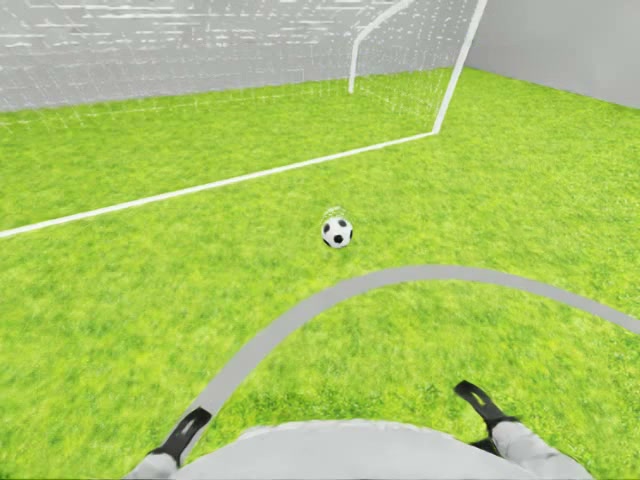}\hfill
    \taskrolloutimg{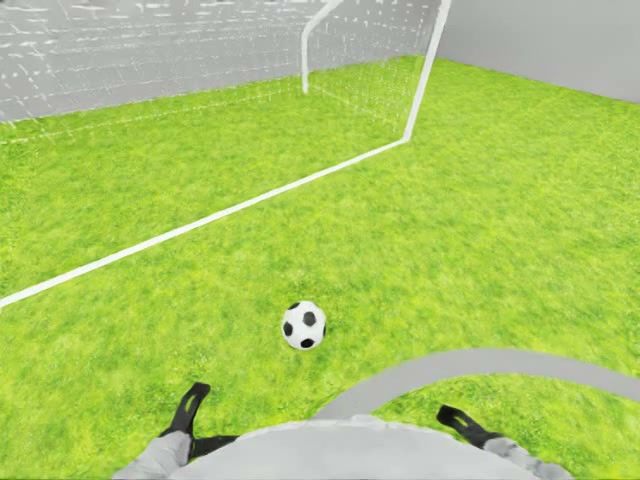}\hfill
    \taskrolloutimg{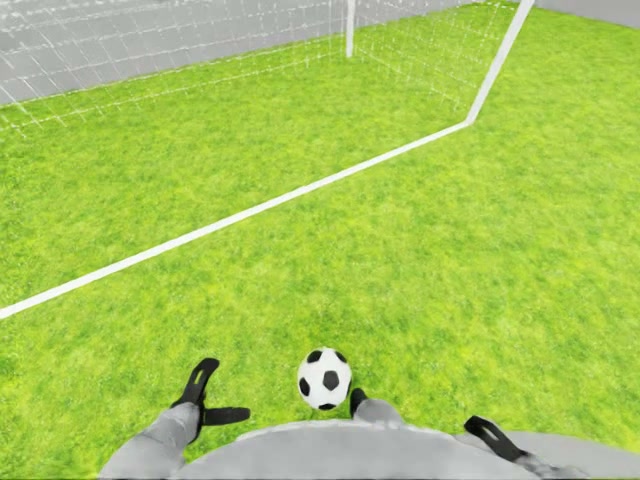}\hfill
    \taskrolloutimg{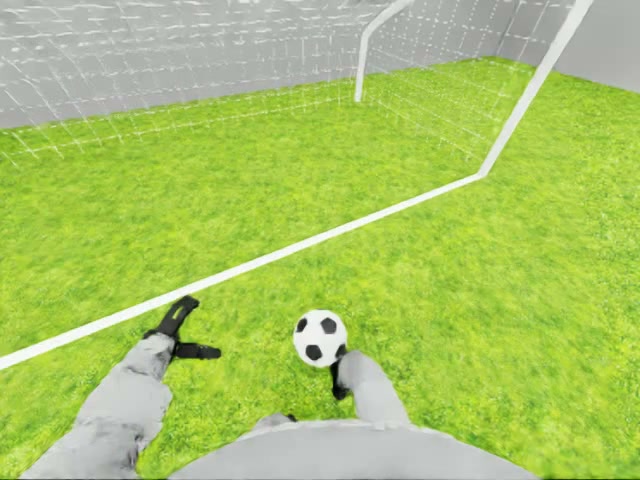}
    \captionof{figure}{Football successful rollout with 10 uniformly sampled frames shown in temporal order from left to right, top to bottom.}
    \label{fig:app_football_success_rollout}
\end{center}

\begin{center}
    \taskrolloutimg{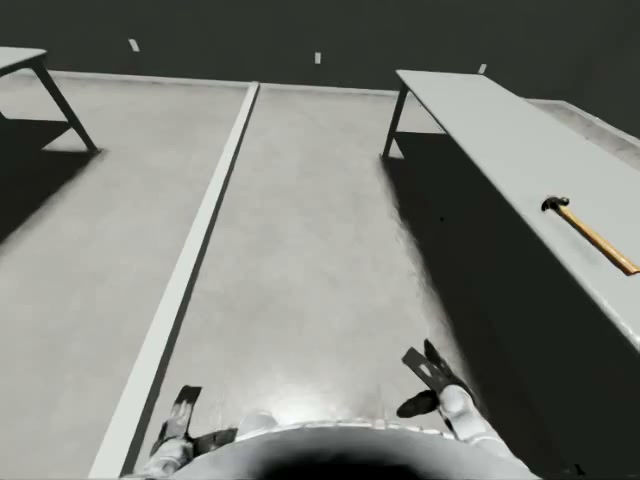}\hfill
    \taskrolloutimg{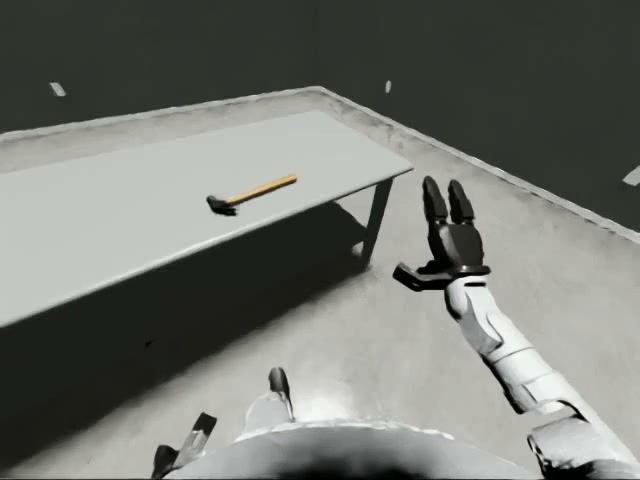}\hfill
    \taskrolloutimg{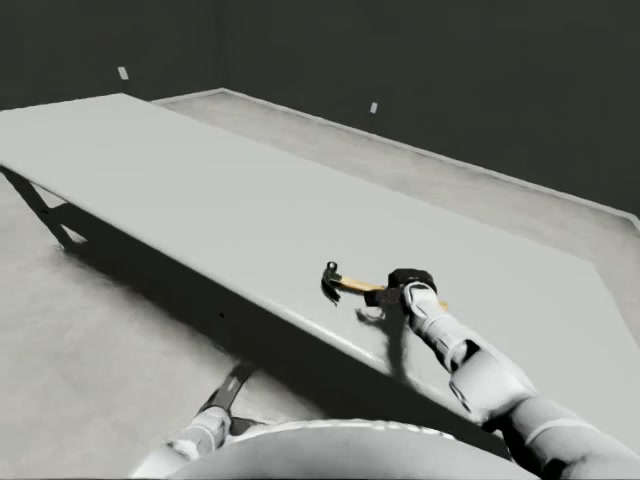}\hfill
    \taskrolloutimg{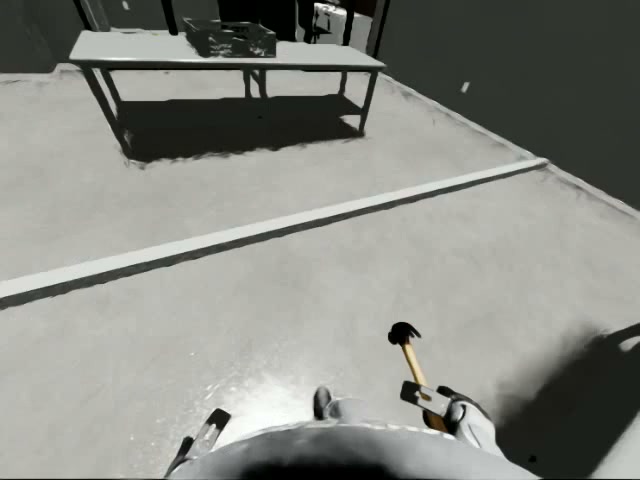}\hfill
    \taskrolloutimg{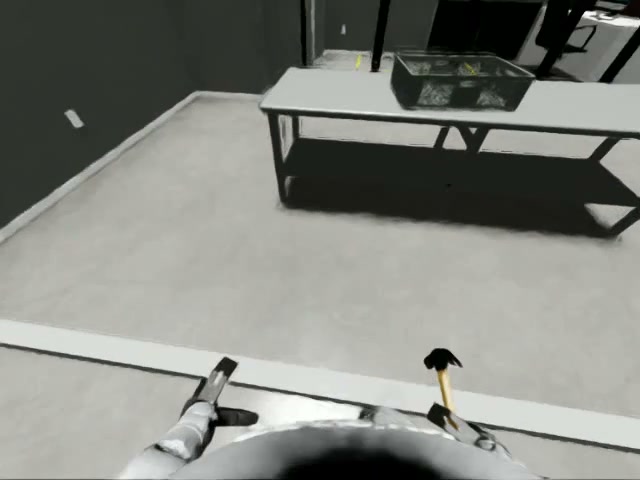}

    \vspace{0.35em}

    \taskrolloutimg{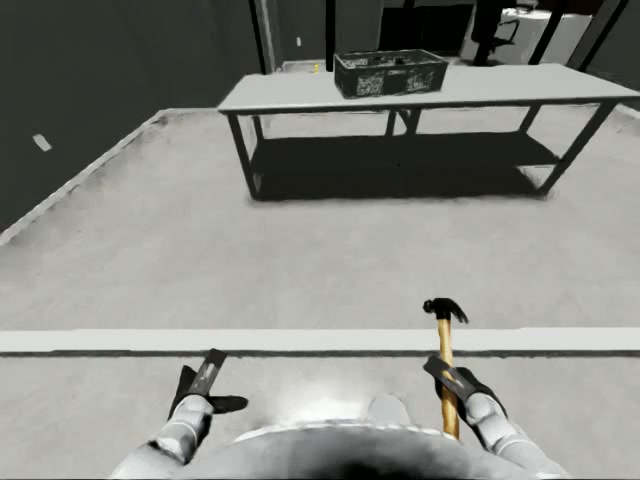}\hfill
    \taskrolloutimg{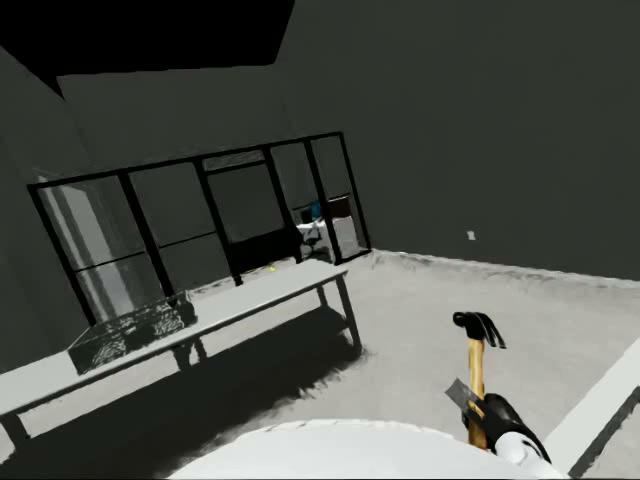}\hfill
    \taskrolloutimg{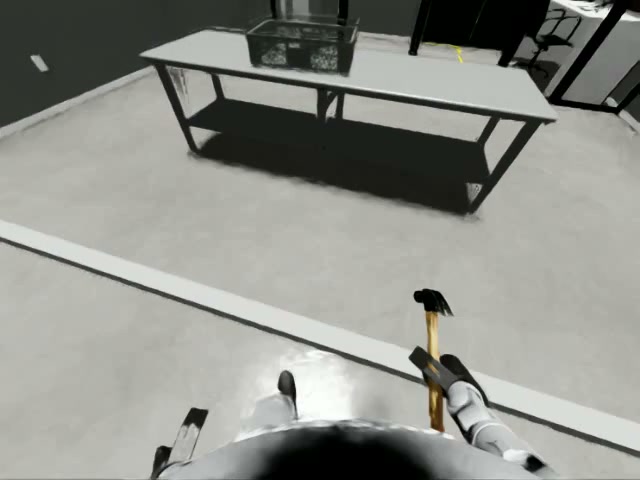}\hfill
    \taskrolloutimg{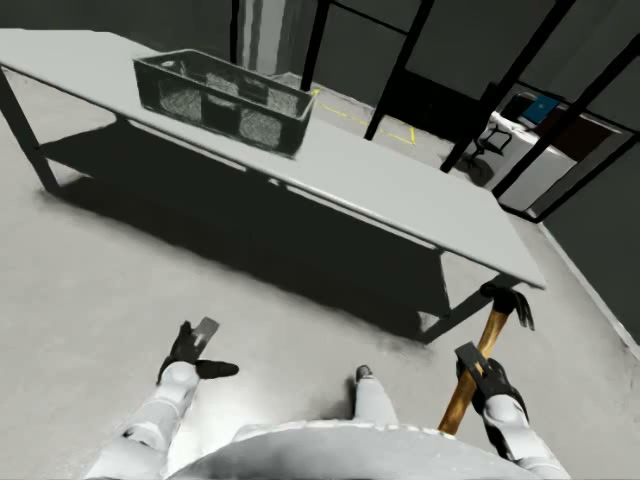}\hfill
    \taskrolloutimg{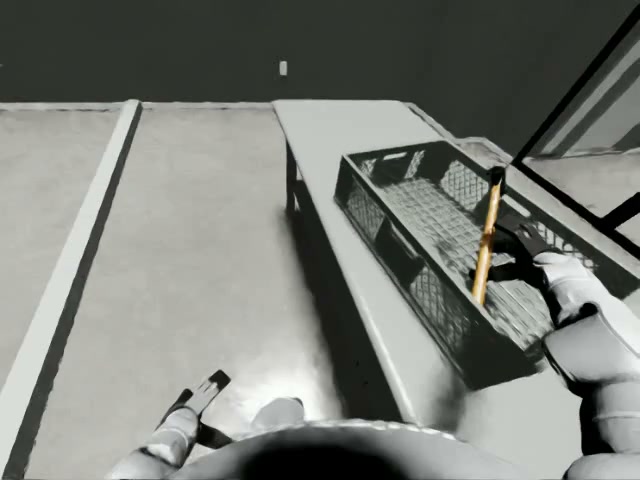}
    \captionof{figure}{DoubleDesk successful rollout with 10 uniformly sampled frames shown in temporal order from left to right, top to bottom.}
    \label{fig:app_doubledesk_success_rollout}
\end{center}

\begin{center}
    \taskrolloutimg{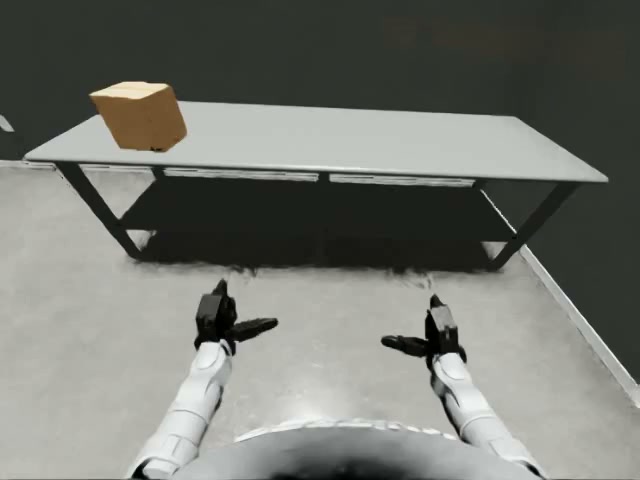}\hfill
    \taskrolloutimg{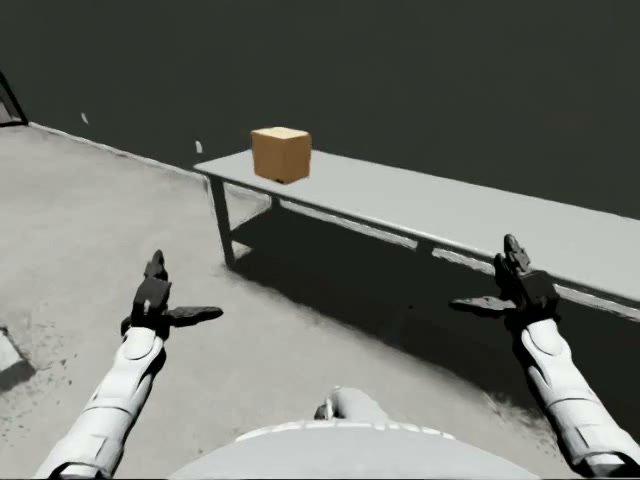}\hfill
    \taskrolloutimg{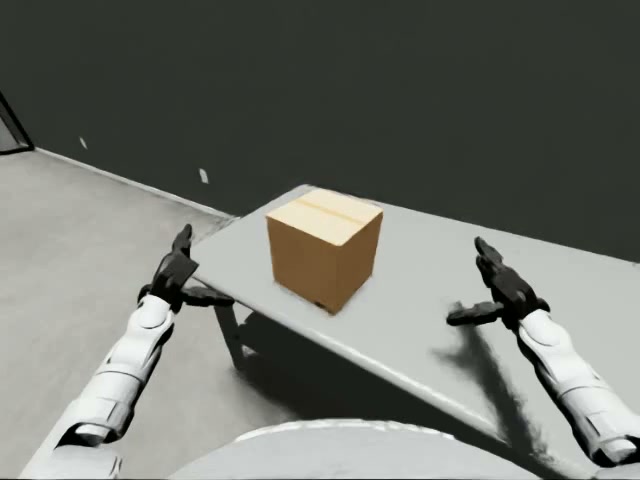}\hfill
    \taskrolloutimg{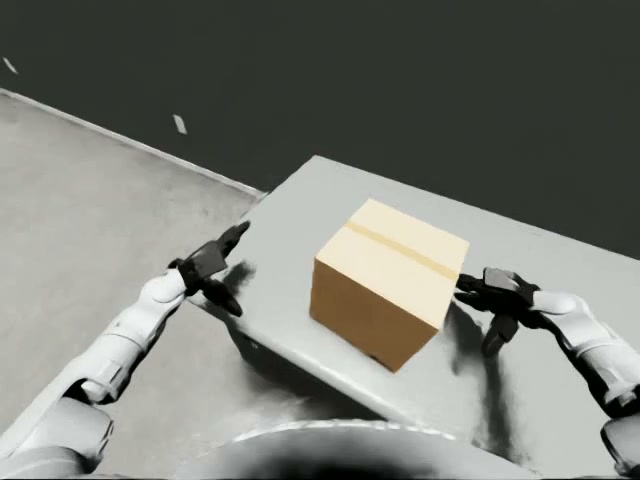}\hfill
    \taskrolloutimg{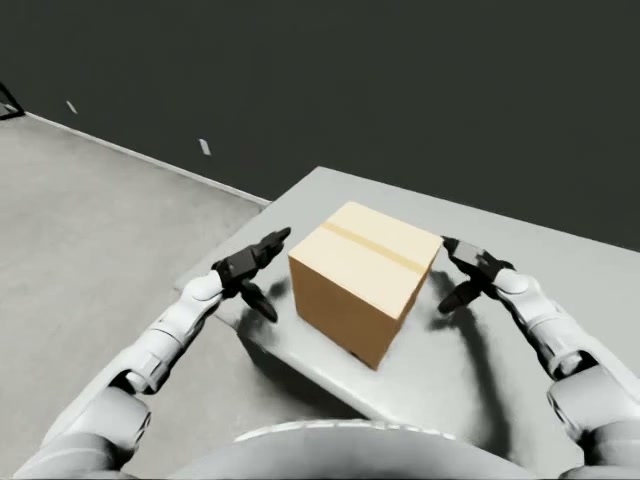}

    \vspace{0.35em}

    \taskrolloutimg{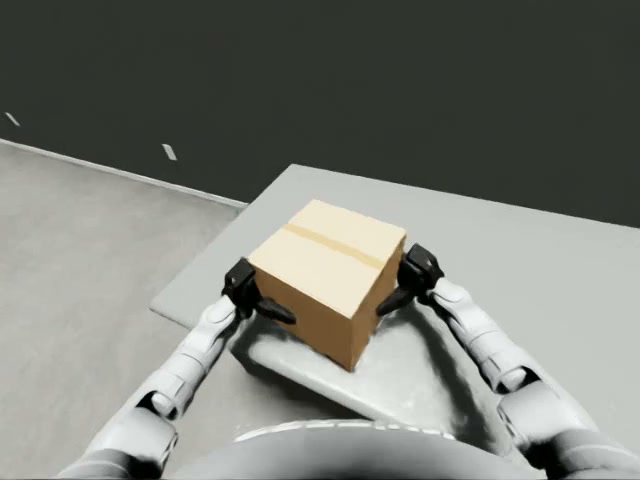}\hfill
    \taskrolloutimg{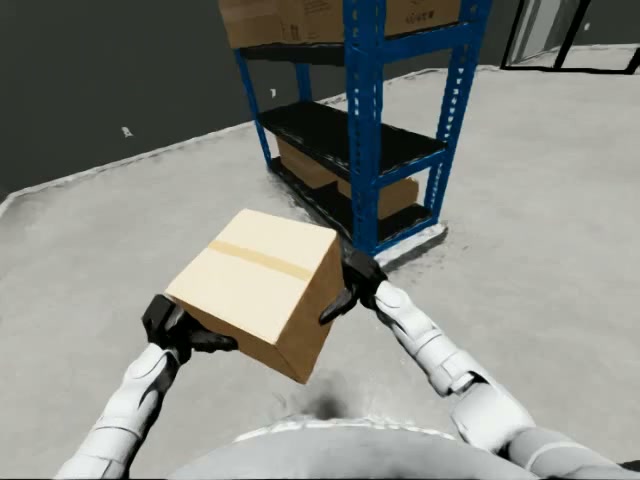}\hfill
    \taskrolloutimg{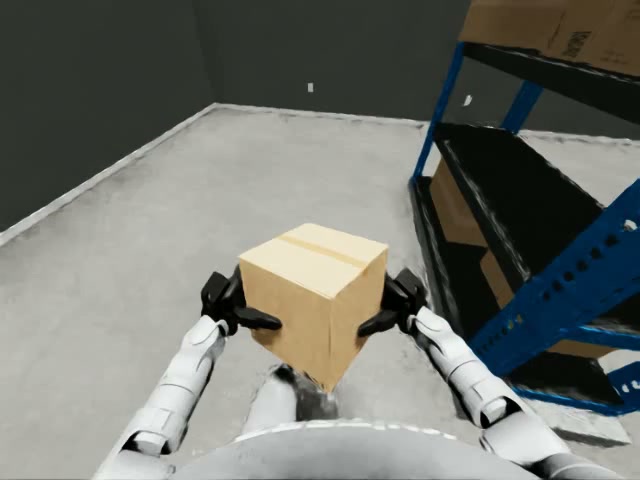}\hfill
    \taskrolloutimg{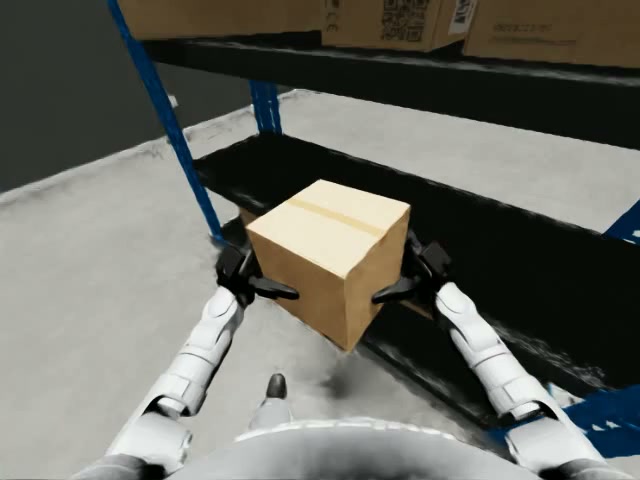}\hfill
    \taskrolloutimg{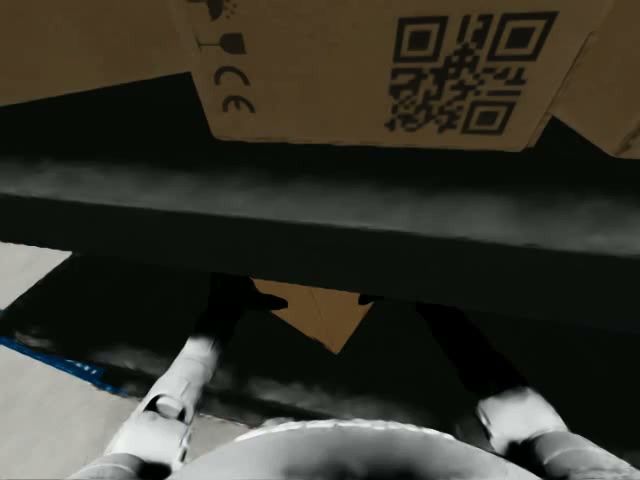}
    \captionof{figure}{P\&PBox successful rollout with 10 uniformly sampled frames shown in temporal order from left to right, top to bottom.}
    \label{fig:app_ppbox_success_rollout}
\end{center}

\begin{center}
    \taskrolloutimg{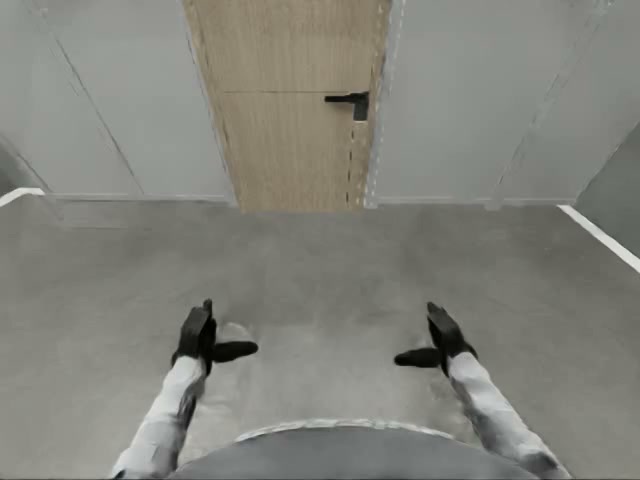}\hfill
    \taskrolloutimg{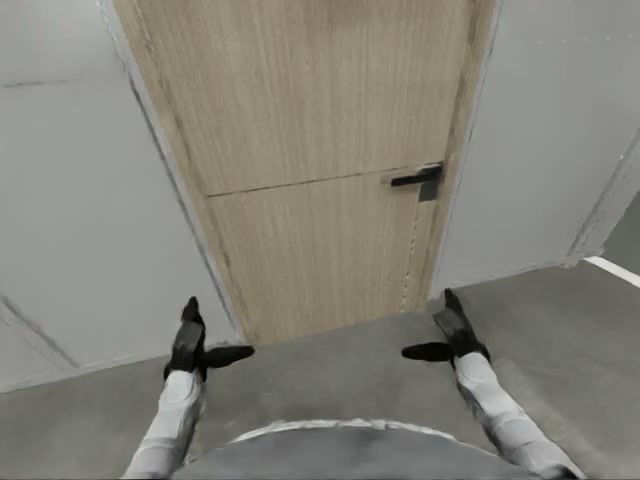}\hfill
    \taskrolloutimg{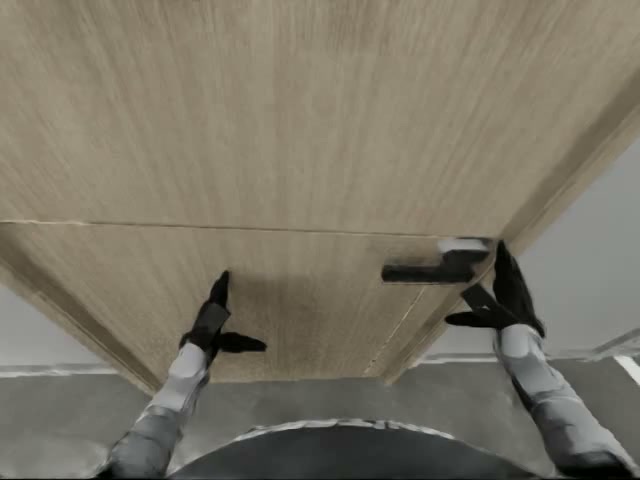}\hfill
    \taskrolloutimg{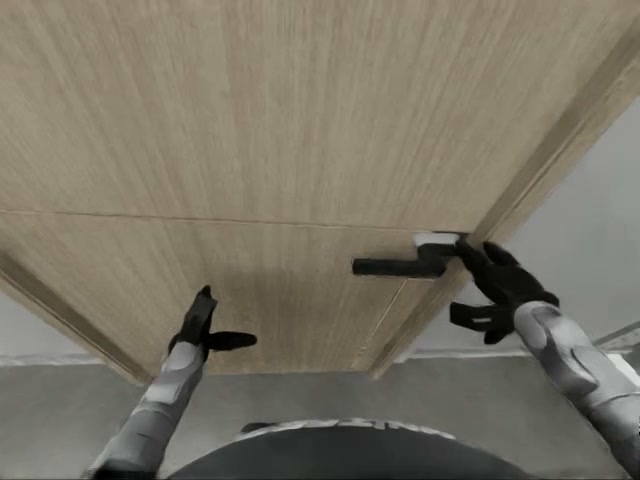}\hfill
    \taskrolloutimg{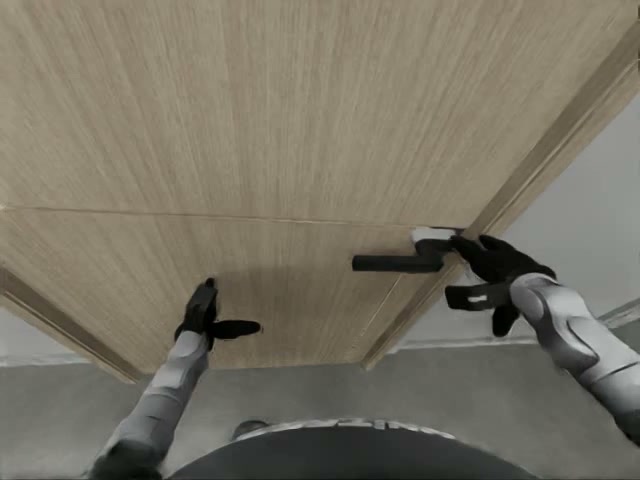}

    \vspace{0.35em}

    \taskrolloutimg{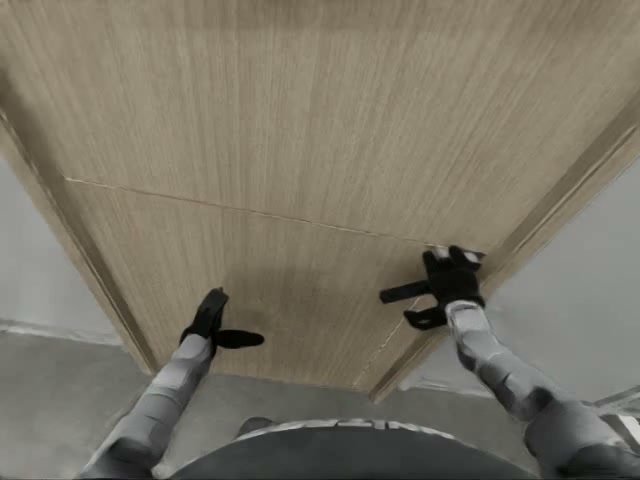}\hfill
    \taskrolloutimg{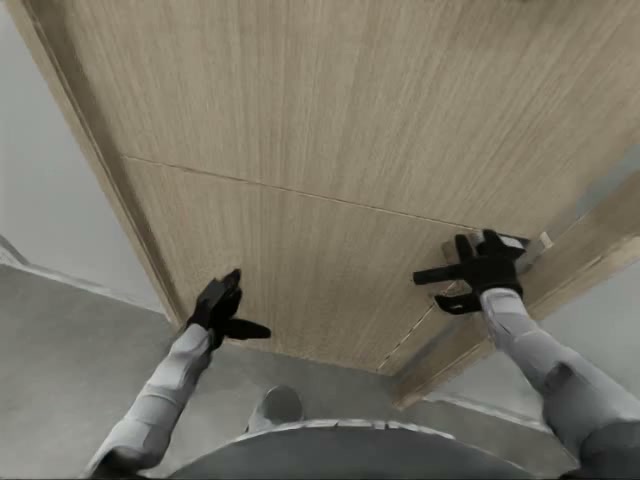}\hfill
    \taskrolloutimg{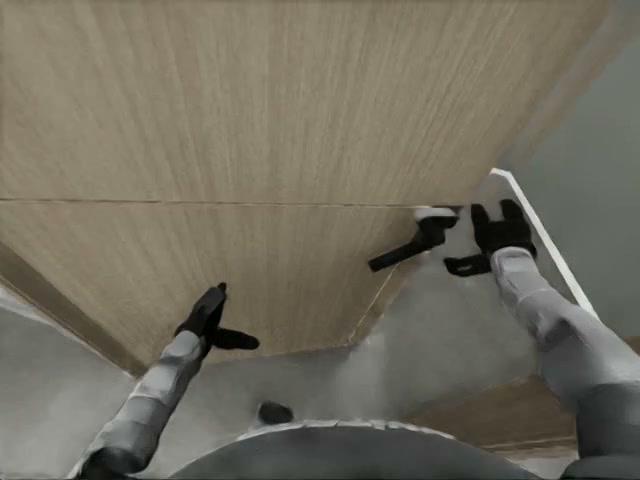}\hfill
    \taskrolloutimg{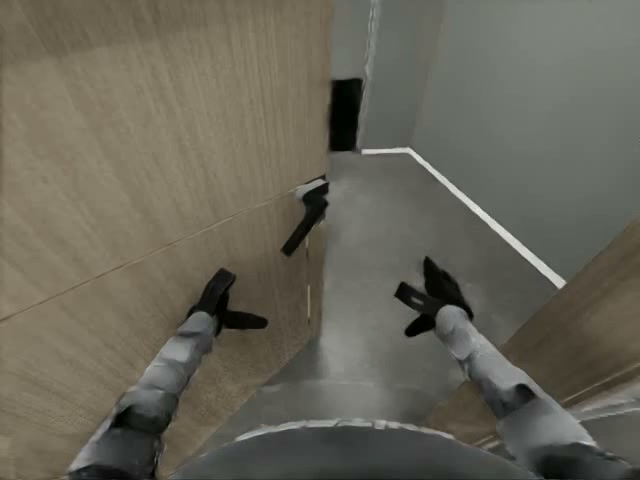}\hfill
    \taskrolloutimg{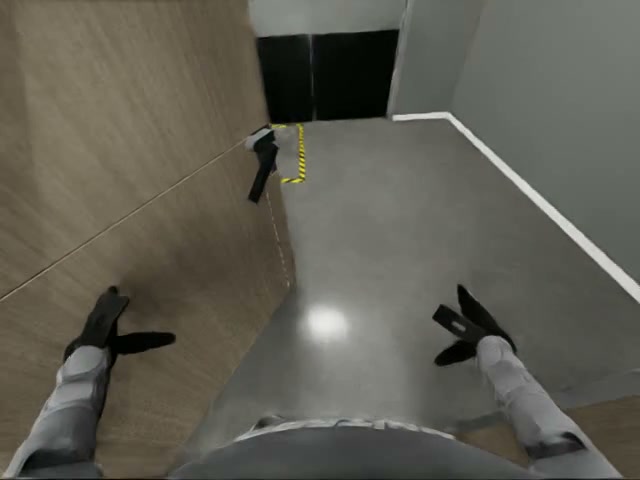}
    \captionof{figure}{OpenDoor successful rollout with 10 uniformly sampled frames shown in temporal order from left to right, top to bottom.}
    \label{fig:app_opendoor_success_rollout}
\end{center}

\begin{center}
    \taskrolloutimg{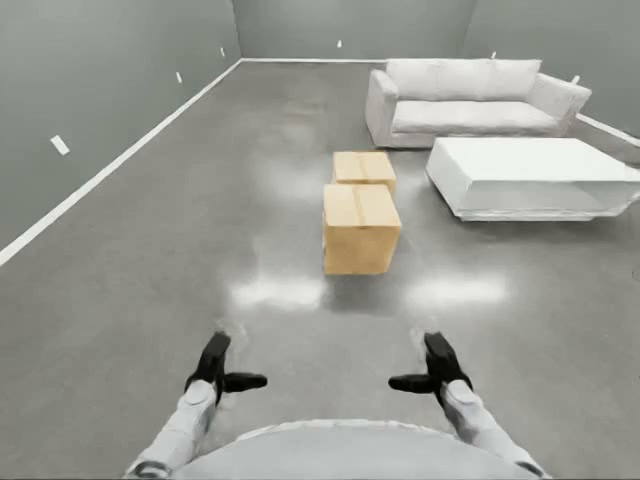}\hfill
    \taskrolloutimg{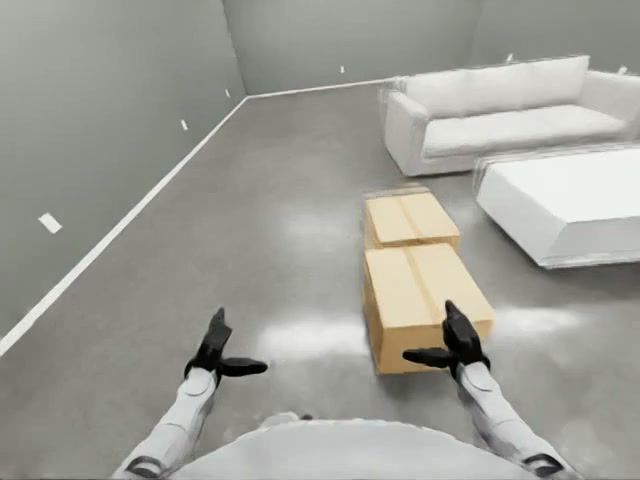}\hfill
    \taskrolloutimg{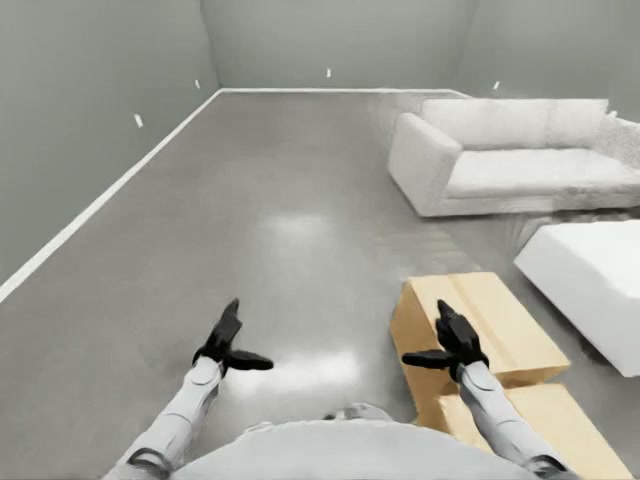}\hfill
    \taskrolloutimg{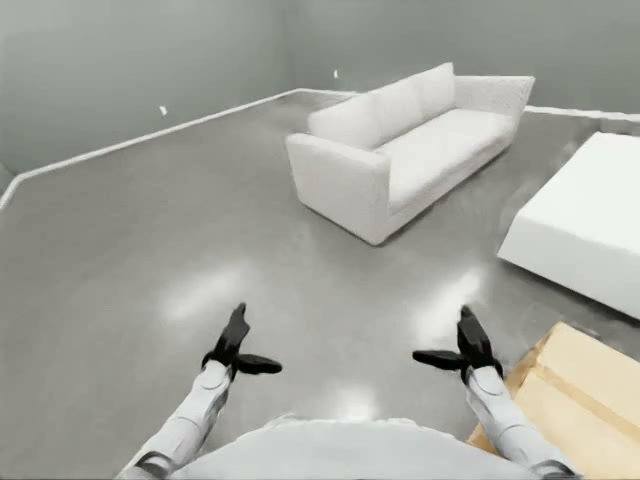}\hfill
    \taskrolloutimg{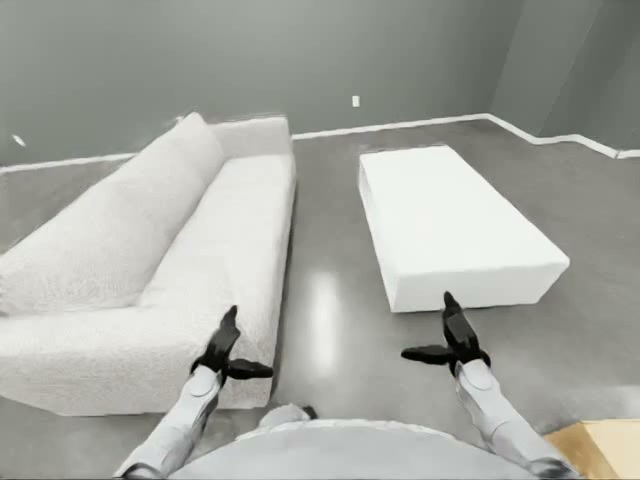}

    \vspace{0.35em}

    \taskrolloutimg{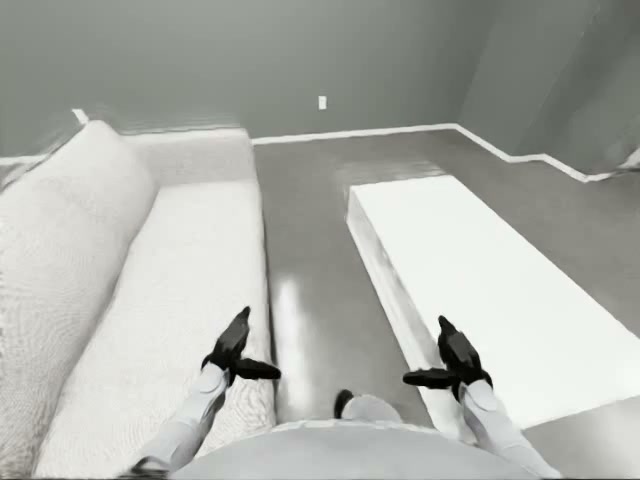}\hfill
    \taskrolloutimg{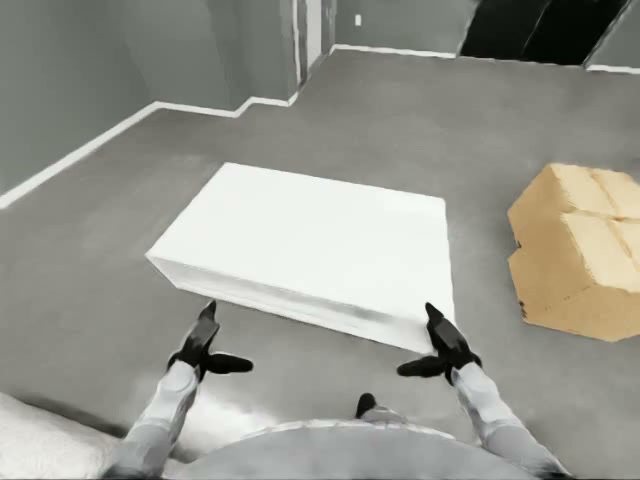}\hfill
    \taskrolloutimg{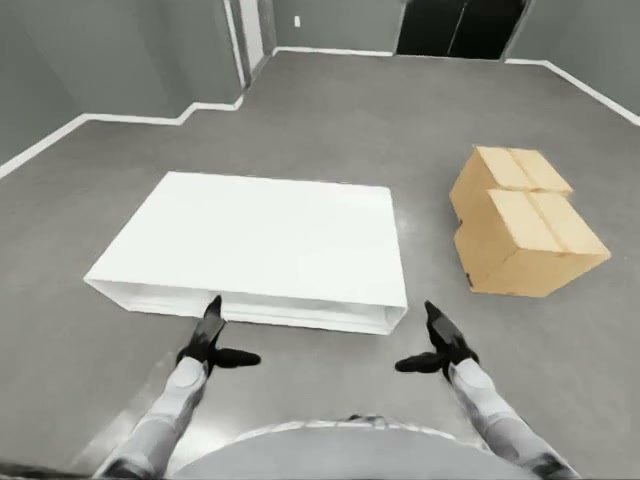}\hfill
    \taskrolloutimg{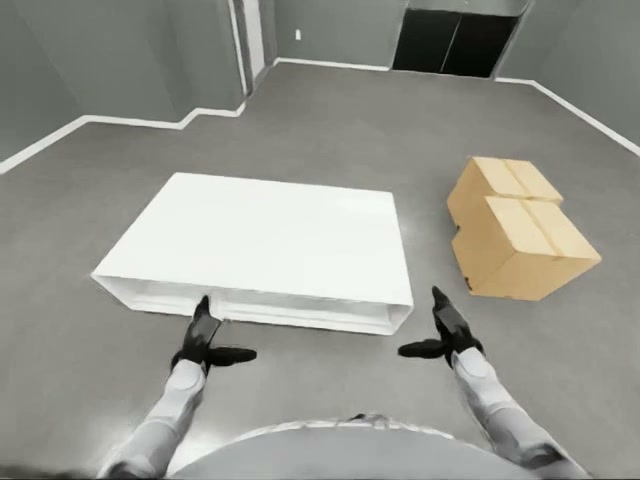}\hfill
    \taskrolloutimg{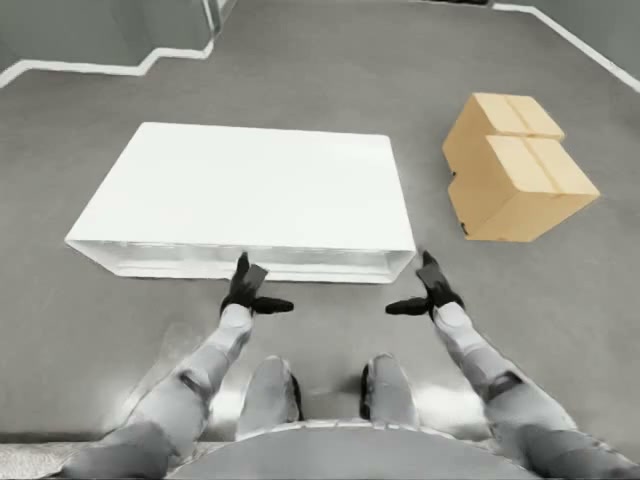}
    \captionof{figure}{SitSofa successful rollout with 10 uniformly sampled frames shown in temporal order from left to right, top to bottom.}
    \label{fig:app_sitsofa_success_rollout}
\end{center}

\begin{center}
    \taskrolloutimg{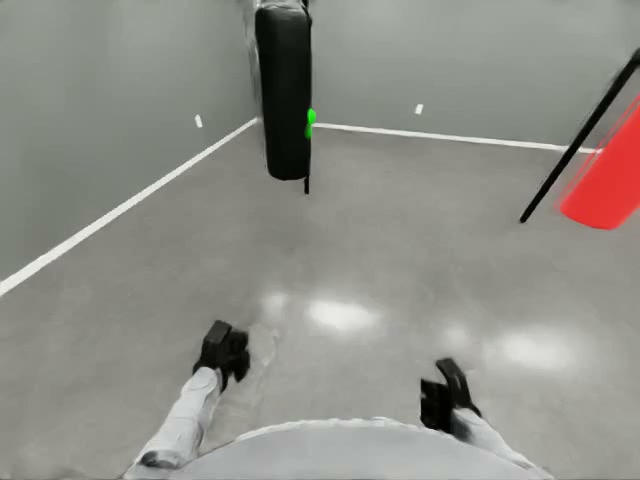}\hfill
    \taskrolloutimg{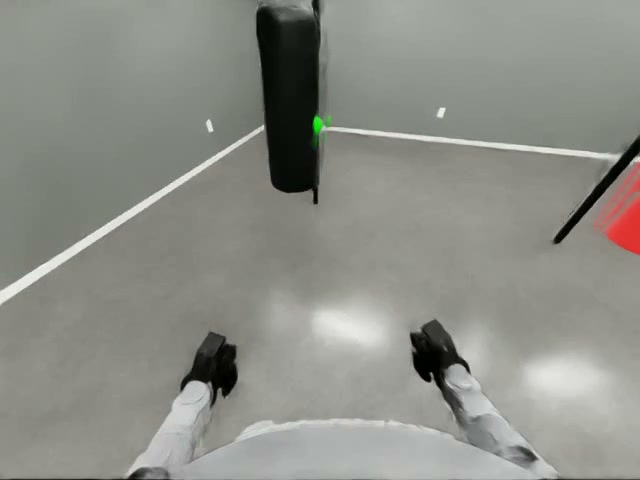}\hfill
    \taskrolloutimg{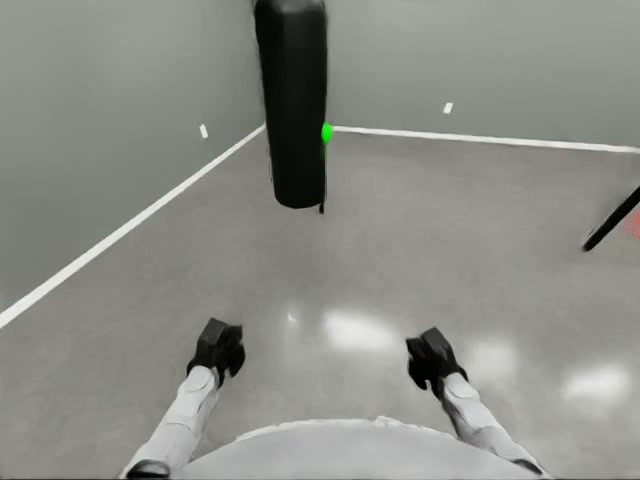}\hfill
    \taskrolloutimg{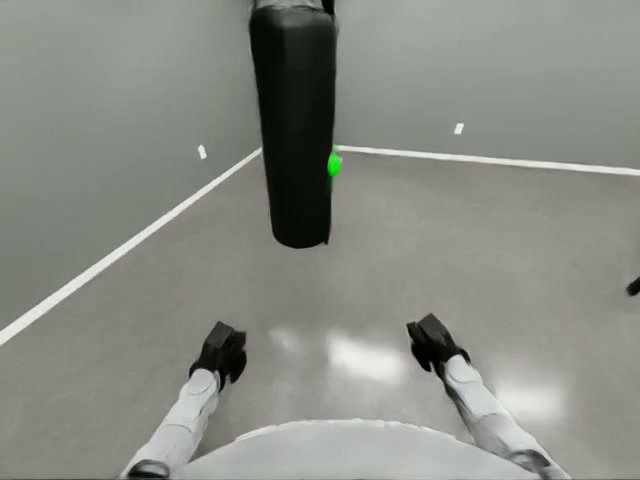}\hfill
    \taskrolloutimg{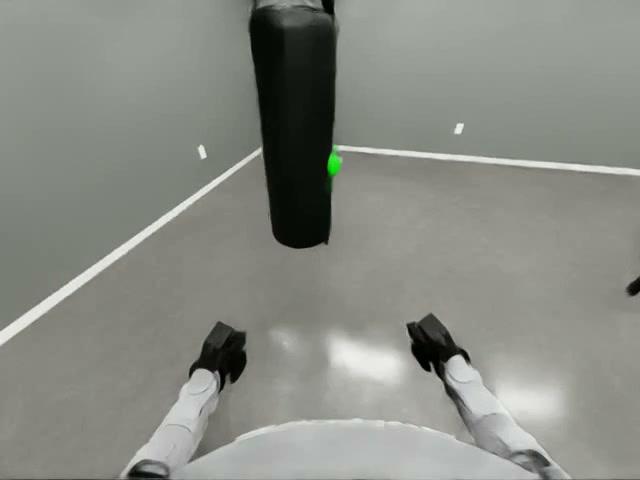}

    \vspace{0.35em}

    \taskrolloutimg{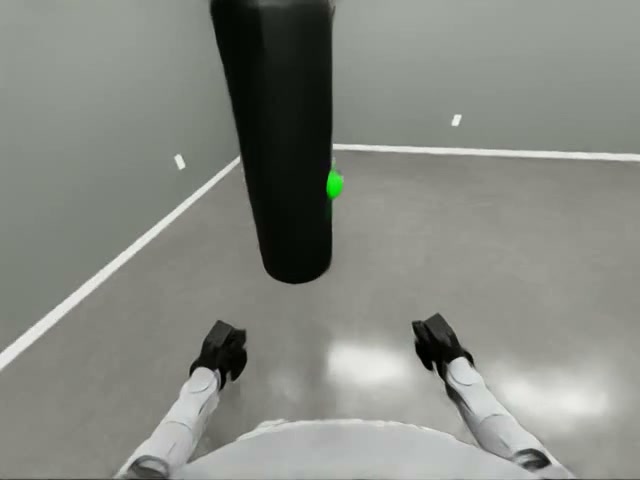}\hfill
    \taskrolloutimg{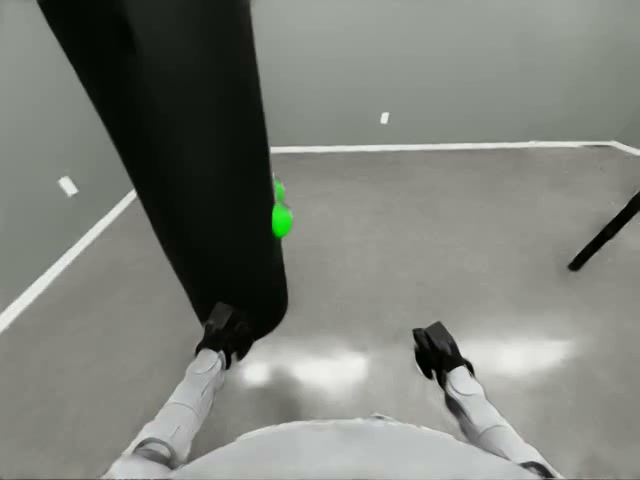}\hfill
    \taskrolloutimg{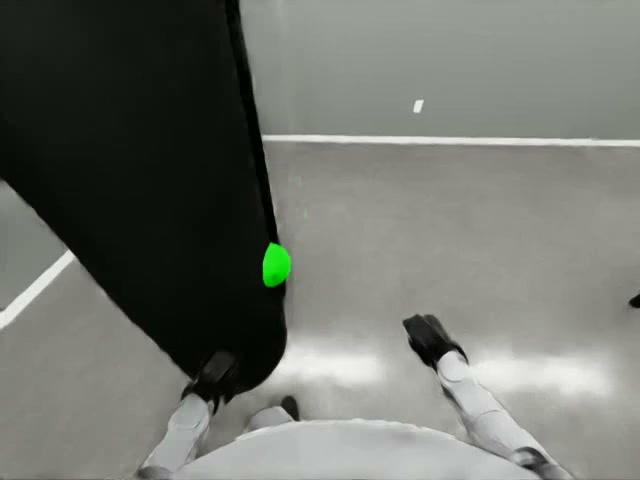}\hfill
    \taskrolloutimg{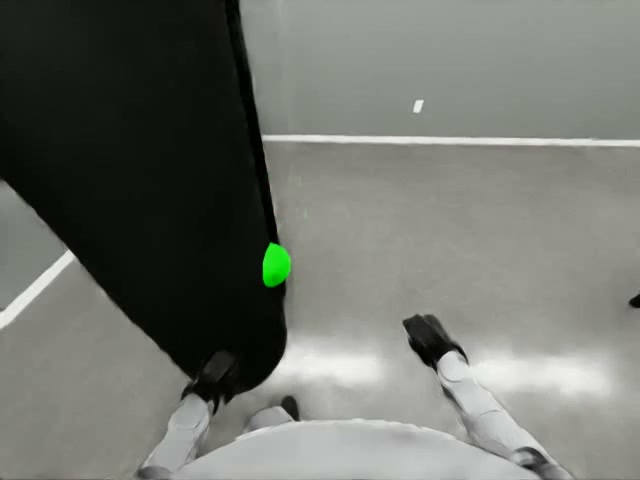}\hfill
    \taskrolloutimg{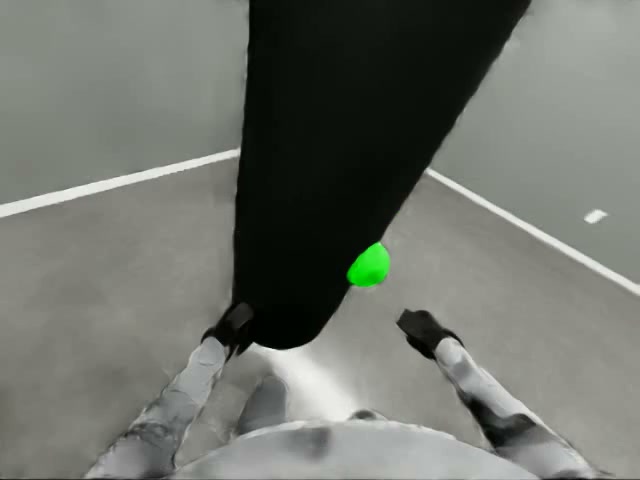}
    \captionof{figure}{Boxing successful rollout with 10 uniformly sampled frames shown in temporal order from left to right, top to bottom.}
    \label{fig:app_boxing_success_rollout}
\end{center}

\begin{center}
    \taskrolloutimg{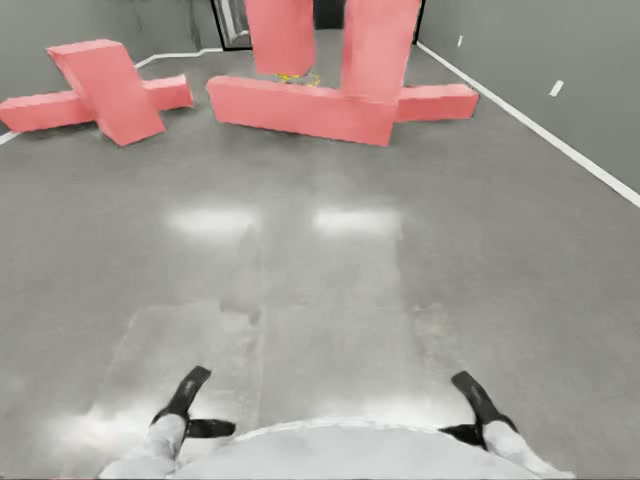}\hfill
    \taskrolloutimg{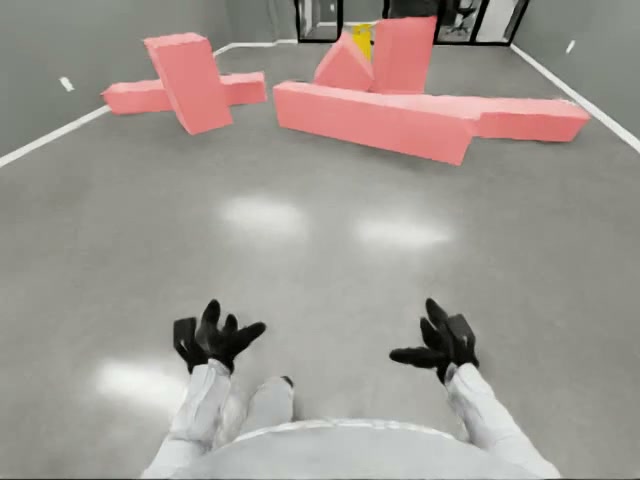}\hfill
    \taskrolloutimg{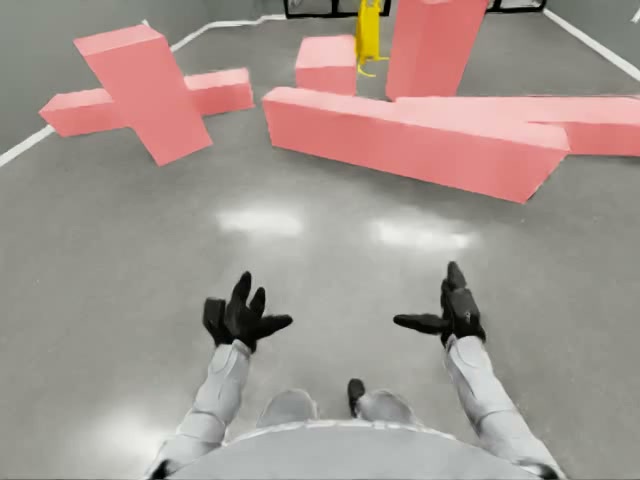}\hfill
    \taskrolloutimg{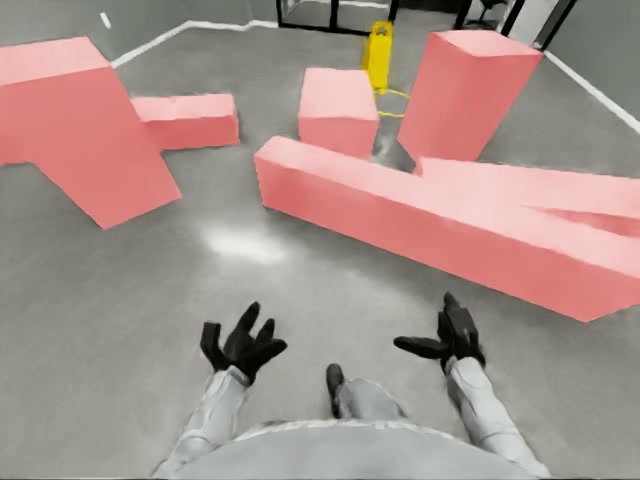}\hfill
    \taskrolloutimg{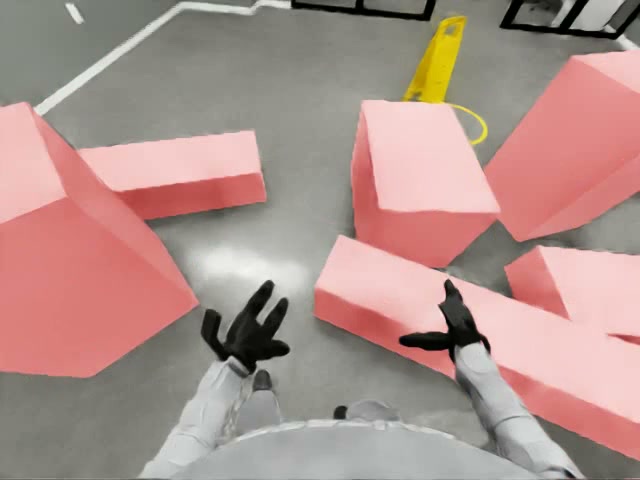}

    \vspace{0.35em}

    \taskrolloutimg{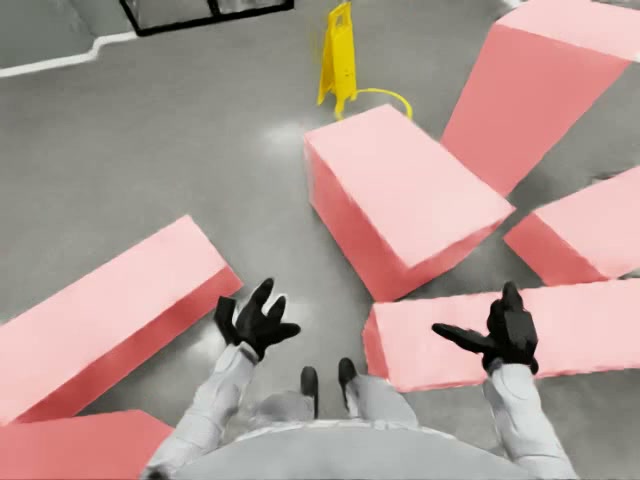}\hfill
    \taskrolloutimg{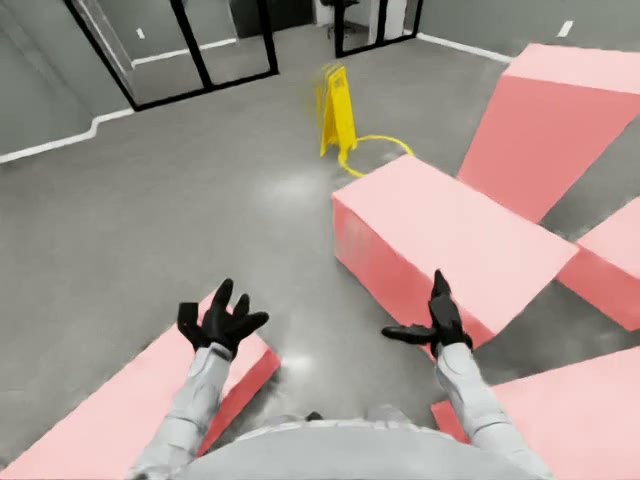}\hfill
    \taskrolloutimg{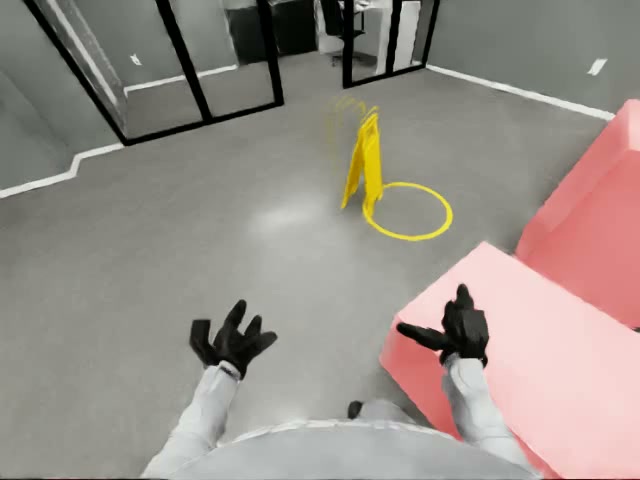}\hfill
    \taskrolloutimg{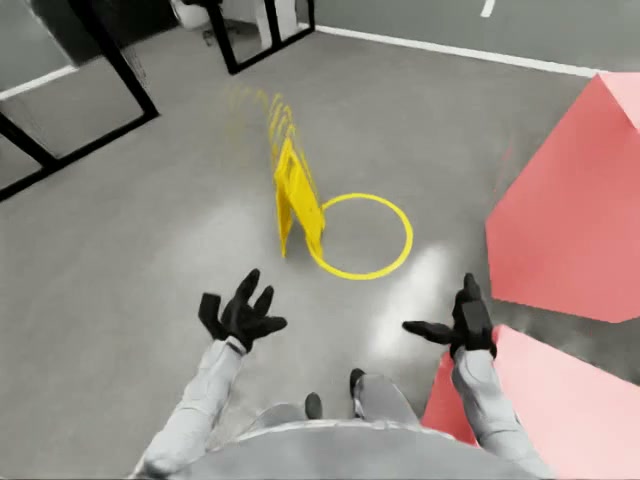}\hfill
    \taskrolloutimg{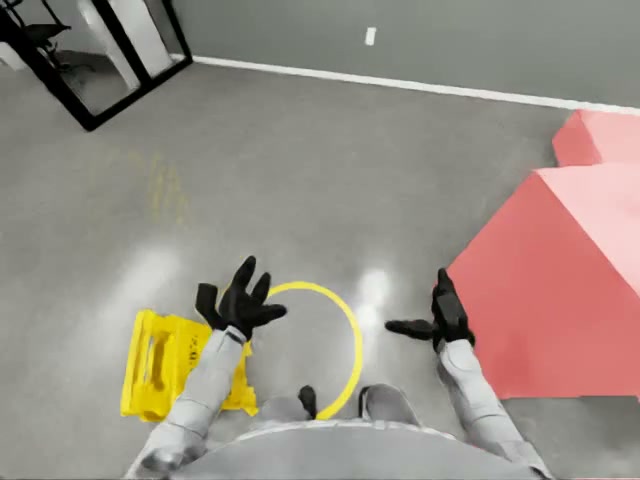}
    \captionof{figure}{VisNavi successful rollout with 10 uniformly sampled frames shown in temporal order from left to right, top to bottom.}
    \label{fig:app_visnavi_success_rollout}
\end{center}

\paragraph{Football (HOI).}
The robot must kick a soccer ball into a single goal net. The ball position is randomized within $[-1.2, 1.2] \times [-1.2, 1.2]$. The scene includes four backdrop walls and a kinematic goal net at 3.0m distance. The ball has physics properties (0.43kg mass, restitution 0.1).
\textbf{Why leg-critical}: Kicking requires the robot to (i) visually locate the ball, (ii) approach and adjust its stance, (iii) balance on the support leg, and (iv) execute a coordinated leg swing with precise impact timing. Each sub-step depends on lower-body dynamics; a stationary-base policy cannot perform the approach-kick sequence.

\paragraph{DoubleDesk (HOI).}
The robot operates between two desks, tasked with picking up a hammer and placing it into a basket. Both the hammer and basket positions are randomized: hammer within $[-3.2, -2.4] \times [-3.15, -3.0]$ and basket within $[-3.6, -2.0] \times [-0.3, -0.005]$ with yaw randomization. If the hammer falls to the ground, it is unrecoverable (training data does not contain ground-pickup demonstrations, and teleoperation-based ground retrieval exceeds current GMT capability boundaries), resulting in eventual timeout.
\textbf{Why leg-critical}: The desktops are at standing height and spaced far enough apart that the robot must walk between the pickup and drop-off locations while carrying a grasped object. Whole-body coordination of locomotion, grasp stability, and target placement is essential; a fixed-station manipulator cannot cover the inter-desk distance.

\paragraph{P\&PBox (HOI).}
The robot picks up a box from a table and places it onto a shelving unit. Both the box and shelf positions are randomized. A dropped box does not cause immediate termination; the robot may attempt to recover if the box remains within reach. Success requires the box center to rest on a detected shelf support surface.
\textbf{Why leg-critical}: The box pickup point (table) and placement target (shelf) are separated horizontally, requiring the robot to walk while holding the box. The shelf's varied heights require the robot to adjust its stance and arm reach, combining locomotion with vertical manipulation.

\paragraph{OpenDoor (HSI).}
The robot must approach and open an articulated door. The door position is randomized within a $[-0.754, 0.55] \times [-0.8, 0.1]$ XY range. The door uses articulation physics with configured joint friction properties. Success requires the robot's root to pass through the door frame plane ($|x_{\text{local}}| \leq 0.55$m, $y_{\text{local}} > 0.02$m past the door) while maintaining an upright standing posture.
\textbf{Why leg-critical}: Unlike fixed-base manipulation, successful door opening requires coordinated base placement, torso orientation, arm contact, and balance recovery under articulated contact forces. The robot must walk through the door while pushing, which demands lower-body locomotion integrated with upper-body manipulation.

\paragraph{SitSofa (HSI).}
The robot navigates a living room environment and must sit on a sofa. Two carton obstacles are randomized in position as distractors. Success is determined when the robot's pelvis or hip joint bodies make stable contact with the sofa seat surface.
\textbf{Why leg-critical}: Success requires obstacle-aware navigation (lower body steers around cartons), precise body-to-sofa alignment (foot placement determines final seating position), and a controlled sitting transition that depends on coordinated leg flexion and torso rotation. A wheeled base cannot execute the sitting motion.

\paragraph{Boxing (HSI).}
The robot stands before a punching bag and must strike a green spherical target mounted on the bag. The target's position is randomized within a 0.15m radius cylinder around two predefined bag centers. Success is determined by the distance between the robot's hand proxy links and the target center falling below a dynamic threshold computed from the asset's bounding radius plus a hand-radius offset (0.08m).
\textbf{Why leg-critical}: The target's randomized height requires the robot to crouch for lower targets and reach high for elevated ones; arm-only motion is insufficient. The dynamic hit-distance threshold depends on the target's world-space bounding box, meaning the robot must coordinate whole-body posture to bring the hand near the target while maintaining balance on two feet.

\paragraph{VisNavi (HSI).}
The robot must locate and walk into a target sign zone within a warehouse environment containing randomized obstacles. The target sign position is randomized with XY pose ranges, and six obstacles (two sets of three) are placed with configurable layout ranges. Success requires both feet to be within the target bounding box while the robot remains upright.
\textbf{Why leg-critical}: The task demands egocentric visual navigation with lower-body steering through cluttered obstacle fields; foot placement accuracy directly determines whether both feet land inside the target zone. Static arm manipulation or a mobile-base abstraction cannot satisfy the dual-foot zone constraint.

\paragraph{Implementation Note.}
To ensure reproducibility and maintain a consistent evaluation interface across tasks, \textsc{HumanoidArena} adopts a modular task implementation. 
Task scenes and object layouts are specified under \texttt{tasks/common\_scene/}, and environment configurations, including initialization ranges and task-level randomization settings, are defined under \texttt{tasks/common\_env\_config/}. 
The reward predicates and success/termination checks for each task are implemented in its corresponding \texttt{mdp/rewards.py} file. 
During evaluation, fall detection is handled uniformly in \texttt{script/eval\_scripts/*/sim\_eval\_vla.py}, where a 60\textdegree\ torso-tilt threshold is treated as a soft fall condition after 5-frame confirmation, while a 75\textdegree\ hard-tilt threshold or a 50N body-contact force under tilt triggers immediate termination.

\section{Data Collection Pipeline and Dataset Specification}
\label{app:data_spec}

This section describes the data collection pipeline and the resulting dataset specification. A core design principle of \textsc{HumanoidArena} is that all demonstrations share a unified \textit{policy-facing canonical action space} (the 40D intermediate whole-body action defined in~\S C), while the low-level execution dynamics differ depending on which GMT executes the action. This separation allows the benchmark to study whether high-level policies learn reusable whole-body intent (via the shared action interface) or overfit to tracker-specific dynamics (via \textit{cross-GMT} evaluation). Switching the GMT backend under the same teleoperation interface yields multi-GMT datasets with identical policy-facing semantics but different tracking characteristics.

\subsection{Collection Pipeline}

Human demonstrations are collected through a shared upstream teleoperation pipeline, followed by backend-specific action interpretation inside Isaac Lab:

\begin{itemize}
    \item \textbf{Shared capture and retargeting (both backends):} The operator receives the humanoid's egocentric video stream via PICO headset. Human motion is captured and retargeted through GMR to produce a 35D robot-space reference signal (\texttt{mimic\_obs}), which is published to Redis for real-time consumption.
    \item \textbf{Backend-specific action interpretation:} Inside Isaac Lab, the 35D \texttt{mimic\_obs} is consumed by a backend-specific action provider — \texttt{action\_provider\_wh\_twist2} for the TWIST2 tracker, and \texttt{action\_provider\_sonic} for the SONIC tracker. Each provider interprets the shared 35D signal into executable G1 joint targets using its own policy (TWIST2: ONNX inference; SONIC: GEAR-SONIC encoder/decoder). The GMT then executes these targets in simulation.
\end{itemize}

Within Isaac Lab, the action provider consumes poses from Redis, runs a pretrained policy, and drives the G1 humanoid robot. A \texttt{RecordingManager} buffers per-frame state and action data, serializing episodes into NPZ archives along with multi-camera RGB video. Teleoperation and data collection use a single egocentric front camera at 640$\times$480 resolution; the multi-camera and camera depth recordings are generated during data re-recording and are available for training and inference. The high-level policy outputs at 50Hz for both backends; GMT-specific decimation aligns both to the same 50Hz control frequency for fair \textit{cross-GMT} comparison. Data collection and teleoperation consistently operate at 50Hz.

\subsection{Canonical State and Action Schema}

All recorded demonstrations are normalized into the shared policy-facing interface defined in Appendix~C. 
Specifically, each episode stores synchronized egocentric observations, a 64D canonical proprioceptive state, and a 40D intermediate whole-body action. 
The detailed state/action schema, coordinate conventions, canonical conversion procedure, and GMT-specific adapter mappings are described in Appendix~C.

\subsection{Dataset Statistics}

Each task provides 100 successful demonstration episodes per GMT backend (TWIST2 and SONIC), totaling 200 episodes per task and 1,400 episodes across all tasks. Although the total demonstration duration may be modest relative to large-scale video pretraining corpora, each episode contains dense 50Hz whole-body state-action supervision, synchronized egocentric observations, and closed-loop GMT-executed humanoid behavior. The dataset is therefore intended primarily for controlled evaluation of hierarchical whole-body policies rather than large-scale pretraining.

\subsection{Dataset Format}

Each recorded episode is stored as an NPZ archive containing per-frame arrays: front-camera RGB images (640$\times$480, with additional multi-camera views available from re-recording), 64D canonical state, and 40D canonical action. Unless otherwise specified, all reported benchmark results use the front egocentric camera as the policy visual input. The NPZ archives are converted to LeRobot-compatible format using the conversion pipeline described below. Training, validation, and test splits are held constant across all evaluated policies and GMT backends. Only successful demonstrations are included in the training set.

\subsection{Conversion Pipeline}

Raw NPZ recordings are converted into LeRobot-compatible datasets using dedicated conversion tools. The canonical extraction helpers read existing canonical fields when available or reconstruct 64D state and 40D action from recorded joint arrays and root orientation. Cleaning tools filter invalid frames due to time discontinuities or held-action motion mismatches. A verification script cross-checks converted dataset rows against canonical arrays reconstructed from source recordings to detect conversion drift.

For multi-camera datasets, re-recording tools replay NPZ episodes through Isaac Lab to generate additional viewpoint images.

\section{Intermediate Whole-Body Action Interface and GMT Adapters}
\label{app:action_gmt}

This appendix defines the policy-facing canonical interface used throughout \textsc{HumanoidArena}. All high-level policies observe the same 64D proprioceptive state and predict the same 40D intermediate whole-body action, irrespective of which GMT executes the motion.

\begin{table}[h]
\centering
\caption{Canonical 64D state and 40D intermediate whole-body action representation.}
\label{tab:canonical-schema}
\begin{tabular}{lll}
\toprule
\textsc{Component} & \textsc{Dims} & \textsc{Description} \\
\midrule
\multicolumn{3}{l}{\textbf{Observation State (64D)}} \\
\textsc{root\_rot6d} & 6  & \textsc{Root orientation in 6D continuous rotation} \\
\textsc{dof\_pos}    & 29 & \textsc{Joint positions for 29-DoF whole-body G1} \\
\textsc{dof\_vel}    & 29 & \textsc{Joint velocities} \\
\multicolumn{3}{l}{\textbf{Intermediate Whole-Body Action (40D)}} \\
\textsc{root\_xy\_delta} & 2  & \textsc{Root translation in xy axis} \\
\textsc{root\_z}         & 1  & \textsc{Root-height target} \\
\textsc{root\_rot6d}     & 6  & \textsc{Target root orientation in 6D continuous rotation} \\
\textsc{joint\_pos}      & 29 & \textsc{Target joint positions} \\
\textsc{hand\_binary}    & 2  & \textsc{Left/right hand open-close commands} \\
\bottomrule
\end{tabular}
\end{table}

The 40D action is intentionally structured: it encodes coordinated whole-body intent (root motion, joint targets, hands) while remaining compact enough for visuomotor policy learning. It delegates dynamic feasibility and fast stabilization to the low-level GMT rather than requiring the policy to predict torque-level commands.

\paragraph{Canonical conversion.} During demonstration collection and policy rollout, the reference signal from the teleoperation pipeline is normalized into the canonical interface. The conversion runs through \texttt{action\_provider/vla\_smpl\_runtime.py}:
\begin{itemize}
    \item \texttt{build\_vla\_observation\_state()} constructs the 64D proprioceptive state from root orientation, joint positions, and joint velocities.
    \item \texttt{build\_vla\_action()} constructs the 40D intermediate action from root motion, orientation, joint targets, and hand commands.
    \item \texttt{UnifiedSMPLActionRuntime} lifts tracker-specific or teleoperation-specific references into the shared canonical space when needed.
\end{itemize}
With this conversion, policies trained on demonstrations collected with different GMTs observe and produce identical state and action formats.

\paragraph{GMT adapters.} Each GMT exposes a different execution interface. A GMT-specific adapter $\psi_m$ maps the canonical 40D action $u_t$ into the reference format required by GMT $G^{(m)}$, producing executable G1 joint-position targets $q_{t+1}^{(m)} = G^{(m)}(\psi_m(u_t))$. The adapter changes only the action-to-tracker mapping; the high-level policy, observation space, and canonical action space remain unchanged across GMT backends. This design makes \textit{in-GMT} and \textit{cross-GMT} evaluation well-defined: \textit{cross-GMT} deployment replaces only the adapter and low-level tracker, while keeping the trained policy and its output space fixed.

\textsc{HumanoidArena} currently instantiates two GMT backends. The TWIST2 adapter maps the canonical 40D action into a mimic-style tracking command for TWIST2's ONNX policy. The SONIC adapter maps the same 40D action into the reference format required by the SONIC execution stack, which uses a GEAR-SONIC encoder/decoder pair for tracking. Both backends consume the same upstream 35D \texttt{mimic\_obs} signal produced by the shared GMR retargeting pipeline (\S B), ensuring consistent canonical conversion and enabling fair \textit{cross-GMT} comparison.

\paragraph{Implementation Note.} The canonical runtime is implemented in \texttt{action\_provider/vla\_smpl\_runtime.py}. The dataset-level conversion pipeline lives under \texttt{tools/data\_tools/}, including \texttt{twist2lerobot\_64\_40.py}, \texttt{sonic2lerobot\_64\_40.py}, and shared helpers in \texttt{smpl\_lerobot\_common.py}.

\section{Evaluation Protocols and Perturbation Design}
\label{app:eval_protocol}

This section specifies the evaluation protocol, the three perturbation axes used for robustness testing, the \textit{in-GMT} and \textit{cross-GMT} deployment protocols, and the complete evaluation coverage matrix. All protocols ensure that different policies and backends are compared under identical rollout seeds, held-out object configurations, and initial robot states.

\subsection{Evaluation Protocol}

All models are evaluated under a consistent protocol: each model is tested with $N_{\text{seeds}} = 3$ random seeds, each seed running $N_{\text{repeats}} = 20$ repeated trials, totaling 60 episodes per configuration. The simulation uses persistent mode (\texttt{persistent\_sim = true}) where the Isaac Lab environment is created once and reset between episodes. Episode seeds are derived deterministically as $\text{sha256}(\text{task\_name} | \text{group\_seed} | \text{repeat\_idx})$. Success is defined as a binary reward: 1 if the task-specific success condition is met (as defined in \S A, Table~\ref{tab:termination}), 0 otherwise. In addition to success rate, we report average fall rate~(AFR) as a stability metric.

\subsection{\textit{In-GMT} and \textit{Cross-GMT} Evaluation}

\textsc{HumanoidArena} supports two GMT deployment protocols that test the generalization of learned high-level policies across execution backends:
\begin{itemize}
    \item \textbf{\textit{In-GMT} (matched)}: The high-level policy is trained on demonstrations collected and executed with GMT-$A$ and evaluated with the same GMT-$A$. This measures standard task performance under a matched data-collection and execution backend.
    \item \textbf{\textit{Cross-GMT} (transferred)}: The high-level policy is trained on GMT-$A$ demonstrations but deployed with GMT-$B$. This measures whether the intermediate whole-body action representation transfers across execution backends.
\end{itemize}

\subsection{Test Modes}

We define four test modes that systematically vary different axes of perturbation while keeping the evaluation protocol identical. Each perturbation axis is designed to change the target test variable without altering the task's underlying semantics:

\begin{table*}[h]
\centering
\caption{Four test settings and their perturbation dimensions}
\label{tab:test-modes}
\begin{adjustbox}{max width=\textwidth}
\begin{tabular}{lcccc}
\toprule
\textsc{Mode} & \textsc{Object Pose Range} & \textsc{Asset} & \textsc{Lighting} \\
\midrule
\textsc{Base Test} & \textsc{Training range} & \textsc{Original} & \textsc{DomeLight (uniform)} \\
\textsc{Execution Test} & \textbf{Expanded} & \textsc{Original} & \textsc{DomeLight (uniform)} \\
\textsc{Semantic Test} & \textsc{Training range} & \textbf{Replaced USD} & \textsc{DomeLight (uniform)} \\
\textsc{Visual Test} & \textsc{Training range} & \textsc{Original} & \textbf{DistantLight + random rotation} \\
\bottomrule
\end{tabular}
\end{adjustbox}
\end{table*}

\subsection{Execution Test: Expanded Pose Randomization}

The execution test increases the randomization range of object positions beyond the training distribution. For each task, the \texttt{deterministic\_object\_resets[].pose\_range} is widened in the execution YAML config relative to the base test. For example, the football task's ball XY range expands from $[-1.2, 1.2]$ to $[-1.45, 1.45]$, and the door position expands from $x \in [-0.754, 0.55]$ to $[-0.9, 0.66]$. All other parameters (control frequency, decimation, physics properties) remain identical to base test.

\subsection{Semantic Test: Asset Replacement}

We use ``semantic'' in the benchmark sense of \textit{semantics-preserving asset substitution}: the task goal and affordance remain unchanged while the visual instance, shape, or distractor configuration varies. The test evaluates robustness to appearance, geometry, and scene-clutter variations by substituting specific scene assets with pre-built variant USD files.

The semantic variants fall into three categories: (i) \textit{geometry-preserving appearance variants} where textures or colors are modified while geometry remains unchanged (OpenDoor door, Football goal net, SitSofa/DoubleDesk room walls); (ii) \textit{affordance-preserving geometry variants} where the object shape changes but the functional affordance and comparable physical extent are preserved (Boxing: green sphere $\rightarrow$ green cube of equal bounding extent); (iii) \textit{scene-clutter distractor variants} where additional non-target objects are introduced to test robustness under scene clutter while preserving the target goal (P\&PBox: distractor objects on the desk surface).

\begin{table}[h]
\centering
\caption{Semantic asset substitutions across the 7 tasks.}
\label{tab:semantic-overrides}
\begin{adjustbox}{max width=\textwidth}
\begin{tabular}{lll}
\toprule
\textsc{Task} & \textsc{Scene Attribute} & \textsc{Semantic Variant} \\
\midrule
\textsc{Football} & \texttt{goal\_net}        & \textsc{Original goal $\rightarrow$ texture-modified goal} \\
\textsc{DoubleDesk} & \texttt{room\_walls}      & \textsc{Original room $\rightarrow$ texture-modified room} \\
\textsc{P\&PBox} & \texttt{room\_walls}     & \textsc{Original room $\rightarrow$ room with distractor objects} \\
\textsc{OpenDoor} & \texttt{door}             & \textsc{Original door $\rightarrow$ textured variant} \\
\textsc{SitSofa} & \texttt{room\_walls}      & \textsc{Original room $\rightarrow$ texture-modified room} \\
\textsc{Boxing} & \texttt{boxing\_target}   & \textsc{Green sphere $\rightarrow$ green cube} \\
\textsc{VisNavi} & \texttt{target\_sign}     & \textsc{Wet floor sign $\rightarrow$ semantic variant} \\
\bottomrule
\end{tabular}
\end{adjustbox}
\end{table}

Asset overrides are specified in YAML test configurations using the environment configuration loader. The loader traverses the config object hierarchy and sets the asset's \texttt{usd\_path} attribute. No code changes are required in the evaluation pipeline — semantic testing is purely configuration-driven. All YAML configs reside under \texttt{tasks/common\_test\_config/semantic/}.

\subsection{Visual Test: Lighting Randomization}

The visual test evaluates robustness to illumination variations. The default DomeLight (uniform ambient illumination) is replaced at runtime with a DistantLight (directional light source) whose pose and intensity are randomized per episode.

\begin{itemize}
    \item \textbf{Light Replacement}: After simulator initialization, the existing DomeLight prim at \texttt{/World/light} is deleted and a DistantLight is created with default parameters: position $(-4, -1, 18)$, intensity $5000$, angle $15^\circ$, color $(0.75, 0.75, 0.75)$.
    \item \textbf{Per-Episode Randomization}: Before each episode reset, the DistantLight parameters are deterministically randomized using the episode seed:
    \begin{itemize}
        \item Rotation: pitch $\sim \mathcal{U}(-20^\circ, +20^\circ)$, roll $\sim \mathcal{U}(-20^\circ, +20^\circ)$, yaw fixed at $0^\circ$.
        \item Intensity: $\sim \mathcal{U}(3000, 7000)$.
        \item Color: RGB each channel $\sim \mathcal{U}(0.65, 0.85)$.
    \end{itemize}
\end{itemize}

The DistantLight's directional nature means that rotating the light changes shadow directions, highlight positions, and overall scene contrast, testing the model's robustness to lighting variations without changing the physical task configuration. Implementation resides in \texttt{tools/augmentation\_utils.py} (light replacement and randomization) and \texttt{tasks/common\_runtime/env\_runtime\_hooks.py} (per-episode integration).
\begin{table}[h]
\centering
\caption{Evaluation matrix. $\checkmark$ = completed, N.E. = not evaluated for the listed mode.}
\label{tab:eval-matrix}
\begin{adjustbox}{max width=\textwidth}
\begin{tabular}{lcccc}
\toprule
\textsc{Configuration} & \textsc{Base Test} & \textsc{Execution} & \textsc{Semantic} & \textsc{Visual} \\
\midrule
\textsc{In-GMT twist2} & $\checkmark$ (7 tasks) & $\checkmark$ (7) & $\checkmark$ (7) & $\checkmark$ (7) \\
\textsc{In-GMT sonic}  & $\checkmark$ (7 tasks) & $\checkmark$ (7) & $\checkmark$ (7) & $\checkmark$ (7) \\
\textsc{Cross-GMT (twist2 $\rightarrow$ sonic)} & $\checkmark$ (7 tasks) & N.E. & N.E. & N.E. \\
\textsc{Cross-GMT (sonic $\rightarrow$ twist2)} & $\checkmark$ (7 tasks) & N.E. & N.E. & N.E. \\
\bottomrule
\end{tabular}
\end{adjustbox}
\end{table}
\subsection{Evaluation Matrix}

Table~\ref{tab:eval-matrix} summarizes the complete evaluation coverage. All results use 3 seeds $\times$ 20 repeats = 60 episodes per entry. In-GMT denotes matched training and evaluation with the same backend; Cross-GMT denotes training on one backend's demonstrations and deployment on the other.

% \clearpage
\section{Additional Experiments and Rollout Analysis}
\label{app:qualitative}

\subsection{Predicted Action Distributions under \textit{Cross-GMT} Observations}
\label{app:tsne_action_distribution}

This section provides a qualitative analysis of why \textit{cross-GMT} deployment is challenging. We ask whether a fixed policy produces similar whole-body actions when conditioned on observations collected with different general motion trackers (GMTs). Instead of visualizing ground-truth dataset actions, we visualize \emph{policy-predicted} actions: each point corresponds to a 40D action predicted by a task-specific diffusion policy from one sampled egocentric observation. The policy input contains the front RGB image and the 64D proprioceptive state, while the t-SNE embedding is computed only from the predicted 40D action after policy post-processing.

For each task and each diffusion policy, we sample 1,000 SONIC observations and 1,000 TWIST2 observations, using 10 uniformly spaced frames from each of the 100 episodes per source. We run the same inference pipeline as the HTTP policy server, including image preprocessing, policy normalization, post-processing, and action unnormalization. For diffusion policies, we use the first action from the predicted action chunk as the immediate policy output, avoiding the stateful action queue used during online \texttt{select\_action} inference.

\begin{figure}[!ht]
\centering
\includegraphics[draft=false,width=0.86\linewidth]{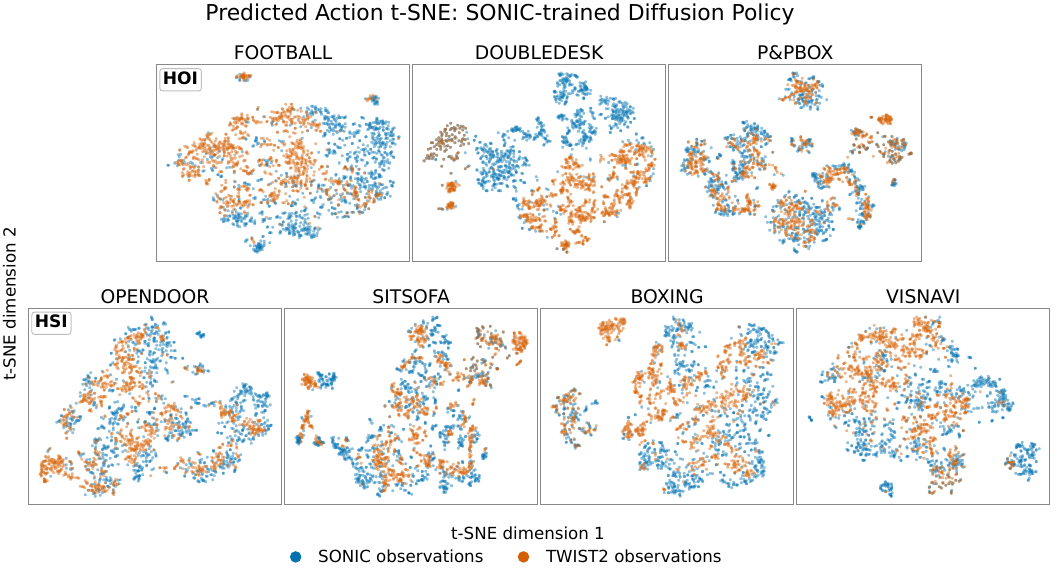}
\caption{\textbf{SONIC-trained diffusion policies.} t-SNE visualization of predicted 40D actions on the seven benchmark tasks. Blue and orange points denote actions induced by SONIC and TWIST2 observations, respectively, using 1,000 samples per source. Larger color separation indicates a stronger GMT-dependent shift in the immediate action distribution.}
\label{fig:tsne-sonic-policy}
\end{figure}

\begin{figure}[!ht]
\centering
\includegraphics[draft=false,width=0.86\linewidth]{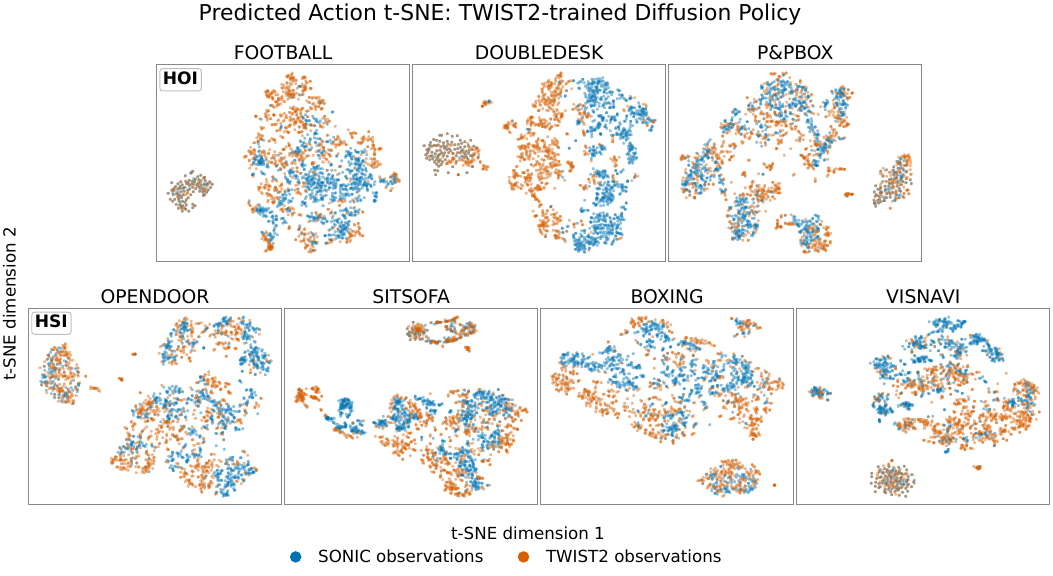}
\caption{\textbf{TWIST2-trained diffusion policies.} t-SNE visualization of predicted 40D actions on the seven benchmark tasks. Blue and orange points denote actions induced by SONIC and TWIST2 observations, respectively, using 1,000 samples per source. Stronger separation provides qualitative evidence of a larger GMT-dependent shift in the policy-predicted action distribution.}
\label{fig:tsne-twist2-policy}
\end{figure}

Figures~\ref{fig:tsne-sonic-policy} and~\ref{fig:tsne-twist2-policy} show the predicted-action distributions of SONIC-trained and TWIST2-trained diffusion policies, respectively. Each subplot fixes one task-specific policy and compares actions induced by SONIC versus TWIST2 observations. Overlapping colors indicate that the policy maps both observation sources to similar regions of the action space, whereas separated clusters indicate a GMT-dependent shift in the predicted whole-body action distribution. Since t-SNE is fitted independently per subplot, the coordinates should only be interpreted within each subplot; absolute positions are not comparable across tasks or policies.

The visualized distributions show that SONIC and TWIST2 observations often form source-specific regions in the predicted action space, especially in tasks involving locomotion, contact, and whole-body coordination. This pattern is consistent with the cross-GMT rollout degradation reported in the benchmark: observations collected through different GMTs are not interchangeable inputs for a fixed policy. Instead, changing the GMT shifts the coupled image--proprioception distribution and leads the same policy to produce different immediate control outputs. These t-SNE results therefore provide qualitative evidence that the SONIC--TWIST2 gap appears not only at the execution-interface level, but also in the policy-predicted action distribution.

\section{Policy Implementations and Training Details}
\label{app:implementation}

All policy baselines are implemented and trained on the LeRobot platform under a unified experimental pipeline. As summarized in Table~\ref{tab:policy_training_config}, we report the main implementation and training configurations for ACT, Diffusion Policy, Flow Matching, and $\pi_{0.5}$, covering model initialization, optimization settings, temporal prediction design, image preprocessing, training schedule, and numerical precision. To ensure a fair comparison, these baselines are trained with consistent settings whenever applicable, including the same observation length, action chunk length, inference action length, batch size, training steps, and random seed. Meanwhile, we retain the model-specific configurations required by each policy architecture, such as backbone initialization, learning rate, policy horizon, optimization objective, preprocessing pipeline, and precision format. This unified yet architecture-aware setup allows different policy baselines to be compared under controlled training conditions while preserving their standard implementation characteristics.

    \begin{table*}[htbp]
        \centering
        \caption{\textbf{Training configurations of policy baselines.}
        We summarize the key implementation and training settings used for ACT, Diffusion Policy, Flow Matching, and $\pi_{0.5}$.}
        \label{tab:policy_training_config}
        \resizebox{\textwidth}{!}{
        \begin{tabular}{c|cccc}
            \toprule
            \rowcolor{gray!15}
            \textbf{Configuration} 
            & \textbf{ACT} 
            & \textbf{Diffusion Policy} 
            & \textbf{Flow Matching} 
            & \textbf{\boldmath$\pi_{0.5}$} \\
            \midrule
            Initialization
            & ResNet-18 
            & ResNet-18 
            & CLIP ViT-B/16 
            & $\pi_{0.5}$ base checkpoint \\
            Objective 
            & -- 
            & -- 
            & Flow matching 
            & -- \\
            Observation steps 
            & 1 
            & 1 
            & 1 
            & 1 \\
            Action chunk length 
            & 20 
            & 20 
            & 20 
            & 20 \\
            Policy horizon 
            & -- 
            & 24 
            & 40 
            & -- \\
            Inference action steps 
            & 20 
            & 20 
            & 20 
            & 20 \\
            Image preprocessing 
            & Resize + padding 
            & Resize + padding 
            & Resize + padding 
            & Image transform enabled \\
            Input image size 
            & $224 \times 224$ 
            & $224 \times 224$ 
            & $224 \times 224$ 
            & -- \\
            Learning rate 
            & $1.0\times10^{-5}$ 
            & $1.0\times10^{-4}$ 
            & $2.0\times10^{-5}$ 
            & $2.5\times10^{-5}$ \\
            Batch size 
            & 64 
            & 64 
            & 64 
            & 64 \\
            Training steps 
            & 100K 
            & 100K 
            & 100K 
            & 100K \\
            Random seed 
            & 42 
            & 42 
            & 42 
            & 42 \\
            Precision
            & float32 
            & float32 
            & float32 
            & bfloat16 \\
            \bottomrule
        \end{tabular}
        }
    \end{table*}

\section{Failure Mode Analysis}
\label{app:failure}

This section analyses representative failure modes in \textsc{HumanoidArena}. 
While the main text reports task success rate and average fall rate, these aggregate metrics do not fully explain where a hierarchical humanoid policy succeeds or fails. 
In \textsc{HumanoidArena}, a rollout may fail because the high-level policy misinterprets the egocentric scene, fails to align the body with the task-relevant object, predicts an intermediate whole-body action that is difficult for the selected GMT to execute, or makes contact with the environment in a physically ineffective way. 
We therefore analyse failures at the level of rollout phases: perception and target localization, approach and body alignment, pre-contact posture preparation, contact execution, post-contact recovery, and final task completion.

\subsection{Failure Taxonomy}

We organize failures according to the stage of the hierarchical whole-body interaction pipeline at which task progress breaks down.

\paragraph{Perception and target-localization failure.}
The policy fails to localize the task-relevant object or scene region from egocentric observations. 
For Football, this may appear as walking toward an incorrect direction, losing the ball after body rotation, or failing to maintain both the ball and goal in the effective field of view.

\paragraph{Approach and body-alignment failure.}
The policy identifies the correct target but fails to bring the humanoid body into a useful interaction pose. 
For kicking, this includes incorrect root position relative to the ball, poor support-foot placement, or torso orientation that prevents a directed swing-leg trajectory.

\paragraph{Pre-contact posture failure.}
The robot approaches the object but does not form a stable pre-contact posture. 
In Football, the robot must coordinate root displacement, support-leg balance, and swing-leg preparation. 
A rollout may remain upright but still fail to produce a posture from which a strong and well-directed kick is feasible.

\paragraph{Contact-execution failure.}
The robot reaches the object but the physical contact does not produce the desired task effect. 
For Football, this includes weak impact, off-center foot--ball contact, incorrect kick direction, or insufficient momentum transfer. 
This failure mode is particularly important for leg-critical tasks because task success depends not only on reaching the object, but also on generating a physically meaningful contact event.

\paragraph{Post-contact recovery failure.}
After an imperfect contact, the robot may need to re-localize the object, recover balance, and continue task execution. 
A policy that can recover from a partial contact failure demonstrates stronger closed-loop competence than one that becomes stuck after the first unsuccessful attempt.

\paragraph{Tracker-induced execution failure.}
In cross-GMT settings, the same 40D canonical action may be realized differently by different GMT-specific adapters and low-level trackers. 
The resulting motion can differ in root displacement, swing-leg timing, torso stabilization, or egocentric camera motion, leading to task failure even when the high-level action encodes a plausible task intent.

\subsection{Football Case Study: Recoverable Contact-Execution Failure}

\begin{figure*}[t]
    \centering
    \begin{subfigure}[t]{0.32\linewidth}
        \centering
        \includegraphics[width=\linewidth]{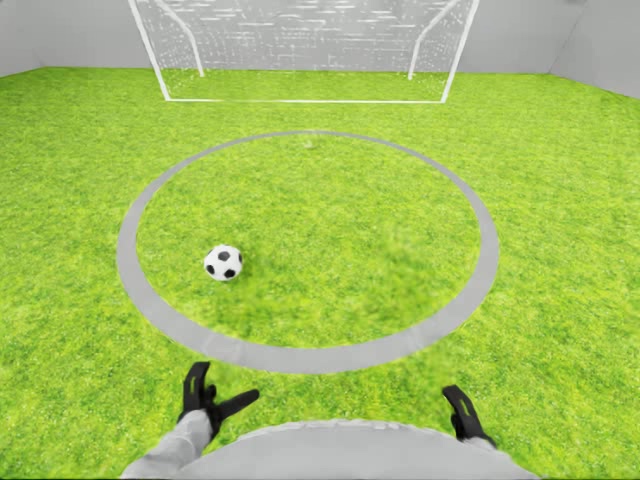}
        \caption{Initial scene}
    \end{subfigure}
    \hfill
    \begin{subfigure}[t]{0.32\linewidth}
        \centering
        \includegraphics[width=\linewidth]{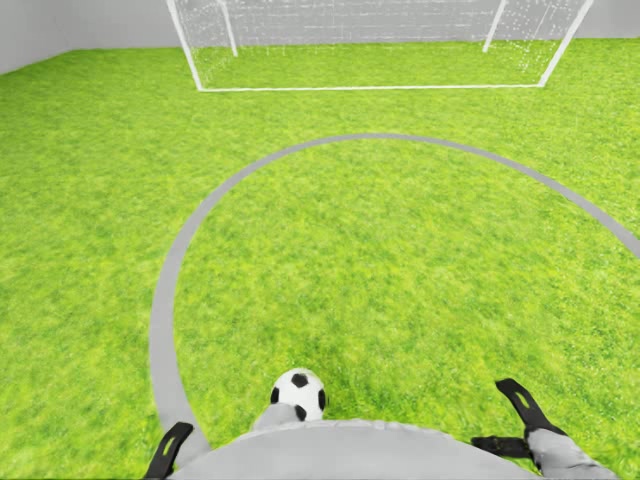}
        \caption{First approach}
    \end{subfigure}
    \hfill
    \begin{subfigure}[t]{0.32\linewidth}
        \centering
        \includegraphics[width=\linewidth]{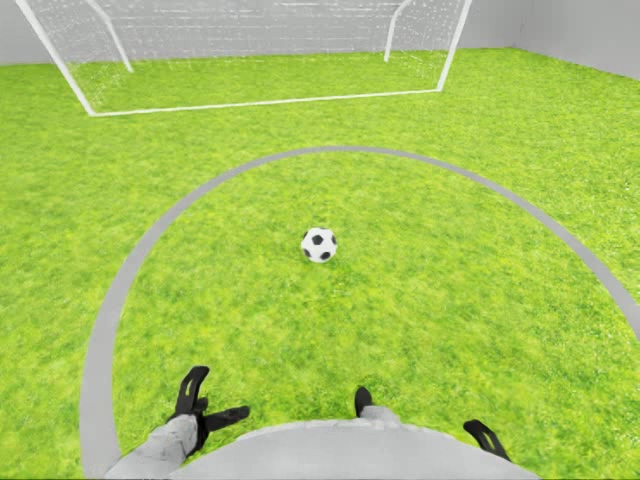}
        \caption{Partial failure}
    \end{subfigure}

    \vspace{0.5em}

    \begin{subfigure}[t]{0.32\linewidth}
        \centering
        \includegraphics[width=\linewidth]{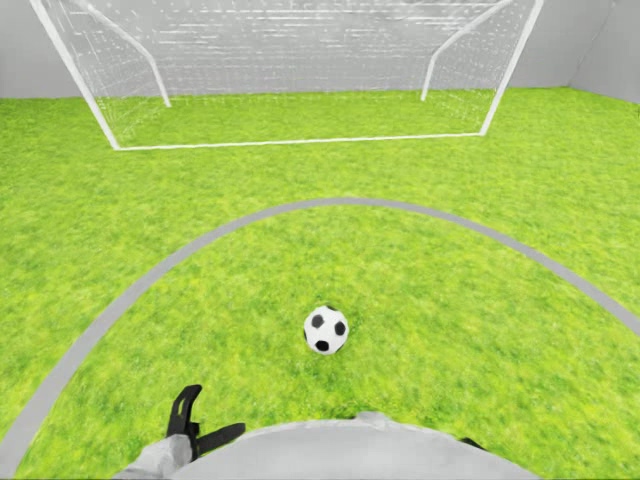}
        \caption{Re-approach}
    \end{subfigure}
    \hfill
    \begin{subfigure}[t]{0.32\linewidth}
        \centering
        \includegraphics[width=\linewidth]{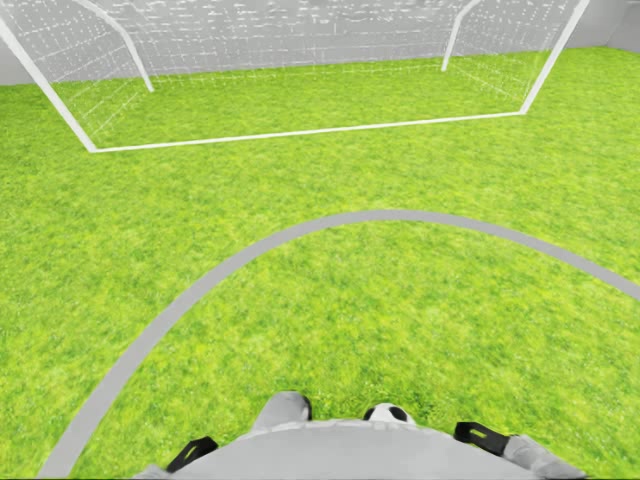}
        \caption{Pre-kick alignment}
    \end{subfigure}
    \hfill
    \begin{subfigure}[t]{0.32\linewidth}
        \centering
        \includegraphics[width=\linewidth]{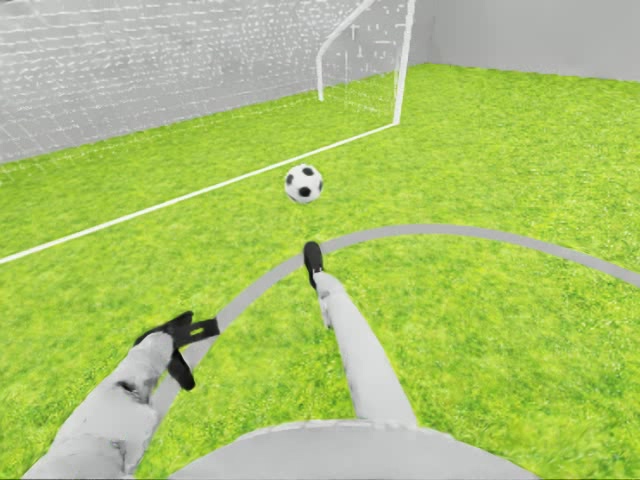}
        \caption{Successful second kick}
    \end{subfigure}

    \caption{
    Football failure-and-recovery rollout. 
    The robot first observes the ball and goal, approaches the ball, and attempts an initial kick. 
    The first contact is physically ineffective: the ball moves forward but does not travel far enough to enter the goal. 
    The policy then continues the task by re-localizing the updated ball position, reducing the distance to the ball, aligning its body and feet before contact, and executing a second kick that sends the ball into the goal. 
    This example shows that some \textsc{HumanoidArena} failures are phase-level and recoverable, rather than immediately terminal.
    }
    \label{fig:football_recovery_failure}
\end{figure*}

Figure~\ref{fig:football_recovery_failure} shows a representative Football rollout. 
The sequence illustrates a recoverable contact-execution failure: the robot initially approaches the ball and attempts a kick, but the first contact does not send the ball far enough to enter the goal. 
Instead of terminating, falling, or remaining idle, the policy continues to track the updated ball position, moves closer, re-aligns its body and feet, and performs a second kick that completes the task.

\paragraph{Stage 1: target localization.}
At the beginning of the rollout, the ball and goal are visible from the robot's egocentric view. 
The policy initiates forward movement toward the task-relevant object, indicating that the rollout does not fail at the initial perception or target-grounding stage.

\paragraph{Stage 2: first approach and kick attempt.}
The robot walks toward the ball and forms a kicking posture. 
This phase requires coordinated root displacement, support-foot placement, torso stabilization, and swing-leg preparation. 
The behavior demonstrates that the high-level policy has learned a plausible approach-and-kick pattern from the demonstration data.

\paragraph{Stage 3: partial contact failure.}
The first kick contacts the ball but does not complete the task. 
The ball moves forward but stops between the robot and the goal. 
This is not a pure perception failure, because the robot has reached the correct object and produced meaningful task progress. 
Instead, it is better categorized as a contact-execution failure: the contact direction, impact strength, or swing timing is insufficient for the ball to satisfy the goal predicate.

\paragraph{Stage 4: recovery and re-approach.}
After the unsuccessful kick, the robot does not remain fixed at the old ball location. 
It continues to move toward the new ball position. 
This behavior suggests that the high-level policy has learned a closed-loop recovery pattern from the demonstration distribution: when the first interaction does not finish the task, the robot can re-localize the object and continue acting on the updated scene state.

\paragraph{Stage 5: pre-contact re-alignment.}
Before the second kick, the robot reduces its distance to the ball and re-aligns its body. 
This stage is important because leg-object interaction depends strongly on the relative pose between the root, support foot, swing leg, ball, and goal. 
The policy does not simply repeat the previous action; it adjusts the approach based on the changed ball position.

\paragraph{Stage 6: successful second contact.}
The second kick sends the ball into the goal. 
This final stage indicates that the rollout contains a recoverable failure rather than an unrecoverable collapse. 
The policy preserves task intent after the first imperfect contact, while the GMT maintains sufficient balance and whole-body feasibility to support another attempt.

\subsection{Implications for Hierarchical Whole-Body Learning}

The failure taxonomy reveals that leg-critical HOI/HSI tasks expose diagnostic signals beyond binary success. A policy may demonstrate competent target localization and approach behavior while still failing at contact execution (e.g., Football's timing-critical kick). Binary success hides this intermediate competence, which is valuable for understanding where a hierarchical policy breaks down. Recoverable failures — where the robot re-approaches after an imperfect contact — further indicate that the policy is not replaying a fixed trajectory but conditioning on updated observations.

\subsection{Relation to Other Failure Modes}

The same rollout can transition between failure categories. A perception error (losing the ball) may lead to an approach failure (incorrect foot placement), which then produces a contact failure (weak kick). \textsc{HumanoidArena}'s perturbation axes help isolate these: visual perturbations primarily stress perception; execution perturbations stress alignment and contact; semantic perturbations stress target grounding. The behavior-level taxonomy above provides a consistent framework for mapping quantitative signals (timeout/fall/visual-decline) to the underlying failure stage.

\section{Reproducibility and Release Details}
\label{app:release}
\paragraph{Data collection.}
All demonstration data are collected in Isaac Lab through teleoperated control. During collection, the operator receives the humanoid's egocentric visual stream and
controls the robot in closed loop; the resulting trajectories are recorded with synchronized observations, proprioceptive states, intermediate whole-body actions, task
metadata, and GMT backend information.

\paragraph{Compute resources.}
All policy models are trained on an 8-GPU NVIDIA H200 server. Validation and benchmark evaluation are performed on a 16-GPU NVIDIA RTX 4090 server. We use the RTX 4090
server for evaluation because our Isaac Lab evaluation pipeline requires graphical rendering, which is not available on the H-series training nodes in our compute
environment.

\paragraph{Assets and licenses.}
The paper cites and credits existing assets, including Isaac Lab (BSD-3-Clause, with Isaac Sim subject to NVIDIA Isaac Sim terms), Unitree simulation/G1-related resources
(Apache-2.0 where derived from \texttt{unitree\_sim\_isaaclab}), GMR (MIT), TWIST2 (MIT), SONIC/GR00T-WholeBodyControl (Apache-2.0), LeRobot and Unitree-LeRobot components
(Apache-2.0), ACT and Diffusion Policy baselines (MIT), Flow Matching implementations based on LeRobot or related open implementations (Apache-2.0 where
applicable), $\pi_{0.5}$/OpenPI code (Apache-2.0), XRoboToolkit teleoperation components used in the data stream (MIT), and Synthesis/Extwin scene assets, which are tracked
through an asset manifest and redistributed only when permitted by the corresponding asset license or terms of use.

\section{More Results}
\label{app:moreresults}
Table~\ref{table:physical-results-full} reports the full perturbation results for in-GMT evaluation, including per-task success rates under the in-distribution, visual, semantic, and execution settings.
These results complement the suite-level robustness profiles in Figure~\ref{fig:perturbation} and provide detailed task-level evidence for the analysis in Section~\ref{sec:experiments}.

\begin{table*}[t]
  \caption{
  \textbf{In-GMT evaluation.}
  We evaluate high-level policy baselines with matched training and deployment GMTs, using either TWIST2 or SONIC as the low-level execution backend. Results are reported as success rate (SR, $\uparrow$; mean $\pm$ standard deviation) on the in-distribution setting and the visual, semantic, and execution
  generalization splits. AVG denotes the average performance within each HOI or HSI task suite.
  The best and second-best SRs for each task and suite average under the same GMT and evaluation split are highlighted in \textbf{bold} and \underline{underlined}, respectively.
  }
  \label{table:physical-results-full}
  \centering
  \begin{adjustbox}{max width=\textwidth}
  \setlength{\tabcolsep}{4.2pt}
  \renewcommand{\arraystretch}{1.08}

  \begin{tabular}{ccc|cccc|ccccc}
  \toprule
  \rowcolor{mygray}
  & & &
  \multicolumn{4}{c|}{HOI} &
  \multicolumn{5}{c}{HSI} \\

  \rowcolor{mygray}
  \multirow{-2}{*}{Difficulty} &
  \multirow{-2}{*}{Method} &
  \multirow{-2}{*}{AFR($\downarrow$)} &
  \textsc{Football} & \textsc{DoubleDesk} & \textsc{P\&Pbox} & \textsc{Avg} &
  \textsc{OpenDoor} & \textsc{SitSofa} & \textsc{Boxing} & \textsc{VisNavi} & \textsc{Avg} \\
  \midrule
  \midrule

  \rowcolor{mygray}\multicolumn{12}{c}{\textbf{TWIST2}} \\
  \multirow{4}{*}{Base} & ACT~\cite{ACT} & 6.43\% & 26.7$\pm$6.2\% & \underline{15.0$\pm$8.2\%} & 43.3$\pm$8.5\% & 28.33$\pm$13.94\% & 50.0$\pm$4.1\% & 66.7$\pm$10.3\% & \underline{58.3$\pm$10.3\%} & 13.3$\pm$4.7\% & 47.08$\pm$21.84\% \\
                              & DP~\cite{chi2023diffusionpolicy} & 10.48\% & \underline{46.7$\pm$6.2\%} & 10.0$\pm$4.1\% & \textbf{50.0$\pm$0.0\%} & \underline{35.56$\pm$18.63\%} & \underline{68.3$\pm$6.2\%} & \textbf{73.3$\pm$6.2\%} & 51.7$\pm$11.8\% & \textbf{30.0$\pm$4.1\%} & \underline{55.83$\pm$18.58\%} \\
                              & FM~\cite{FM_objective} & 7.38\% & \textbf{53.3$\pm$8.5\%} & \textbf{16.7$\pm$8.5\%} & 38.3$\pm$8.5\% & \textbf{36.11$\pm$17.28\%} & \textbf{76.7$\pm$2.4\%} & \underline{68.3$\pm$6.2\%} & \textbf{63.3$\pm$9.4\%} & \underline{26.7$\pm$20.9\%} & \textbf{58.75$\pm$22.56\%} \\
                              & $\pi_{0.5}$~\cite{pi05} & 3.33\% & 26.7$\pm$4.7\% & 1.7$\pm$2.4\% & \underline{46.7$\pm$8.5\%} & 25.00$\pm$19.29\% & 28.3$\pm$9.4\% & 60.0$\pm$4.1\% & 51.7$\pm$8.5\% & 13.3$\pm$8.5\% & 38.33$\pm$20.14\% \\
  \midrule
  \multirow{4}{*}{Visual} & ACT~\cite{ACT} & 6.67\% & 15.0$\pm$8.2\% & 0.0$\pm$0.0\% & 28.3$\pm$2.4\% & 14.44$\pm$12.57\% & 43.3$\pm$2.4\% & 51.7$\pm$8.5\% & \textbf{63.3$\pm$6.2\%} & 23.3$\pm$12.5\% & 45.42$\pm$16.77\% \\
                              & DP~\cite{chi2023diffusionpolicy} & 12.38\% & \textbf{40.0$\pm$7.1\%} & \textbf{8.3$\pm$8.5\%} & \textbf{41.7$\pm$6.2\%} & \textbf{30.00$\pm$17.00\%} & \textbf{68.3$\pm$4.7\%} & \textbf{78.3$\pm$2.4\%} & 51.7$\pm$8.5\% & \underline{25.0$\pm$4.1\%} & \textbf{55.83$\pm$20.90\%} \\
                              & FM~\cite{FM_objective} & 9.05\% & \underline{26.7$\pm$4.7\%} & \underline{5.0$\pm$4.1\%} & 23.3$\pm$6.2\% & 18.33$\pm$10.80\% & \underline{65.0$\pm$10.8\%} & \underline{53.3$\pm$6.2\%} & \underline{56.7$\pm$18.9\%} & \textbf{30.0$\pm$10.8\%} & \underline{51.25$\pm$18.04\%} \\
                              & $\pi_{0.5}$~\cite{pi05} & 1.67\% & 25.0$\pm$4.1\% & 1.7$\pm$2.4\% & \underline{33.3$\pm$11.8\%} & \underline{20.00$\pm$15.28\%} & 31.7$\pm$2.4\% & 43.3$\pm$10.3\% & 43.3$\pm$13.1\% & 5.0$\pm$4.1\% & 30.83$\pm$17.89\% \\
  \midrule
  \multirow{4}{*}{Semantic} & ACT~\cite{ACT} & 4.29\% & 20.0$\pm$4.1\% & 6.7$\pm$2.4\% & 35.0$\pm$8.2\% & 20.56$\pm$12.79\% & \underline{46.7$\pm$8.5\%} & 51.7$\pm$17.0\% & \textbf{61.7$\pm$11.8\%} & \underline{33.3$\pm$15.5\%} & \underline{48.33$\pm$17.00\%} \\
                              & DP~\cite{chi2023diffusionpolicy} & 6.43\% & \underline{50.0$\pm$0.0\%} & \textbf{16.7$\pm$4.7\%} & \underline{46.7$\pm$6.2\%} & \textbf{37.78$\pm$15.65\%} & \textbf{56.7$\pm$9.4\%} & \textbf{75.0$\pm$10.8\%} & \underline{60.0$\pm$4.1\%} & \textbf{40.0$\pm$8.2\%} & \textbf{57.92$
                              \pm$15.06\%} \\
                              & FM~\cite{FM_objective} & 7.14\% & \textbf{58.3$\pm$6.2\%} & \underline{8.3$\pm$2.4\%} & 43.3$\pm$4.7\% & \underline{36.67$\pm$21.47\%} & 23.3$\pm$13.1\% & \underline{66.7$\pm$8.5\%} & 60.0$\pm$10.8\% & 20.0$\pm$14.1\% & 42.50$\pm$24.11\% \\
                              & $\pi_{0.5}$~\cite{pi05} & 1.43\% & 13.3$\pm$9.4\% & 0.0$\pm$0.0\% & \textbf{61.7$\pm$12.5\%} & 25.00$\pm$27.99\% & 6.7$\pm$2.4\% & 50.0$\pm$4.1\% & 55.0$\pm$4.1\% & 16.7$\pm$9.4\% & 32.08$\pm$21.55\% \\
  \midrule
  \multirow{4}{*}{Execution} & ACT~\cite{ACT} & 5.71\% & 26.7$\pm$13.1\% & \textbf{13.3$\pm$2.4\%} & 33.3$\pm$13.1\% & 24.44$\pm$13.63\% & 48.3$\pm$15.5\% & \underline{65.0$\pm$4.1\%} & \underline{55.0$\pm$12.2\%} & \underline{35.0$\pm$10.8\%} & 50.83$\pm$15.79\% \\
                              & DP~\cite{chi2023diffusionpolicy} & 8.33\% & \textbf{43.3$\pm$4.7\%} & 8.3$\pm$4.7\% & \textbf{45.0$\pm$4.1\%} & \textbf{32.22$\pm$17.50\%} & \underline{63.3$\pm$6.2\%} & 60.0$\pm$10.8\% & 48.3$\pm$10.3\% & \textbf{36.7$\pm$13.1\%} & \underline{52.08$\pm$14.78\%} \\
                              & FM~\cite{FM_objective} & 6.67\% & \underline{41.7$\pm$6.2\%} & \underline{13.3$\pm$4.7\%} & \underline{33.3$\pm$6.2\%} & \underline{29.44$\pm$13.22\%} & \textbf{80.0$\pm$8.2\%} & \textbf{76.7$\pm$8.5\%} & \textbf{58.3$\pm$13.1\%} & 23.3$\pm$9.4\% & \textbf{59.58$\pm$24.62\%} \\
                              & $\pi_{0.5}$~\cite{pi05} & 2.62\% & 21.7$\pm$2.4\% & 0.0$\pm$0.0\% & 33.3$\pm$8.5\% & 18.33$\pm$14.72\% & 31.7$\pm$13.1\% & 51.7$\pm$8.5\% & 50.0$\pm$7.1\% & 16.7$\pm$8.5\% & 37.50$\pm$17.26\% \\
  \midrule

  \rowcolor{mygray}\multicolumn{12}{c}{\textbf{SONIC}} \\
  \multirow{4}{*}{Base} & ACT~\cite{ACT} & 8.57\% & \underline{16.7$\pm$6.2\%} & 18.3$\pm$4.7\% & 56.7$\pm$6.2\% & 30.56$\pm$19.36\% & \underline{78.3$\pm$9.4\%} & 73.3$\pm$8.5\% & 56.7$\pm$6.2\% & \underline{33.3$\pm$2.4\%} & \underline{60.42$\pm$18.98\%} \\
                              & DP~\cite{chi2023diffusionpolicy} & 8.33\% & \textbf{45.0$\pm$10.8\%} & 36.7$\pm$4.7\% & \textbf{75.0$\pm$4.1\%} & \textbf{52.22$\pm$17.97\%} & \textbf{85.0$\pm$10.8\%} & \textbf{78.3$\pm$14.3\%} & \textbf{76.7$\pm$2.4\%} & 23.3$\pm$6.2\% & \textbf{65.83$\pm$26.52\%} \\
                              & FM~\cite{FM_objective} & 5.71\% & 13.3$\pm$2.4\% & \underline{38.3$\pm$4.7\%} & \underline{73.3$\pm$6.2\%} & \underline{41.67$\pm$25.06\%} & 70.0$\pm$4.1\% & 15.0$\pm$7.1\% & \underline{70.0$\pm$8.2\%} & \textbf{38.3$\pm$14.3\%} & 48.33$\pm$24.94\% \\
                              & $\pi_{0.5}$~\cite{pi05} & 5.24\% & 10.0$\pm$4.1\% & \textbf{43.3$\pm$6.2\%} & 71.7$\pm$11.8\% & 41.67$\pm$26.46\% & 66.7$\pm$6.2\% & \underline{73.3$\pm$2.4\%} & 70.0$\pm$0.0\% & 23.3$\pm$6.2\% & 58.33$\pm$20.85\% \\
  \midrule
  \multirow{4}{*}{Visual} & ACT~\cite{ACT} & 9.76\% & \underline{18.3$\pm$2.4\%} & 6.7$\pm$2.4\% & 50.0$\pm$10.8\% & 25.00$\pm$19.44\% & \underline{63.3$\pm$2.4\%} & \underline{35.0$\pm$8.2\%} & 53.3$\pm$10.3\% & \underline{38.3$\pm$2.4\%} & \underline{47.50$\pm$13.31\%} \\
                              & DP~\cite{chi2023diffusionpolicy} & 8.33\% & \textbf{30.0$\pm$4.1\%} & \textbf{28.3$\pm$2.4\%} & \underline{55.0$\pm$4.1\%} & \textbf{37.78$\pm$12.72\%} & \textbf{71.7$\pm$8.5\%} & \textbf{46.7$\pm$22.5\%} & \textbf{75.0$\pm$14.1\%} & 35.0$\pm$4.1\% & \textbf{57.08$\pm$21.93\%} \\
                              & FM~\cite{FM_objective} & 8.10\% & 6.7$\pm$6.2\% & 13.3$\pm$6.2\% & 46.7$\pm$20.9\% & 22.22$\pm$21.87\% & 31.7$\pm$4.7\% & 8.3$\pm$2.4\% & \underline{73.3$\pm$2.4\%} & \textbf{41.7$\pm$8.5\%} & 38.75$\pm$23.90\% \\
                              & $\pi_{0.5}$~\cite{pi05} & 5.00\% & 15.0$\pm$0.0\% & \underline{13.3$\pm$2.4\%} & \textbf{73.3$\pm$10.3\%} & \underline{33.89$\pm$28.56\%} & 63.3$\pm$2.4\% & 35.0$\pm$10.8\% & 63.3$\pm$13.1\% & 18.3$\pm$6.2\% & 45.00$\pm$21.31\% \\
  \midrule
  \multirow{4}{*}{Semantic} & ACT~\cite{ACT} & 6.43\% & 8.3$\pm$6.2\% & \underline{15.0$\pm$4.1\%} & \underline{63.3$\pm$2.4\%} & 28.89$\pm$24.92\% & \underline{56.7$\pm$6.2\%} & 21.7$\pm$6.2\% & 46.7$\pm$10.3\% & \underline{30.0$\pm$8.2\%} & \underline{38.75$\pm$15.83\%} \\
                              & DP~\cite{chi2023diffusionpolicy} & 6.67\% & \textbf{33.3$\pm$14.3\%} & 13.3$\pm$2.4\% & 56.7$\pm$6.2\% & \underline{34.44$\pm$19.92\%} & \textbf{78.3$\pm$2.4\%} & \textbf{60.0$\pm$7.1\%} & \textbf{71.7$\pm$8.5\%} & 26.7$\pm$8.5\% & \textbf{59.17$\pm$21.10\%} \\
                              & FM~\cite{FM_objective} & 5.71\% & 8.3$\pm$8.5\% & 8.3$\pm$4.7\% & 60.0$\pm$4.1\% & 25.56$\pm$25.10\% & 3.3$\pm$4.7\% & 28.3$\pm$2.4\% & \underline{68.3$\pm$13.1\%} & \textbf{35.0$\pm$7.1\%} & 33.75$\pm$24.51\% \\
                              & $\pi_{0.5}$~\cite{pi05} & 2.38\% & \underline{16.7$\pm$8.5\%} & \textbf{25.0$\pm$0.0\%} & \textbf{75.0$\pm$7.1\%} & \textbf{38.89$\pm$26.54\%} & 8.3$\pm$4.7\% & \underline{41.7$\pm$13.1\%} & 65.0$\pm$4.1\% & 16.7$\pm$2.4\% & 32.92$\pm$23.40\% \\
  \midrule
  \multirow{4}{*}{Execution} & ACT~\cite{ACT} & 6.90\% & \underline{23.3$\pm$6.2\%} & 11.7$\pm$8.5\% & 65.0$\pm$4.1\% & 33.33$\pm$23.80\% & 68.3$\pm$13.1\% & \underline{50.0$\pm$4.1\%} & 46.7$\pm$6.2\% & \underline{36.7$\pm$6.2\%} & \underline{50.42$\pm$14.06\%} \\
                              & DP~\cite{chi2023diffusionpolicy} & 5.00\% & \textbf{45.0$\pm$4.1\%} & \underline{30.0$\pm$14.7\%} & \textbf{78.3$\pm$6.2\%} & \textbf{51.11$\pm$22.33\%} & \underline{81.7$\pm$6.2\%} & \textbf{80.0$\pm$4.1\%} & \textbf{80.0$\pm$7.1\%} & 33.3$\pm$11.8\% & \textbf{68.75$\pm$21.90\%} \\
                              & FM~\cite{FM_objective} & 5.24\% & 20.0$\pm$8.2\% & \textbf{31.7$\pm$8.5\%} & 63.3$\pm$8.5\% & 38.33$\pm$20.14\% & \textbf{83.3$\pm$4.7\%} & 6.7$\pm$4.7\% & 66.7$\pm$11.8\% & \textbf{45.0$\pm$4.1\%} & 50.42$\pm$29.54\% \\
                              & $\pi_{0.5}$~\cite{pi05} & 5.48\% & 15.0$\pm$4.1\% & 26.7$\pm$2.4\% & \underline{78.3$\pm$6.2\%} & \underline{40.00$\pm$27.89\%} & 63.3$\pm$6.2\% & 41.7$\pm$9.4\% & \underline{68.3$\pm$6.2\%} & 18.3$\pm$2.4\% & 47.92$\pm$20.86\% \\

  \bottomrule
  \end{tabular}
  \end{adjustbox}
  \end{table*}

Table~\ref{table:mergeall} further studies whether the shared intermediate whole-body action interface enables dataset transfer across tasks and GMT backends. Compared with the base in-GMT results in Table~\ref{table:physical-results-full}, merged training brings clear gains in several settings. Under TWIST2$\times$7, DP improves from 35.56$\pm$18.63\% to 38.89$\pm$20.25\% on HOI, and more substantially from 55.83$\pm$18.58\% to 73.33$\pm$23.48\% on HSI. $\pi_{0.5}$ also improves on TWIST2 HSI from 38.33$\pm$20.14\% to 55.00$\pm$25.50\%. Under SONIC$\times$7, $\pi_{0.5}$ improves its HSI average from 58.33$\pm$20.85\% to 65.42$\pm$21.06\%. These results suggest that the canonical action interface can reuse task-level whole-body intent across tasks, supporting effective dataset transfer beyond isolated task training.

However, the gains are not uniform. For example, SONIC$\times$7 reduces DP on HOI from 52.22$\pm$17.97\% to 29.44$\pm$23.27\%, and TWIST2+SONIC$\times$14 improves DP on HSI to 68.12$\pm$23.00\% but leaves HOI at 31.67$\pm$26.11\%. We attribute these drops to GMT-conditioned action distribution mismatch. Although different GMT datasets share the same 40D action space, TWIST2~\cite{ze2025twist2} and SONIC~\cite{luo2025sonic} differ in tracking capability, motion priors, stabilization behavior, and contact response. During closed-loop teleoperation, operators therefore rely on different corrective motions, foot placements, body postures, and contact timings to complete the same task. Such GMT-specific realizations have a stronger effect on contact-rich HOI tasks, where small execution differences can affect grasp stability, kicking timing, and object-contact effectiveness. Overall, merge-all training demonstrates that the shared intermediate action interface provides a practical basis for dataset transfer, while revealing how GMT-conditioned execution diversity shapes the effectiveness of multi-GMT policy learning.

\begin{table*}[t]
  \caption{
  \textbf{Merge-All evaluation.}
  We evaluate high-level policy baselines trained with merged demonstrations from TWIST2-only seven task-GMT settings, SONIC-only seven task-GMT settings, or the combined TWIST2+SONIC fourteen task-GMT settings. Results are reported as success rate (SR, $\uparrow$; mean $\pm$ standard deviation). AVG denotes the
  average SR within each HOI or HSI task suite.
  The best and second-best distinct AVG are highlighted in \textbf{bold} and \underline{underlined}, respectively.
  }
  \label{table:mergeall}
  \centering
  \begin{adjustbox}{max width=\textwidth}
  \setlength{\tabcolsep}{4.2pt}
  \renewcommand{\arraystretch}{1.08}

  \begin{tabular}{ccccccc|ccccc}
  \toprule
  & & &
  \multicolumn{4}{c|}{HOI} &
  \multicolumn{5}{c}{HSI} \\
  \cmidrule(lr){4-7} \cmidrule(lr){8-12}

  \multirow{-2}{*}{Setting} &
  \multirow{-2}{*}{Method} &
  \multirow{-2}{*}{AFR~($\downarrow$)} &
  \textsc{Football} & \textsc{DoubleDesk} & \textsc{P\&Pbox} & \textsc{Avg~($\uparrow$)} &
  \textsc{OpenDoor} & \textsc{SitSofa} & \textsc{Boxing} & \textsc{VisNavi} & \textsc{Avg~($\uparrow$)} \\
  \midrule

  \multirow{4}{*}{TWIST2$\times$7} & ACT~\cite{ACT} & 7.75\% & 48.3$\pm$2.4\% & 3.3$\pm$2.4\% & 17.5$\pm$2.5\% & \cellcolor{mygray}23.75$\pm$19.96\% & 21.7$\pm$4.7\% & 78.3$\pm$6.2\% & 8.3$\pm$6.2\% & 18.3$\pm$16.5\% & \cellcolor{mygray}31.67$\pm$29.04\% \\
    & DP~\cite{chi2023diffusionpolicy} & 7.62\% & 61.7$\pm$9.4\% & 18.3$\pm$12.5\% & 36.7$\pm$6.2\% & \textbf{\cellcolor{mygray}38.89$\pm$20.25\%} & 93.3$\pm$6.2\% & 91.7$\pm$2.4\% & 61.7$\pm$8.5\% & 46.7$\pm$22.5\% & \textbf{\cellcolor{mygray}73.33$\pm$23.48\%} \\
    & FM~\cite{FM_objective} & 6.90\% & 35.0$\pm$4.1\% & 18.3$\pm$2.4\% & 30.0$\pm$4.1\% & \underline{\cellcolor{mygray}27.78$\pm$7.86\%} & 68.3$\pm$6.2\% & 68.3$\pm$6.2\% & 55.0$\pm$8.2\% & 25.0$\pm$7.1\% & \cellcolor{mygray}54.17$\pm$19.02\% \\
    & $\pi_{0.5}$~\cite{pi05} & 4.52\% & 33.3$\pm$8.5\% & 0.0$\pm$0.0\% & 43.3$\pm$6.2\% & \cellcolor{mygray}25.56$\pm$19.50\% & 73.3$\pm$6.2\% & 81.7$\pm$4.7\% & 46.7$\pm$6.2\% & 18.3$\pm$6.2\% & \underline{\cellcolor{mygray}55.00$\pm$25.50\%} \\

  \midrule

  \multirow{4}{*}{SONIC$\times$7} & ACT~\cite{ACT} & 7.38\% & 20.0$\pm$4.1\% & 30.0$\pm$4.1\% & 60.0$\pm$12.2\% & \underline{\cellcolor{mygray}36.67$\pm$18.71\%} & 58.3$\pm$2.4\% & 21.7$\pm$2.4\% & 58.3$\pm$10.3\% & 28.3$\pm$4.7\% & \cellcolor{mygray}41.67$\pm$17.83\% \\
    & DP~\cite{chi2023diffusionpolicy} & 7.14\% & 53.3$\pm$6.2\% & 35.0$\pm$10.8\% & 0.0$\pm$0.0\% & \cellcolor{mygray}29.44$\pm$23.27\% & 83.3$\pm$6.2\% & 83.3$\pm$4.7\% & 21.7$\pm$30.6\% & 50.0$\pm$14.7\% & \underline{\cellcolor{mygray}59.58$\pm$31.12\%} \\
    & FM~\cite{FM_objective} & 7.14\% & 21.7$\pm$8.5\% & 20.0$\pm$7.1\% & 0.0$\pm$0.0\% & \cellcolor{mygray}13.89$\pm$11.73\% & 76.7$\pm$2.4\% & 48.3$\pm$6.2\% & 18.3$\pm$25.9\% & 38.3$\pm$6.2\% & \cellcolor{mygray}45.42$\pm$25.12\% \\
    & $\pi_{0.5}$~\cite{pi05} & 6.43\% & 10.0$\pm$7.1\% & 30.0$\pm$7.1\% & 75.0$\pm$4.1\% & \textbf{\cellcolor{mygray}38.33$\pm$27.89\%} & 85.0$\pm$4.1\% & 70.0$\pm$14.7\% & 71.7$\pm$12.5\% & 35.0$\pm$4.1\% & \textbf{\cellcolor{mygray}65.42$\pm$21.06\%} \\

  \midrule

  \multirow{4}{*}{TWIST2+SONIC$\times$14} & ACT~\cite{ACT} & 6.95\% & 32.5$\pm$18.4\% & 9.2$\pm$10.2\% & 36.0$\pm$12.8\% & \underline{\cellcolor{mygray}25.29$\pm$18.67\%} & 10.0$\pm$2.9\% & 34.2$\pm$6.7\% & 53.3$\pm$16.7\% & 13.3$\pm$9.0\% & \cellcolor{mygray}27.71$\pm$20.21\% \\
    & DP~\cite{chi2023diffusionpolicy} & 9.40\% & 33.3$\pm$5.5\% & 17.5$\pm$12.8\% & 44.2$\pm$30.2\% & \cellcolor{mygray}31.67$\pm$26.11\% & 82.5$\pm$4.8\% & 89.2$\pm$9.8\% & 63.3$\pm$9.0\% & 37.5$\pm$17.5\% & \textbf{\cellcolor{mygray}68.12$\pm$23.00\%} \\
    & FM~\cite{FM_objective} & 9.88\% & 26.7$\pm$12.1\% & 18.3$\pm$13.4\% & 28.3$\pm$26.9\% & \cellcolor{mygray}24.44$\pm$19.21\% & 13.3$\pm$13.1\% & 40.8$\pm$24.7\% & 36.7$\pm$15.5\% & 10.0$\pm$7.6\% & \cellcolor{mygray}25.21$\pm$21.38\% \\
    & $\pi_{0.5}$~\cite{pi05} & 8.69\% & 35.0$\pm$23.8\% & 15.0$\pm$15.8\% & 45.0$\pm$25.3\% & \textbf{\cellcolor{mygray}31.67$\pm$25.33\%} & 51.7$\pm$21.5\% & 58.3$\pm$21.5\% & 55.8$\pm$11.7\% & 23.3$\pm$10.3\% & \underline{\cellcolor{mygray}47.29$\pm$22.13\%} \\

  \bottomrule
  \end{tabular}
  \end{adjustbox}
  \vspace{-10pt}
  \end{table*}
  
\section{Limitations and Future Directions}
\label{app:discussion}

\paragraph{Limited task and scene coverage.}
The current benchmark contains seven leg-critical HOI/HSI tasks covering object interaction, scene interaction, navigation, kicking, sitting, door opening, and target striking. 
Although these tasks expose important whole-body behaviors such as foot placement, balance maintenance, posture adjustment, and contact-rich execution, they still represent only a subset of everyday humanoid interaction. 
Future versions can expand the benchmark with richer household activities, multi-object rearrangement, human-robot interaction, tool use, outdoor locomotion-interaction tasks, and longer-horizon task sequences.

%%%%%%%%%%%%%%%%%%%%%%%%%%%%%%%%%%%%%%%%%%%%%%%%%%%%%%%%%%%%

\newpage

\end{document}